\documentclass{article}

\usepackage[final]{neurips_2022}

\usepackage{microtype}
\usepackage{graphicx}
\usepackage{booktabs} % for professional tables

\usepackage{hyperref}

\usepackage{amsfonts}       % blackboard math symbols
\usepackage{nicefrac}       % compact symbols for 1/2, etc.
\usepackage{microtype}      % microtypography
\usepackage{enumitem}
\usepackage{mathtools}
\usepackage{makecell}
\usepackage{multirow}
\usepackage{wasysym}
\usepackage{multirow}

\usepackage{caption}
\usepackage{subcaption}
\usepackage{wrapfig}
\usepackage{units}
\usepackage[dvipsnames]{xcolor}

\usepackage[ruled,vlined]{algorithm2e}
\usepackage{float}

\usepackage{amsmath,amssymb,amsthm}

\newtheorem{lemma}{Lemma}

\DeclarePairedDelimiter{\norm}{\lVert}{\rVert} 

\usepackage{enumitem}
\setlist{leftmargin=*,itemsep=0pt}

\DeclareMathOperator*{\argmax}{arg\,max}
\DeclareMathOperator*{\argmin}{arg\,min}

\title{Wasserstein Iterative Networks\\for Barycenter Estimation}

\author{%
  Alexander Korotin
    \\
  Skolkovo Institute of Science and Technology\\
  Artificial Intelligence Research Institute\\
  \textit{Moscow, Russia} \\
  \texttt{a.korotin@skoltech.ru} \\
  \And
  Vage Egiazarian \\
Skolkovo Institute of Science and Technology \\
  \textit{Moscow, Russia} \\
\texttt{vage.egiazarian@skoltech.ru} \\
  \And
  Lingxiao Li \\
Massachusetts Institute of Technology \\
\textit{Cambridge, Massachusetts, USA} \\
\texttt{lingxiao@mit.edu} \\
  \And
Evgeny Burnaev
    \\
  Skolkovo Institute of Science and Technology\\
  Artificial Intelligence Research Institute\\
  \textit{Moscow, Russia} \\
  \texttt{e.burnaev@skoltech.ru} \\
}

\begin{document}

\maketitle

\vspace{-4mm}\begin{abstract}
\vspace{-2mm}Wasserstein barycenters have become popular due to their ability to represent the average of probability measures in a geometrically meaningful way. In this paper, we present an algorithm to approximate the Wasserstein-2 barycenters of continuous measures via a generative model. Previous approaches rely on regularization (entropic/quadratic) which introduces bias or on input convex neural networks which are not expressive enough for large-scale tasks. In contrast, our algorithm does not introduce bias and allows using arbitrary neural networks. In addition, based on the celebrity faces dataset, we construct Ave, celeba! \textit{dataset} which can be used for quantitative evaluation of barycenter algorithms by using standard metrics of generative models such as FID. 
\end{abstract}

\vspace{-3mm}\section{Introduction}
\label{sec-intro}

\vspace{-2mm}\begin{wrapfigure}{r}{0.5\textwidth}\vspace{-10mm}
\includegraphics[width=\linewidth]{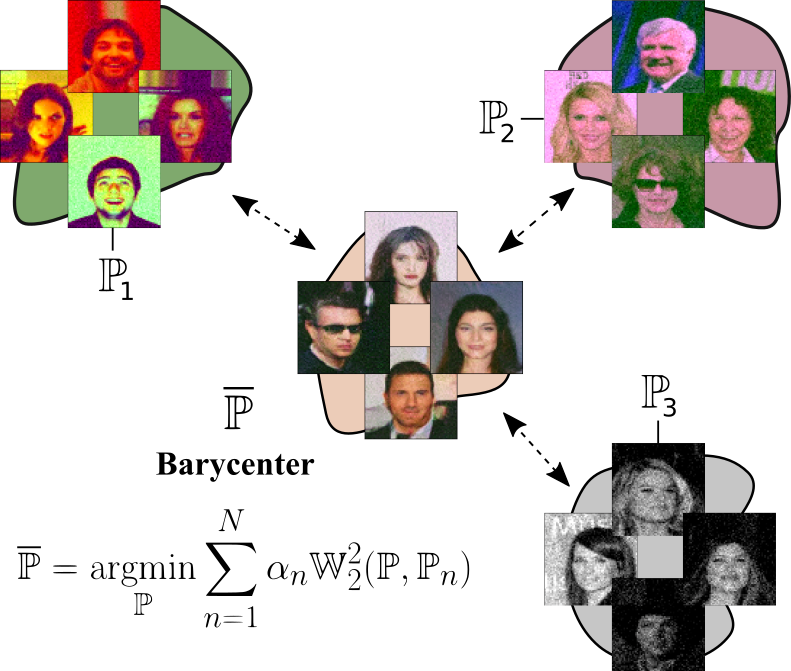}
\vspace{-3mm}\caption{\centering The barycenter w.r.t. weights ${(\alpha_{1},\alpha_{2},\alpha_{3})=(\frac{1}{4},\frac{1}{2},\frac{1}{4})}$ of Ave, Celeba! subsets computed by our Algorithm \ref{algorithm-win}. \newline The figure shows random samples from the input subsets and generated images from the barycenter.}
% \lingxiao{The 4 images are not pushforward of each other via the arrows right? Maybe we need to clarify this in the caption}\alex{Is this better (added the last sentence)?}
\label{fig:ave-celeba-teaser}\vspace{-4mm}
\end{wrapfigure}Wasserstein barycenters \citep{agueh2011barycenters} provide a geometrically meaningful notion of the average of probability measures based on optimal transport (OT, see \cite{villani2008optimal}). Methods for computing barycenters have been successfully applied to various practical problems. In geometry processing, shape interpolation can be performed by barycenters \citep{solomon2015convolutional}. In image processing, barycenters are used for color and style translation \citep{rabin2014adaptive,mroueh2019wasserstein}, texture mixing \citep{rabin2011wasserstein} and image interpolation \citep{lacombe2021learning,simon2020barycenters}. In language processing, barycenters can be applied to text evaluation \citep{colombo2021automatic}. In online learning, barycenters are used for aggregating probabilistic forecasts of experts \citep{korotin2021mixability,paris2021online,koldasbayeva2022large}. In Bayesian inference, the barycenter of subset posteriors converges to the full data posterior \citep{srivastava2015wasp,srivastava2018scalable} allowing efficient computation of full posterior based on barycenters. In reinforcement learning, barycenters are used for uncertainty propagation \citep{metelli2019propagating}. Other applications are data augmentation \citep{bespalov2021data}, multivariate density registration \citep{bigot2019data}, distributions alignment \citep{inouye2021iterative}, domain generalization \citep{lyu2021barycenteric} and adaptation \cite{montesuma2021wasserstein}, model ensembling \citep{dognin2019wasserstein}, averaging of persistence diagrams \cite{vidal2019progressive,barannikov2021manifold,barannikov2021representation}.

% ,li2020continuous}.
% \lingxiao{This paragraph could use some better transitions, instead of just listing ``In XXX, .... In XXX, ....''}

% Thus, Wasserstein-2 barycenter provides the weighted averaging of probability measures in a suitable sense.
% behaving a lot similar to averaging of vectors in $\mathcal{L}^{2}$.

The bottleneck of obtaining barycenters is the computational complexity. For \textit{discrete} measures, fast and accurate barycenter algorithms exist for low-dimensional problems; see \cite{peyre2019computational} for a survey. 
However, discrete methods scale poorly with the number of support points of the barycenter. Consequently, they cannot approximate continuous barycenters well, especially in high dimensions.

% For \textit{continuous} measures, only a few algorithms exist \citep{li2020continuous, fan2020scalable, korotin2021continuous}.
% , see \citep[Tables 1-3]{korotin2021continuous} or \citep[Figure 5]{fan2020scalable} for comparisons with continuous methods.

Existing continuous barycenter approaches \citep{li2020continuous, fan2020scalable, korotin2021continuous} are mostly based on entropic/quadratic regularization or parametrization of Brenier potentials with input-convex neural networks (ICNNs, see \cite{amos2017input}). The regularization-based $[\text{CR}\mathbb{W}\text{B}]$ algorithm by \cite{li2020continuous} recovers a barycenter \textit{biased} from the true one. Algorithms $[\text{C}\mathbb{W}_{2}\text{B}]$ by \cite{korotin2021continuous} and $[\text{SC}\mathbb{W}_{2}\text{B}]$ by \cite{fan2020scalable} based on ICNNs resolve this issue, see \citep[Tables 1-3]{korotin2021continuous}. 
However, despite the growing popularity of ICNNs in OT applications \citep{makkuva2019optimal,korotin2019wasserstein,mokrov2021large},  they could be suboptimal architectures according to a recent study \citep{korotin2021neural}. According to the authors, more expressive networks without the convexity constraint outperform ICNNs in practical OT problems.
% \lingxiao{mention some conclusion from \citet{korotin2021neural}. What is more successful than ICNN? You can foreshadow what is to come. This should be the motivation of the current work}\alex{I think this might confuse the reader. The benchmark paper is about the solvers, not the barycenters. There are no algorithms for continuous barycenters without entropy/icnnS}\lingxiao{I still think we could just add to the last sentence ".. according to a recent study XX where more expressive networks without convexity constraint outperform ICNNs."}

Furthermore, \textit{evaluation} of barycenter algorithms is challenging due to the limited number of continuous measures with explicitly known barycenters. It can be computed when the input measures are location-scatter (e.g. Gaussians) \citep[\wasyparagraph 4]{alvarez2016fixed} or $1$-dimensional \citep[\wasyparagraph 2.3]{bonneel2015sliced}. Recent works \citep{li2020continuous,korotin2021continuous,fan2020scalable} consider the Gaussian case in dimensions $\leq 256$ for quantitative evaluation. In higher dimensions, the computation of the ground truth barycenter is hard even for the Gaussian case: it involves matrix inversion and square root extraction \citep[Algorithm 1]{altschuler2021averaging} with the cubic complexity in the dimension.
% of the ambient space.
% Thus, quantitative evaluation in high dimensions is challenging even in the Gaussian case.

\paragraph{Contributions.} 
\begin{itemize}
    \item We develop a novel \textit{iterative algorithm} (\wasyparagraph\ref{sec-algorithm}) for estimating Wasserstein-2 barycenters based on the fixed point approach by \citep{alvarez2016fixed} combined with a neural solver for optimal transport \citep{korotin2021neural}. Unlike predecessors, our algorithm does not introduce bias and allows arbitrary network architectures. %\lingxiao{mention that we do not need to be limited to convex networks, and in fact any OT solvers can be used.}\alex{(1) But doesn't the phrase "allows using arbitrary neural network architectures" already mean that we are not limited to ICNNs? (2) I think it is better to stick to a particular OT solver for simplicity since the algorithm uses it. I initially tried to write the text like "the method can use any solver", but it became a total mess. Also, people will ask why don't you test the other solvers?}    % \lingxiao{Should probably be more explicit that we are extending Alvarez et al. to the neural setting, and what is new. The novelty is not clear to me. Also should mention whether the method works only for W2 or not}\alex{edited}
    % As a result, it is capable to recover the barycenter in dimensions not handled by existing algorithms.
    \item We construct the \textit{Ave, celeba!} (\textit{ave}raging \textit{celeb}rity faces, \wasyparagraph\ref{sec-ave-celeba}) dataset consisting of $64\times 64$ RGB images for large-scale quantitative evaluation of continuous Wasserstein-2 barycenter algorithms. The dataset includes 3 subsets of degraded images of faces (Figure \ref{fig:ave-celeba-teaser}). The barycenter of these subsets corresponds to the original clean faces.
    % \lingxiao{mention some features of this dataset here}\alex{What is better to emphasize here?}
\end{itemize}

Our algorithm is suitable for large-scale Wasserstein-2 barycenters applications. The developed dataset will allow quantitative evaluation of barycenter algorithms at a large scale improving transparency and allowing healthy competition in the optimal transport research.

\textbf{Notation.} 
We work in a Euclidean space $\mathbb{R}^D$ for some $D$.
All the integrals are computed over $\mathbb{R}^{D}$ unless stated otherwise. We denote the set of all Borel probability measures on $\mathbb{R}^{D}$ with finite second moment by $\mathcal{P}_{2}(\mathbb{R}^{D})$. We use $\mathcal{P}_{2,\text{ac}}(\mathbb{R}^{D})\subset \mathcal{P}_{2}(\mathbb{R}^{D})$ to denote the subset of absolutely continuous measures. We denote its subset of measures with positive density by $\mathcal{P}_{2,\text{ac}}^{+}(\mathbb{R}^{D})\subset \mathcal{P}_{2,\text{ac}}(\mathbb{R}^{D})$. We denote the set of probability measures on $\mathbb{R}^{D} \times \mathbb{R}^{D}$ with marginals $\mathbb{P}$ and $\mathbb{Q}$ by $\Pi(\mathbb{P},\mathbb{Q})$.
For a measurable map $T:  \mathbb{R}^{D}\rightarrow\mathbb{R}^{D}$, we denote the associated push-forward operator by $T\sharp$. For $\phi : \mathbb{R}^{D} \rightarrow \mathbb{R}$, we denote by $\overline{\phi}$ its Legendre-Fenchel transform \citep{fenchel1949conjugate} defined by $\overline{\phi} (y) = \max_{x\in\mathbb{R}^{D}} [\langle x,y \rangle - \phi(x)]$. Recall that $\overline{\phi} $ is a convex function, even when $\phi$ is not.

\section{Preliminaries}
\label{sec-preliminaries}

\textbf{Wasserstein-2 distance}. For ${\mathbb{P},\mathbb{Q}\!\in\!\mathcal{P}_{2}(\mathbb{R}^{D})}$, Monge's \emph{primal} formulation of the squared Wasserstein-2
distance, i.e., OT with \textit{quadratic cost}, is 
\begin{equation}
\mathbb{W}_{2}^{2}(\mathbb{P},\mathbb{Q})\stackrel{\text{def}}{=}\min_{T\sharp\mathbb{P}=\mathbb{Q}}\ \int \frac{1}{2}\|x-T(x)\|^{2}d\mathbb{P}(x),
\label{ot-primal-form-monge}
\end{equation}
where the minimum is taken over measurable functions (transport maps) $T:\mathbb{R}^{D}\rightarrow\mathbb{R}^{D}$ mapping $\mathbb{P}$ to $\mathbb{Q}$. 
The optimal $T^{*}$ is called the \textit{optimal transport map}. Note that \eqref{ot-primal-form-monge} is not symmetric, and this formulation does not allow mass splitting. That is, for some
${\mathbb{P},\mathbb{Q}\in\mathcal{P}_{2}(\mathbb{R}^{D})}$, there might be no map $T$ that satisfies $T\sharp\mathbb{P}=\mathbb{Q}$. Thus, \cite{kantorovitch1958translocation} proposed the following relaxation:
\begin{equation}\mathbb{W}_{2}^{2}(\mathbb{P},\mathbb{Q})\stackrel{\text{def}}{=}\min_{\pi\in\Pi(\mathbb{P},\mathbb{Q})}\int_{\mathbb{R}^{D}\times \mathbb{R}^{D}}\frac{1}{2}\|x-y\|^{2}d\pi(x,y),
\label{ot-primal-form}
\end{equation}
where the minimum is taken over all transport plans $\pi$, i.e., measures on $\mathbb{R}^{D}\times\mathbb{R}^{D}$ whose marginals are $\mathbb{P}$ and $\mathbb{Q}$. The optimal $\pi^{*}\in\Pi(\mathbb{P},\mathbb{Q})$ is called the \textit{optimal transport plan}. 
If $\pi^{*}$ is of the form $[\text{id} , T^{*}]\sharp \mathbb{P}\in\Pi(\mathbb{P},\mathbb{Q})$ for some $T^*$, then
$T^{*}$ minimizes \eqref{ot-primal-form-monge}.
The \textit{dual form} \citep{villani2003topics} of $\mathbb{W}_{2}^{2}$ is:
\begin{eqnarray}\mathbb{W}_{2}^{2}(\mathbb{P}, \mathbb{Q})\!=\!
\max_{u\oplus v\leq \frac{\|\cdot\|^{2}}{2}}\bigg[\int u(x)d\mathbb{P}(x)+\int v(y)d\mathbb{Q}(y)\bigg],
\label{ot-dual-form}
\end{eqnarray}
where the maximum is taken over ${u\in \mathcal{L}^{1}(\mathbb{P})}$, ${v\in \mathcal{L}^{1}(\mathbb{Q})}$ satisfying ${u(x)+v(y)\!\leq\!\frac{1}{2}\|x-y\|^{2}}$
for all ${x,y\!\in\!\mathbb{R}^{D}}$. The functions $u$ and $v$ are called \textit{potentials}. There exist optimal $u^{*}, v^{*}$ satisfying $u^{*}=(v^{*})^{c}$, where ${f^{c}(y)\stackrel{def}{=}\min\limits_{x\in\mathbb{R}^{D}}\big[\frac{1}{2}\|x-y\|^{2}-f(x)\big]}$ is the $c$-transform of $f$. We rewrite \eqref{ot-dual-form} as
\begin{eqnarray}\mathbb{W}_{2}^{2}(\mathbb{P}, \mathbb{Q})=
\max_{v}\bigg[\int v^{c}(x)d\mathbb{P}(x)+\int v(y)d\mathbb{Q}(y)\bigg],
\label{ot-dual-form-c}
\end{eqnarray}
where the maximum is taken over all $v\in \mathcal{L}^{1}(\mathbb{Q})$. It is customary \citep[Cases 5.3 \& 5.17]{villani2008optimal} to define ${u(x)\!=\!\frac{1}{2}\|x\|^{2}\!-\!\psi(x)}$ and ${v(y)\!=\!\frac{1}{2}\|y\|^{2}\!-\!\phi(x)}$. 
% \lingxiao{Trick? I think it's better to define $\psi$ and $\phi$ here, and put ``it is customary to define ...''}
There exist convex optimal $\psi^{*}$ and $\phi^{*}$ satisfying $\overline{\psi^{*}}=\phi^{*}$ and $\overline{\phi^{*}}=\psi^{*}$. If $\mathbb{P}\in\mathcal{P}_{2,ac}(\mathbb{R}^{D})$, then the optimal $T^{*}$ of \eqref{ot-primal-form-monge} always exists and can be recovered from the dual solution $u^{*}$ (or $\psi^{*}$) of \eqref{ot-dual-form}: $T^{*}(x)=x-\nabla u^{*}(x)=\nabla\psi^{*}(x)$ \citep[Theorem 1.17]{santambrogio2015optimal}. The map $T^{*}$ is a gradient of a convex function, see the Brenier Theorem \citep{brenier1991polar}.

\textbf{Wasserstein-2 barycenter}. Let $\mathbb{P}_{1},\dots,\mathbb{P}_{N}\in \mathcal{P}_{2,ac}(\mathbb{R}^{D})$ such that at least one of them has bounded density.
Their barycenter w.r.t. weights $\alpha_{1},\dots,\alpha_{N}$ ($\alpha_{n}\!>\!0$; ${\sum_{n=1}^{N}\alpha_{n}=1}$) is given by \citep{agueh2011barycenters}: 
\begin{equation}\overline{\mathbb{P}}\stackrel{\text{def}}{=}\argmin_{\mathbb{P}\in\mathcal{P}_{2}(\mathbb{R}^{D})}\sum_{n=1}^{N}\alpha_{n}\mathbb{W}_{2}^{2}(\mathbb{P}_{n},\mathbb{P}).
\label{w2-barycenter-def}
\end{equation}
The barycenter $\overline{\mathbb{P}}$ exists uniquely and $\overline{\mathbb{P}}\in\mathcal{P}_{2,ac}(\mathbb{R}^{D})$. Moreover, its density is bounded \citep[Definition 3.6 \& Theorem 5.1]{agueh2011barycenters}.
For $n\in\{1,2,\dots,N\}$, let $T_{\overline{\mathbb{P}}\rightarrow\mathbb{P}_{n}}=\nabla\psi_{n}^{*}$ be the OT maps from $\overline{\mathbb{P}}$ to $\mathbb{P}_{n}$.
The following holds $\overline{\mathbb{P}}$-almost everywhere:
\begin{equation}\sum_{n=1}^{N}\alpha_{n}T_{\overline{\mathbb{P}}\rightarrow\mathbb{P}_{n}}(x)=\sum_{n=1}^{N}\alpha_{n}\nabla\psi^{*}_{n}(x)=x,
\label{bar-condition}
\end{equation}
see \citep[\wasyparagraph 3]{alvarez2016fixed}. If $\overline{\mathbb{P}}\in\mathcal{P}_{2,\text{ac}}^{+}(\mathbb{R}^{D})$, then \eqref{bar-condition} holds for every $x\in\mathbb{R}^{D}$, i.e., $\sum_{n=1}^{N}\alpha_{n}\psi_{n}^{*}(x)=\frac{\|x\|^{2}}{2}+c$. We call such convex potentials $\psi_{n}^{*}$ \textit{congruent}.

\vspace{-2mm}\section{Related Work}

% Methods for OT problems are of two types: (semi-)\textit{discrete} \citep{peyre2019computational} and \textit{continuous}. \lingxiao{semidiscrete can also be regarded as neither continuous or discrete} Applying discrete methods to the continuous measures requires discretization. Computationally feasible discretization underscore continuous methods, e.g., see \citep[Tables 1-3]{korotin2021continuous} for comparisons in barycenter problems. \lingxiao{Underscore? I do not understand previous sentence.} Continuous OT algorithms employ neural networks or kernel expansions to estimate transport maps or dual solutions. This scales OT to higher-dimensional tasks not handled by discrete methods.\lingxiao{Not sure if this paragraph is adding anything. Consider combining with the last bit of intro.}\alex{True. May be we just remove this paragraph? In principle, the part in the intro already seems sufficient?}\lingxiao{I agree. Let's just remove this paragraph.}

\vspace{-2mm}Below we review existing continuous methods for OT. In \wasyparagraph\ref{sec-solvers}, we discuss methods for OT problems \eqref{ot-primal-form-monge}, \eqref{ot-primal-form}, \eqref{ot-dual-form}. In \wasyparagraph\ref{sec-solvers-bar}, we review algorithms that compute barycenters \eqref{w2-barycenter-def}.

\vspace{-1mm}\subsection{Continuous OT Solvers for $\mathbb{W}_{2}$}
\label{sec-solvers}
% \lingxiao{W2 transport seems ambiguous. W2 distance?}\alex{Actually, we are more interested in transport map/OT gradient rather than the distance}\lingxiao{It's just I never see the phrase "W2 transport". Maybe "Continuous OT Solvers for W2"}
We use the phrase \textit{OT solver} to denote any method capable of recovering $T^{*}$ or $u^{*}$ (or $\psi^{*}$).

\textbf{Primal-form} solvers based on \eqref{ot-primal-form-monge} or \eqref{ot-primal-form}, e.g., \cite{xie2019scalable,lu2020large}, parameterize $T^*$ using complicated generative modeling techniques with adversarial losses to handle the pushforward constraint $T\sharp\mathbb{P}=\mathbb{Q}$ in the primal form \eqref{ot-primal-form-monge}. They depend on careful hyperparameter search and complex optimization \citep{lucic2018gans}.
% \lingxiao{I guess the challenge is you need to match $Q$? You are also using generative modeling techniques here.}\alex{Truee, nut the generative model is used for the barycenter, not for OT solver:)}\lingxiao{Then maybe add something like "... using complicated generative modeling techniques to handle the pushforward constraint in the primal form."}

\textbf{Dual-form} continuous solvers \citep{genevay2016stochastic,seguy2017large,nhan2019threeplayer,taghvaei20192,korotin2019wasserstein} based on \eqref{ot-dual-form} or \eqref{ot-dual-form-c} have straightforward optimization procedures and can be adapted to various tasks without extensive hyperparameter search.
% \lingxiao{Should probably mention why dual-form is easier, e.g. the constraint is easier to handle. Right now these two paragraphs basically say primal is hard because it's hard, and dual is easier because it's straightforward.}\alex{I fixed the primal form paragraph, but it is not clear how to fix the dual form paragraph. Not all solvers are "unconstrained" -- recall ICNN ones}

A comprehensive overview and a benchmark of dual-form solvers are given in \cite{korotin2021neural}. According to the evaluation, the best performing OT solver is \textit{reversed maximin solver} $\lfloor \text{MM:R}\rceil$, a modification of the idea proposed by \cite{nhan2019threeplayer} in the context of Wasserstein-1 GANs \citep{arjovsky2017wasserstein}.
% Other solvers work worse due to bias, gradient deviation or poor performance of ICNNs.
In this paper, we employ this solver as a part of our algorithm. We review it below.

\textit{Reversed Maximin Solver}.
In \eqref{ot-dual-form-c}, $v^{c}(x)$ can be expanded through $v$ via the definition of $c$-transform:
\begin{eqnarray}\max_{v}\!\int\!\min_{y\in\mathbb{R}^{D}}\big[\frac{\|x-y\|^{2}_{2}}{2}-v(y)\big]d\mathbb{P}(x)\!+\!\int v(y)d\mathbb{Q}(y)\!=
% \label{semi-dual-opt}
\nonumber
\\
% \max_{v}\int\min_{T}\big[\frac{1}{2}\|x-T(x)\|^{2}_{2}-v\big(T(x)\big)\big]d\mathbb{P}(x)+\int v(y)d\mathbb{Q}(y)=
% \nonumber
% \\
\hspace{-3mm}\max_{v}\min_{T}\!\int\hspace{-1.5mm}\bigg[\frac{\|x\!-\!T(x)\|^{2}_{2}}{2}\!-\!v\big(T(x)\big)\!\bigg]\!d\mathbb{P}(x)\!+\hspace{-1.5mm}\int \hspace{-1mm}v(y)d\mathbb{Q}(y).\hspace{-2mm}
% \nonumber
\label{semi-dual-T}
\end{eqnarray}
% In transition between \eqref{semi-dual-opt} and \eqref{semi-dual-T}, 
% \lingxiao{The equation number is way out of the boundary.}\alex{True, but let us not fix this yet -- not enough space}
In \eqref{semi-dual-T}, the optimization over ${y\in \mathbb{R}^{D}}$ is replaced by the equivalent optimization over functions ${T:\mathbb{R}^{D}\rightarrow\mathbb{R}^{D}}$. This is done by the interchanging of integral and minimum, see \citep[Theorem 3A]{rockafellar1976integral}.
% \lingxiao{Isn't this simply because $T$ needs to be optimal pointwise?}\alex{True, but you point is a little bit informal -- min may not exist (should be replaced by inf actually). Therefore, this theorem is a formal answer to any reviewer's worries.}\lingxiao{But if $T$ is just a function from set to set, then I don't see why the min should not exist, if it exists for each $x$. Do you mean $T$ needs to have additional regularity conditions?}\alex{I suppose it also has to be, for example, measurable. This property does not directly follow from your proof. Therefore, it is better to rely  on a stong explanatory theorem here which handles all these tricky questions which we may miss.}

The key point of this reformulation is that the optimal solution of this maximin problem is given by $(v^{*},T^{*})$, where $T^{*}$ is the OT map from $\mathbb{P}$ to $\mathbb{Q}$, see discussion in \citep[\wasyparagraph 2]{korotin2021neural} or  \citep[\wasyparagraph 4.1]{rout2021generative}. 
In practice, the potential $v:\mathbb{R}^{D}\rightarrow\mathbb{R}$ and the map ${T:\mathbb{R}^{D}\rightarrow\mathbb{R}^{D}}$ are parametrized by neural networks $v_{\omega},T_{\theta}$. To train $\theta$ and $\omega$, stochastic gradient ascent/descent (SGAD) over mini-batches from $\mathbb{P},\mathbb{Q}$ is used.

% \textit{Variants.} If the parametrization is ${v_{\omega}(y)=\frac{\|y\|^{2}}{2}-\psi_{\omega}(y)}$ and ${T_{\theta}=\nabla\psi_{\theta}}$ with ICNNs $\psi_{\omega},\phi_{\theta}$, then $\lceil \text{MM:R}\rfloor$ becomes the solver by \cite{makkuva2019optimal}. One may add the cycle-consistency regularizer \citep{korotin2019wasserstein} to remove maximinity of the problem. Following the benchmark by \cite{korotin2021neural}, these ICNN-based solvers exhibit similar performance. We denote them by $\lceil \text{W2}\rfloor$.

\vspace{-1mm}\subsection{Algorithms for Continuous $\mathbb{W}_{2}$ Barycenters}
\label{sec-solvers-bar}

\textbf{Variational optimization.} Problem \eqref{w2-barycenter-def} is optimization over probability measures. 
% This can be done using the generic algorithm by \cite{cohen2020estimating} who employ generative networks to compute barycenters w.r.t.\ arbitrary discrepancies.
% They test their method with the maximum mean discrepancy and Sinkhorn divergence.
% Applying it to $\mathbb W_2$ barycenters requires estimation of $\mathbb{W}_{2}^{2}$. \cite{fan2020scalable} test this approach using the OT solver by \cite{makkuva2019optimal} based on ICNNs to compute $\mathbb{W}_{2}^{2}$, yielding $[\text{SC}\mathbb{W}_{2}\text{B}]$ algorithm.
% We further discuss the generic variational approach in \wasyparagraph\ref{sec-generative-connection}.
To estimate $\overline{\mathbb{P}}$, one may employ a generator $G_{\xi}:\mathbb{R}^{H}\rightarrow\mathbb{R}^{D}$ with a latent measure $\mathbb{S}$ on $\mathbb{R}^{H}$ and train $\xi$  by minimizing
\begin{equation}\sum_{n=1}^{N}\alpha_{n}\mathbb{W}_{2}^{2}(\underbrace{G_{\xi}\sharp\mathbb{S}}_{\mathbb{P}_{\xi}},\mathbb{P}_{n})\rightarrow\min_{\xi}\!.
\label{bar-gen-opt}
\end{equation}
Optimization \eqref{bar-gen-opt} can be performed by using SGD on random mini-batches from measures  $\mathbb{P}_{n}$ and $\mathbb{S}$. The difference between possible variational algorithms lies in the particular estimation method for $\mathbb{W}_{2}^{2}$ terms.
% \lingxiao{Not sure what the previouse sentence means}\alex{To optimize this objective, we need to estimate W2 terms somehow, i.e. use some OT solver}
To our knowledge, only ICNN-based minimax solver \citep{makkuva2019optimal} has been used to compute $\mathbb{W}_{2}^{2}$ in \eqref{bar-gen-opt} yielding $[\text{SC}\mathbb{W}_{2}\text{B}]$ algorithm \citep{fan2020scalable}.

\textbf{Potential-based optimization.} \cite{li2020continuous,korotin2021continuous} recover the optimal potentials $\{\psi^{*}_{n},\phi^{*}_{n}\}$ for each pair $(\overline{\mathbb{P}},\mathbb{P}_{n})$ via a non-minimax regularized dual formulation. No generative model is needed: the barycenter is recovered by pushing forward measures using gradients of potentials or by barycentric projection. However, the non-trivial choice of the \textit{prior} barycenter distribution is required. Algorithm $[\text{CR}\mathbb{W}\text{B}]$ by \cite{li2020continuous} use entropic or quadratic regularization and $[\text{C}\mathbb{W}_{2}\text{B}]$ algorithm by \cite{korotin2021continuous} uses ICNNs, congruence and cycle-consistency 
\citep{korotin2019wasserstein} regularization.

\textbf{Other methods.} Recent work \cite{chi2021variational} combines the variational \eqref{bar-gen-opt} and potential-based optimization via the $c$-cyclical monotonity regularization. In \cite{daaloul2021sampling}, an algorithm to sample from the continuous Wasserstein barycenter via the gradient flows is proposed.

\vspace{-2mm}\section{Iterative $\mathbb{W}_{2}$-Barycenter Algorithm}
\label{sec-algorithm}

\vspace{-2mm}Our proposed algorithm is based on the \textit{fixed point approach} by \cite{alvarez2016fixed} which we recall in \wasyparagraph\ref{sec-fixed-point}. In \wasyparagraph\ref{sec-practical-algorithm}, we formulate our algorithm for computing Wasserstein-2 barycenters. In \wasyparagraph\ref{sec-generative-connection}, we show that our algorithm generalizes the variational barycenter approach.

\vspace{-1mm}\subsection{Theoretical Fixed Point Approach}
\label{sec-fixed-point}

Following \cite{alvarez2016fixed}, we define an operator $\mathcal{H}:\mathcal{P}_{2,ac}(\mathbb{R}^{D})\!\rightarrow\!\mathcal{P}_{2,ac}(\mathbb{R}^{D})$ by
${\mathcal{H}(\mathbb{P})=[\sum_{n=1}^{N}\alpha_{n}T_{\mathbb{P}\rightarrow\mathbb{P}_{n}}]\sharp \mathbb{P},}$
where $T_{\mathbb{P}\rightarrow\mathbb{P}_{n}}$ denotes the OT map from $\mathbb{P}$ to $\mathbb{P}_{n}$. The measure $\mathcal{H}(\mathbb{P})$ obtained by the operator is indeed absolutely continuous, see \citep[Theorem 3.1]{alvarez2016fixed}. According to \eqref{bar-condition}, the barycenter $\overline{\mathbb{P}}$ defined by \eqref{w2-barycenter-def} is a \textit{fixed point} of $\mathcal{H}$, i.e., $\mathcal{H}(\overline{\mathbb{P}})=\overline{\mathbb{P}}$.
This suggests a way to compute $\overline{\mathbb{P}}$ by picking some ${\mathbb{P}\in\mathcal{P}_{2,ac}(\mathbb{R}^{D})}$ and recursively applying $\mathcal{H}$ until convergence.
However, there are several \textbf{challenges}:
\begin{enumerate}[label=(\alph*), wide, labelindent=0pt]
\item[\bf (a)] \label{challenge_a}
A fixed point $\mathbb{P}\in\mathcal{P}_{2,\text{ac}}(\mathbb{R}^{D})$ satisfying $\mathcal{H}(\mathbb{P})=\mathbb{P}$ may be not the barycenter
\citep[Example 3.1]{alvarez2016fixed}. The situation is analogous to that of the iterative $k$-means algorithm for a different problem -- clustering. There may be fixed points which are not globally optimal.
\item[\bf (b)] \label{challenge_b}
The sequence $\{\mathcal{H}^{k}(\mathbb{P})\}_k$ is tight \citep[Theorem 3.6]{alvarez2016fixed} so it has a subsequence converging in $\mathcal{P}_{2,ac}(\mathbb{R}^{D})$, but the entire sequence may not converge. Nevertheless, the value of the objective \eqref{w2-barycenter-def} decreases for $\mathcal{H}^k(\mathbb{P})$ as $k\rightarrow\infty$ \citep[Prop. 3.3]{alvarez2016fixed}.
\item[\bf (c)] \label{challenge_c}
Efficient parametrization of the evolving measure $\mathcal{H}^{k}(\mathbb{P})$ is required. Moreover, to get $\mathcal{H}^{k+1}(\mathbb{P})$ from $\mathcal{H}^{k}(\mathbb{P})$, one needs to compute $N$ optimal transport maps $T_{\mathcal{H}^{k}(\mathbb{P})\rightarrow\mathbb{P}_{n}}$ which can be costly.
\end{enumerate}
In \cite{chewi2020gradient} and \cite{altschuler2021averaging}, the fixed point approach is considered in the Gaussian case where the sequence $\mathcal{H}^{k}(\mathbb{P})$ is guaranteed to converge to the unique fixed point -- the barycenter.
The Gaussian case also makes parameterization \textbf{(c)} simple since both measures $\mathbb{P}_{n}$ and $\mathcal{H}^{k}(\mathbb{P})$ can be parametrized by means and covariance matrices, and the maps $T_{\mathcal{H}^{k}(\mathbb{P})\rightarrow\mathbb{P}_{n}}$ are linear with closed form.
% Their approach has a dimension-free convergence rate.

For general continuous measures $\mathbb{P}_{n}$, it remains an open problem to find sharp conditions on inputs $\mathbb{P}_n$ and the initial measure $\mathbb{P}$ of the fixed-point iteration for the sequence $\{\mathcal{H}^k(\mathbb{P})\}_k$ to converge to the barycenter. In this work, we empirically verify that the fixed point approach works well for the input measures that we consider and for a randomly initialized generative model representing the evolving barycenter (\wasyparagraph\ref{sec-practical-algorithm}).
We tackle challenge \textbf{(c)} and develop a scalable optimization procedure that requires only sample access to ${\mathbb{P}_{n}\!\in\!\mathcal{P}_{2,ac}(\mathbb{R}^{D})}$.

\subsection{Practical Iterative Optimization Procedure}
\label{sec-practical-algorithm}

We employ a generative model to parametrize the evolving measure, i.e., put ${\mathbb{P}_{\xi}=G_{\xi}\sharp\mathbb{S}}$, where $\mathbb{S}$ is a latent measure, e.g., $\mathbb{S}=\mathcal{N}(0,I_{H})$, and $G_{\xi}$ is a neural network $\mathbb{R}^{H}\rightarrow\mathbb{R}^{D}$ with parameters $\xi$. Our approach to compute the operator $\mathcal{H}(\mathbb{P}_{\xi})$ and update $G_{\xi}$ consists of two steps. 

First, we approximately recover $N$ maps
$T_{\mathbb{P}_{\xi}\rightarrow\mathbb{P}_{n}}$
via $\lceil \text{MM:R}\rfloor$ solver, i.e., we use $N$ pairs or networks $\{T_{\theta_{n}}, v_{\omega_{n}}\}$ and train them by optimizing \eqref{semi-dual-T} with ${\mathbb{P}\leftarrow \mathbb{P}_{\xi}}$ and ${\mathbb{Q}\leftarrow\mathbb{P}_{n}}$. For each $n=1,2,\dots,N$, we perform SGAD by using batches from $G_{\xi}\sharp\mathbb{S}$ and $\mathbb{P}_{n}$ and get $T_{\theta_{n}}\approx T_{\mathbb{P}_{\xi}\rightarrow\mathbb{P}^{n}}$.

Second, we update $G_{\xi}$ to represent $\mathcal{H}(\mathbb{P}_{\xi})$ instead of $\mathbb{P}_{\xi}$. Inspired by \cite{chen2019gradual}, we do this via regression. We introduce $G_{\xi_0}$, a fixed copy of $G_{\xi}$. Next, we regress $G_{\xi}(\cdot)$ onto $\sum_{n=1}^{N}\alpha_{n} T_{\theta_{n}}\big( G_{\xi_0}(\cdot)\big)$
$$\int_{z}\boldsymbol\ell\bigg(G_{\xi}(z),\sum_{n=1}^{N}\alpha_{n}T_{\theta_{n}}\big(G_{\xi_0}(z)\big)\bigg)d\mathbb{S}(z)\rightarrow\min_{\xi}$$
by performing SGD on random batches from $\mathbb{S}$, e.g., by using \textit{squared error} ${\boldsymbol\ell(x,x')\!\stackrel{\text{def}}{=}\!\frac{1}{2}\|x-x'\|^{2}}$. Thus, generator $G_{\xi}(\cdot)$ becomes close to ${\sum_{n=1}^{N}\!\alpha_{n} T_{\theta_{n}}\big(G_{\xi_0}(\cdot)\big)}$ as a function $\mathbb{R}^{H}\!\rightarrow\!\mathbb{R}^{D}$. We get
\begin{eqnarray}\mathbb{P}_{\xi}=G_{\xi}\sharp\mathbb{S}\approx \big[\sum_{n=1}^{N}\alpha_{n} T_{\theta_{n}}\big]\sharp \big[G_{\xi_0}\sharp\mathbb{S}\big]=
\big[\sum_{n=1}^{N}\alpha_{n} T_{\theta_{n}}\big]\sharp \mathbb{P}_{\xi_0}\approx\big[\sum_{n=1}^{N}\alpha_{n} T_{\mathbb{P}_{\xi_0}\rightarrow\mathbb{P}_{n}}\big]\sharp \mathbb{P}_{\xi_0}= \mathcal{H}(\mathbb{P}_{\xi_0}),
\nonumber
\end{eqnarray}
i.e., the new generated $G_{\xi}\sharp\mathbb{S}$ measure approximates $\mathcal{H}(\mathbb{P}_{\xi_0})$.

\begin{figure*}[!t]
\begin{subfigure}[b]{0.49\linewidth}
\centering
\includegraphics[width=0.9\linewidth]{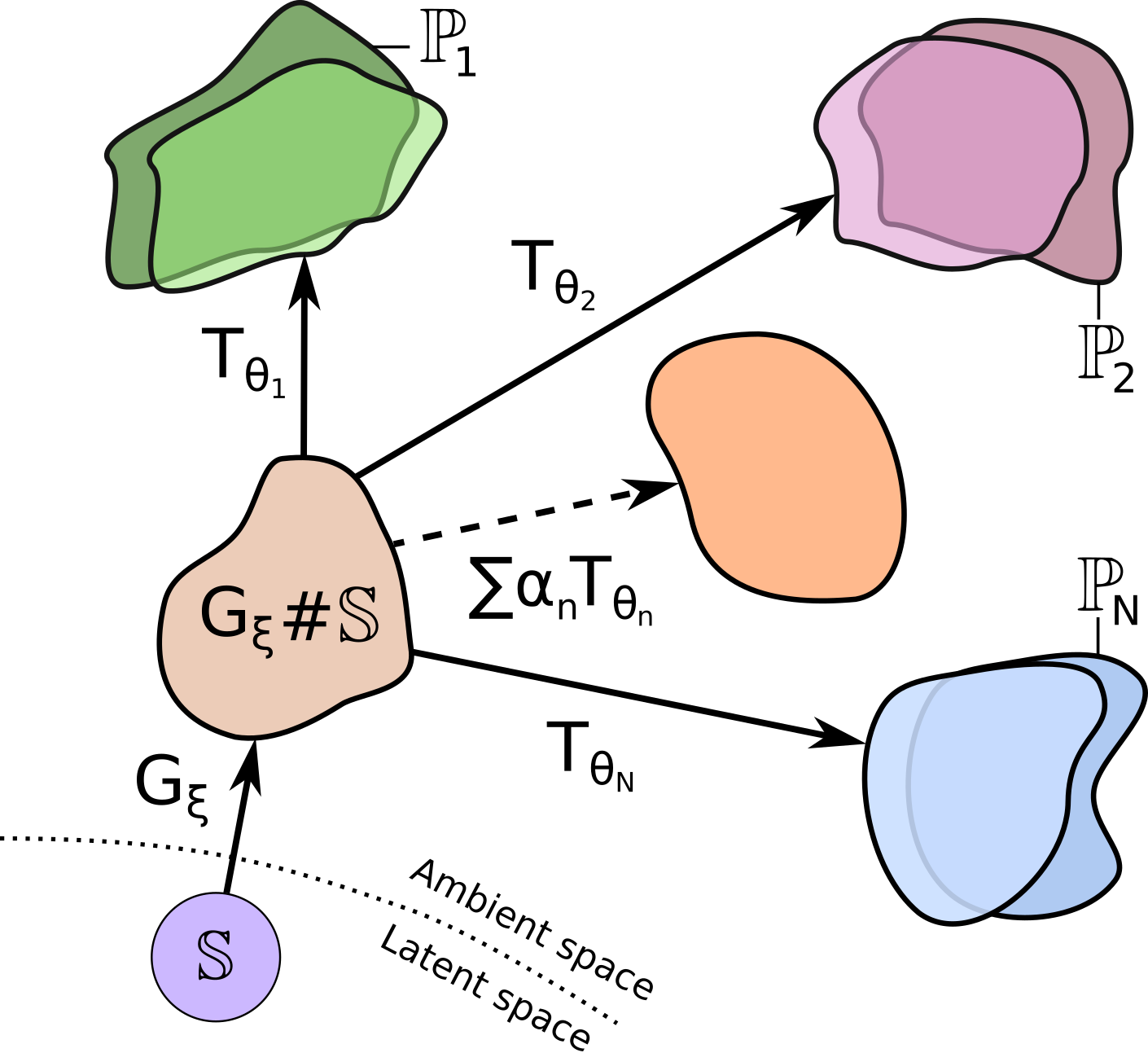}
\vspace{1mm}\caption{
\centering \textbf{Step 1.} We compute $N$ approximate OT maps $T_{\theta_{n}}$ from generated measure $\mathbb{P}_{\xi}=G_{\xi}\sharp\mathbb{S}$ \protect\linebreak to input measures $\mathbb{P}_{n}$.}
\label{fig:win-step-1}
\end{subfigure}\hspace{1mm}\vrule\hspace{1mm}
\begin{subfigure}[b]{0.49\linewidth}
\centering
\includegraphics[width=0.9\linewidth]{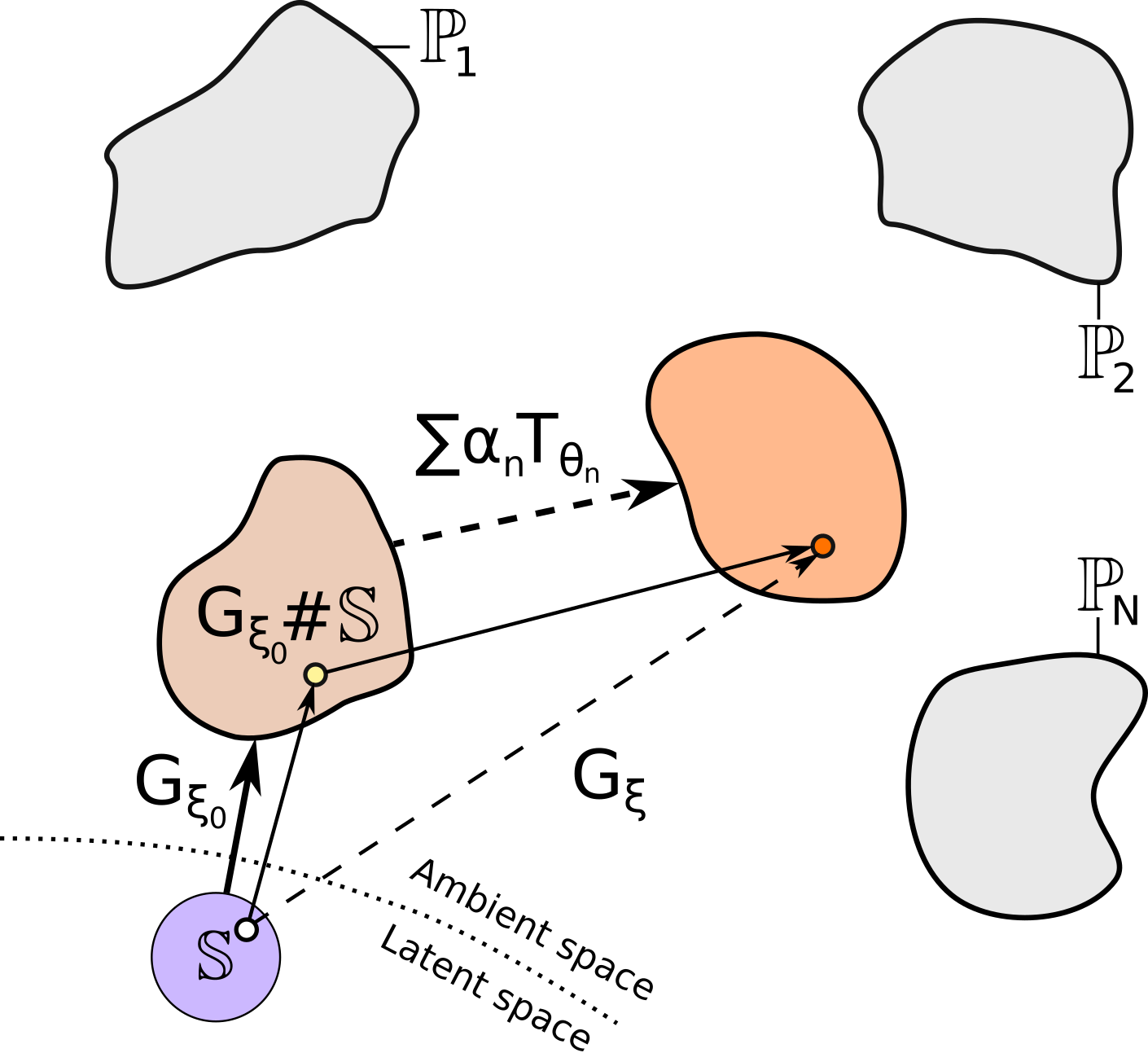}
\caption{
\centering \textbf{Step 2.} We regress $G_{\xi}(\cdot)$ onto $\sum_{n=1}^{N}\alpha_{n} T_{\theta_{n}}\big( G_{\xi_0}(\cdot)\big)$, where $G_{\xi_0}$\protect\linebreak is a fixed copy of $G_{\xi}$ before the update.}
\label{fig:win-step-2}
\end{subfigure}
\vspace{-4mm}
\caption{\centering Our proposed two-step implementation of the fixed-point operator $\mathcal{H}(\cdot)$..}
\label{fig:win-pipeline}
\vspace{-4mm}
%  that we use to compute the barycenter
\end{figure*}

\textbf{Summary.} Our two-step approach iteratively recomputes OT maps $T_{\mathbb{P}_{\xi}\rightarrow\mathbb{P}_{n}}$ (Figure \ref{fig:win-step-1}) and then uses regression to update the generator (Figure \ref{fig:win-step-2}). The \textit{optimization procedure} is detailed in Algorithm \ref{algorithm-win}. Note that when fitting OT maps $T_{\mathbb{P}_{\xi}\rightarrow\mathbb{P}^{n}}$, we start from previously used $\{T_{\theta_n},v_{\omega_n}\}$ rather than re-initialize them. Empirically, this works better.

\vspace{-2mm}\subsection{Relation to Variational  Barycenter Algorithms}
\label{sec-generative-connection}

We show that our Algorithm \ref{algorithm-win} reduces to variational approach (\wasyparagraph\ref{sec-solvers-bar}) when the number of generator updates, $K_G$, is equal to $1$. 
%\lingxiao{Generalize in what sense?
%Lemma 1 seems to be showing equivalence not generalization} %\alex{Variational algorithm is ours only for kG=1. If kG>1, it is a different algorithm}
More specifically, we show the equivalence of the gradient update w.r.t. parameters $\xi$ of the \textit{generator} in our iterative Algorithm \ref{algorithm-win} and that of \eqref{bar-gen-opt}. 
We assume that $\mathbb{W}_{2}^{2}$ terms are computed exactly in \eqref{bar-gen-opt} regardless of the particular OT solver. Similarly, in Algorithm \ref{algorithm-win}, we assume that maps $G_{\xi_0}\sharp\mathbb{S}\rightarrow\mathbb{P}_{n}$ before the generator update are always exact, i.e., $T_{\theta_{n}}=T_{\mathbb{P}_{\xi_0}\rightarrow\mathbb{P}_{n}}$.

\begin{lemma}
\label{lemma-equivalence}
\hspace{-1mm}Assume that $\mathbb{P}_{\xi}\!=\!G_{\xi}\sharp\mathbb{S}\!\in\!\mathcal{P}_{2,ac}(\mathbb{R}^{D})$. Consider $K_G = 1$ for the iterative Algorithm \ref{algorithm-win}, i.e., we do a single gradient step regression update per OT solvers' update. Assume that $\boldsymbol\ell(x,x')=\frac{1}{2}\|x-x'\|^{2}$, i.e., the squared loss is used for regression. Then the generator's gradient update in Algorithm \ref{algorithm-win} is the same as in the variational algorithm:
\begin{eqnarray}\frac{\partial}{\partial\xi}\int_{z}\frac{1}{2}\norm[\big]{G_{\xi}(z)\!-\!\sum_{n=1}^{N}\alpha_{n} T_{\mathbb{P}_{\xi_0}\rightarrow\mathbb{P}_{n}}\big(G_{\xi_0}(z)\big)}^{2}d\mathbb{S}(z)=
\frac{\partial}{\partial\xi}\sum_{n=1}^{N}\alpha_{n}\mathbb{W}_{2}^{2}(G_{\xi}\sharp\mathbb{S},\mathbb{P}_{n}),
\label{equivalence-grads}
\end{eqnarray}
where the derivatives are evaluated at $\xi=\xi_0$.
% , i.e. the current generator parameters state.
\end{lemma}

\vspace{-2mm}We prove the lemma in Appendix \ref{sec-proofs}. 
%On the one hand, to follow the fixed point approach (\wasyparagraph\ref{sec-fixed-point}), one needs to optimize generator $G_{\xi}(\cdot)$ until it converges to $\sum_{n=1}^{N}\alpha_{n}T_{\mathbb{P}_{\xi_0}\rightarrow\mathbb{P}_{n}}\big( G_{\xi_0}(\cdot)\big)$, i.e., $K_{G}\rightarrow\infty$. On the other hand, due to our Lemma \ref{lemma-equivalence}, using $K_{G}=1$ simply leads to the variational approach \eqref{bar-gen-opt}.
In practice, we choose $K_{G} = 50$ as it empirically works better.
% more iterations of generative update make training more stable. <--- I reduced this to save space :))
% We use $K_{G}=50$ in experiments.

\begin{algorithm}[t!]
\SetInd{0.5em}{0.3em}
    {
        \SetAlgorithmName{Algorithm}{empty}{Empty}
        \SetKwInOut{Input}{Input}
        \SetKwInOut{Output}{Output}
        \Input{latent $\mathbb{S}$ and input
        $\mathbb{P}_{1},\dots,\mathbb{P}_{N}$ measures;
        weights $\alpha_{1},\dots,\alpha_{N}>0$ ($\sum_{n=1}^{N}\alpha_{n}=1$);
        number of iters per network: $K_{G}$, $K_{T}$, $K_{v}$;
        generator $G_{\xi}:\mathbb{R}^{H}\rightarrow\mathbb{R}^{D}$;\\ mapping networks $T_{\theta_{1}},\dots,T_{\theta_{N}}:\mathbb{R}^{D}\rightarrow\mathbb{R}^{D}$; potentials $v_{\omega_{1}},\dots,v_{\omega_{N}}:\mathbb{R}^{D}\rightarrow\mathbb{R}$;\\ 
        regression loss $\boldsymbol\ell:\mathbb{R}^{D}\times\mathbb{R}^{D}\rightarrow\mathbb{R}_{+}$; 
        }
        \Output{generator satisfying $G_{\xi}\sharp\mathbb{S}\approx \overline{\mathbb{P}}$; OT maps satisfying $T_{\theta_{n}}\sharp(G_{\xi}\sharp\mathbb{S})\approx \mathbb{P}_{n}$\;
        }
        
        \Repeat{not converged}{
        \# \textit{OT solvers update} \\
        \For{$n = 1,2, \dots, N$}{
                \For{$k_{v} = 1,2, \dots, K_{v}$}{
                    Sample batches $Z\!\sim\! \mathbb{S}$, $Y\!\sim\! \mathbb{P}_{n}$; $X\!\leftarrow\! G_{\xi}(Z)$\;
                    $\mathcal{L}_{v}\leftarrow \frac{1}{|X|}\sum\limits_{x\in X}v_{\omega_{n}}\big(T_{\theta_{n}}(x)\big)-\frac{1}{|Y|}\sum\limits_{y\in Y}v_{\omega_{n}}\big(y\big)$\;
                    Update $\omega_{n}$ by using $\frac{\partial \mathcal{L}_{v}}{\partial \omega_{n}}$\;
                    
                    \For{$k_{T} = 1,2, \dots, K_{T}$}{
                        Sample batch $Z\sim \mathbb{S}$; $X\leftarrow G_{\xi}(Z)$\;
                        ${\mathcal{L}_{T}\!=\!\frac{1}{|X|}\!\sum\limits_{x\in X}\!\big[\frac{1}{2}\|x\!-\!T_{\theta_{n}}(x)\|^{2}\!-\!v_{\omega_{n}}\!\big(T_{\theta_{n}}(x)\big)\!\big]}$\;
                        Update $\theta_{n}$ by using $\frac{\partial \mathcal{L}_{T}}{\partial \theta_{n}}$\;
                    }
                }
            }
        
            \# \textit{Generator update (regression)}\\ 
            $G_{\xi_0}\leftarrow \text{copy}\big(G_{\xi}\big)$\;
            \For{$k_{G} = 1,2, \dots, K_{G}$}{
                Sample batch $Z\sim \mathbb{S}$\;
                ${\mathcal{L}_{G}\!\leftarrow \!\frac{1}{|Z|}\!\sum\limits_{z\in Z}\boldsymbol\ell\bigg(G_{\xi}(z),\sum_{n=1}^{N}\alpha_{n} T_{\theta_{n}}\big(G_{\xi_0}(z)\big)\bigg)}$\;
                Update $\xi$ by using $\frac{\partial \mathcal{L}_{G}}{\partial \xi}$\;
            }
        }
        
        \caption{Wasserstein Iterative Networks (WIN) for Barycenter Estimation
        }
        \label{algorithm-win}
    }
\end{algorithm}

\vspace{-2mm}\section{Ave, celeba! Images Dataset}
\label{sec-ave-celeba}

\vspace{-2mm}In this section, we develop a generic methodology for building measures with known $\mathbb{W}_{2}$ barycenter. We then use it to construct Ave, celeba! dataset for quantitative evaluation of barycenter algorithms.

% We use CelebA $64\times64$ dataset containing $\approx 200$K aligned RGB images of faces \citep{liu2015faceattributes} to create $3$ empirical measures containing $\approx 67$K samples. Measures are produced from a random disjoint subsamples of the initial dataset by applying artificially created noises. The barycenter of $3$ noise measures w.r.t. weights $(\frac{1}{4}, \frac{1}{2}, \frac{1}{4})$ are the original clean faces. As the result, the dataset can be used for quantitative evaluation barycenter algorithms and FID \citep{heusel2017gans} can be used as the metric. Below we detail our generic methodology for producing such a dataset.

\textbf{Key idea.} Consider $\alpha_{1},\dots,\alpha_{N}>0$ with ${\sum_{n=1}^{N}\alpha_{n}=1}$, congruent convex functions ${\psi_{1},\dots\psi_{N}}:\mathbb{R}^{D}\rightarrow\mathbb{R}$, and a measure $\mathbb{P}\in\mathcal{P}_{2,\text{ac}}^{+}(\mathbb{R}^{D})$ with positive density. Define ${\mathbb{P}_{n}=\nabla\psi_{n}\sharp\mathbb{P}}$. Thanks to Brenier's theorem \cite{brenier1991polar}, $\nabla\psi_{n}$ is the unique OT map from $\mathbb{P}$ to $\mathbb{P}_{n}$. Since the support of $\mathbb{P}$ is $\mathbb{R}^{D}$, $\psi_{n}$ is the unique (up to a constant) dual potential  for $(\mathbb{P},\mathbb{P}_{n})$ \cite{staudt2022uniqueness}. Since potentials $\psi_{n}$'s are congruent, the barycenter $\overline{\mathbb{P}}$ of $\mathbb{P}_{n}$ w.r.t. weights $\alpha_{1},\dots,\alpha_{N}$ is $\mathbb{P}$ itself \citep[C.2]{chewi2020gradient}. If $\psi_{n}$'s are such that all $\mathbb{P}_{n}$ are absolutely continuous, then $\mathbb{P}=\overline{\mathbb{P}}$ is the unique barycenter (\wasyparagraph\ref{sec-preliminaries}).

If one obtains $N$ congruent $\psi_{n}$, then for any $\mathbb{P}\in\mathcal{P}_{2,ac}(\mathbb{R}^{D})$, pushforward measures ${\mathbb{P}_{n}=\nabla\psi_{n}\sharp\mathbb{P}}$ can be used as the input measures for the barycenter task. For $\mathbb{P}$ accessible by samples, measures $\mathbb{P}_{n}$ are also accessible by samples: one may sample $x\sim\mathbb{P}$ and push samples forward by $\nabla\psi_{n}$. 

The challenging part is to construct non-trivial congruent convex functions $\psi_{n}$. First, we provide a novel method to transform a single convex function $\psi$ into a \textit{pair} $(\psi^{l},\psi^{r})$ of convex functions satisfying $\alpha\nabla\psi^{l}(x)+(1-\alpha)\nabla\psi^{r}(x)=x$ for all $x\in\mathbb{R}^{D}$ (Lemma \ref{lemma-conj-pair}). Next, we extend the method to generate congruent $N$-\textit{tuples} (Lemma \ref{lemma-conj-n}). 

\begin{figure*}[!t]
\vspace{-2mm}
    \centering
    \begin{subfigure}[b]{0.47\textwidth}
         \centering
         \includegraphics[width=0.93\linewidth]{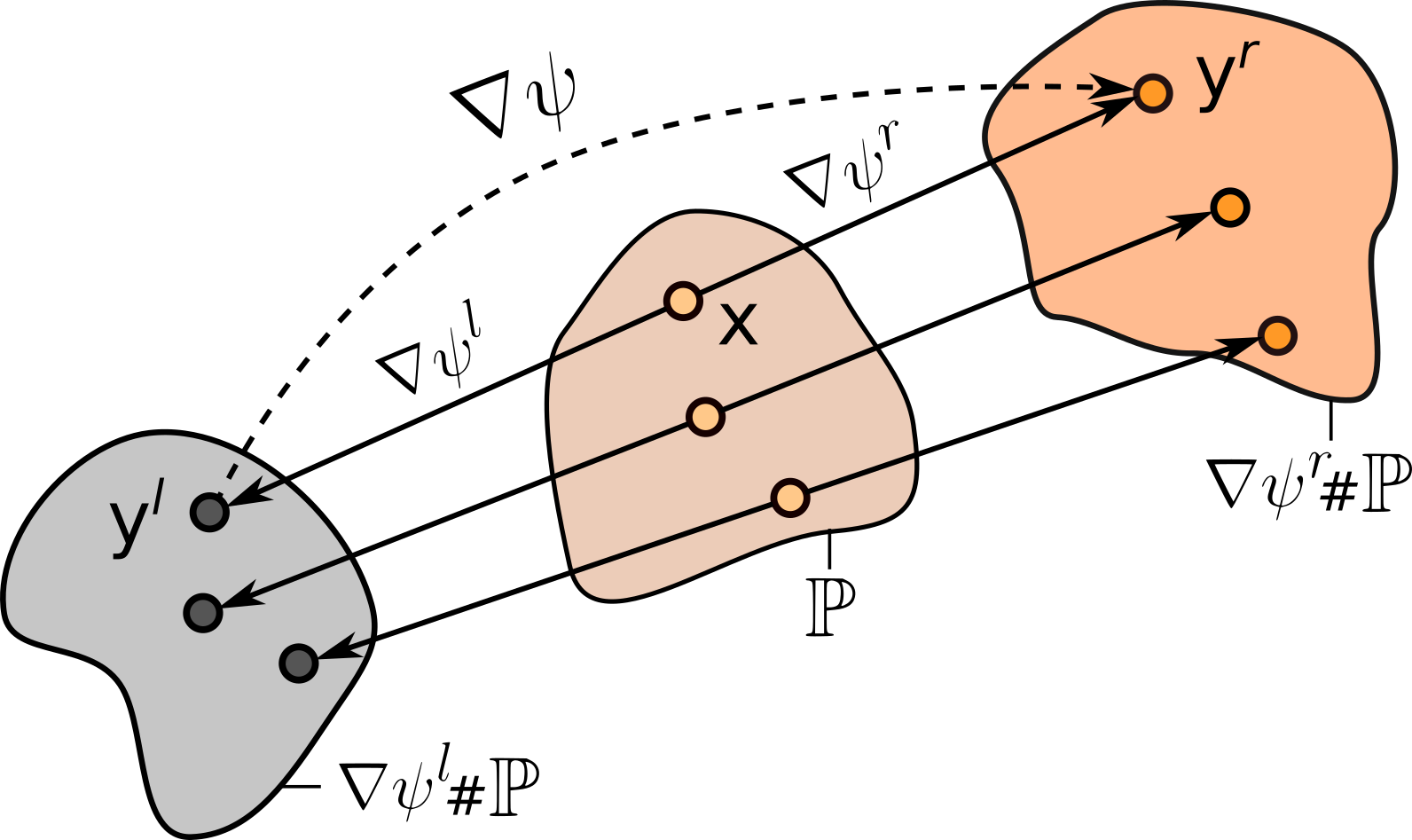}\vspace{3mm}
         
        \caption{\centering Construction of a pair of congruent functions $\psi^{l},\psi^{r}$ from a convex $\psi$, see Lemma \ref{lemma-cong-pair}.}
        \label{fig:pair-construction}\vspace{4mm}
    \end{subfigure}\vrule
     \begin{subfigure}[b]{0.47\textwidth}
        \centering
        \includegraphics[width=0.93\linewidth]{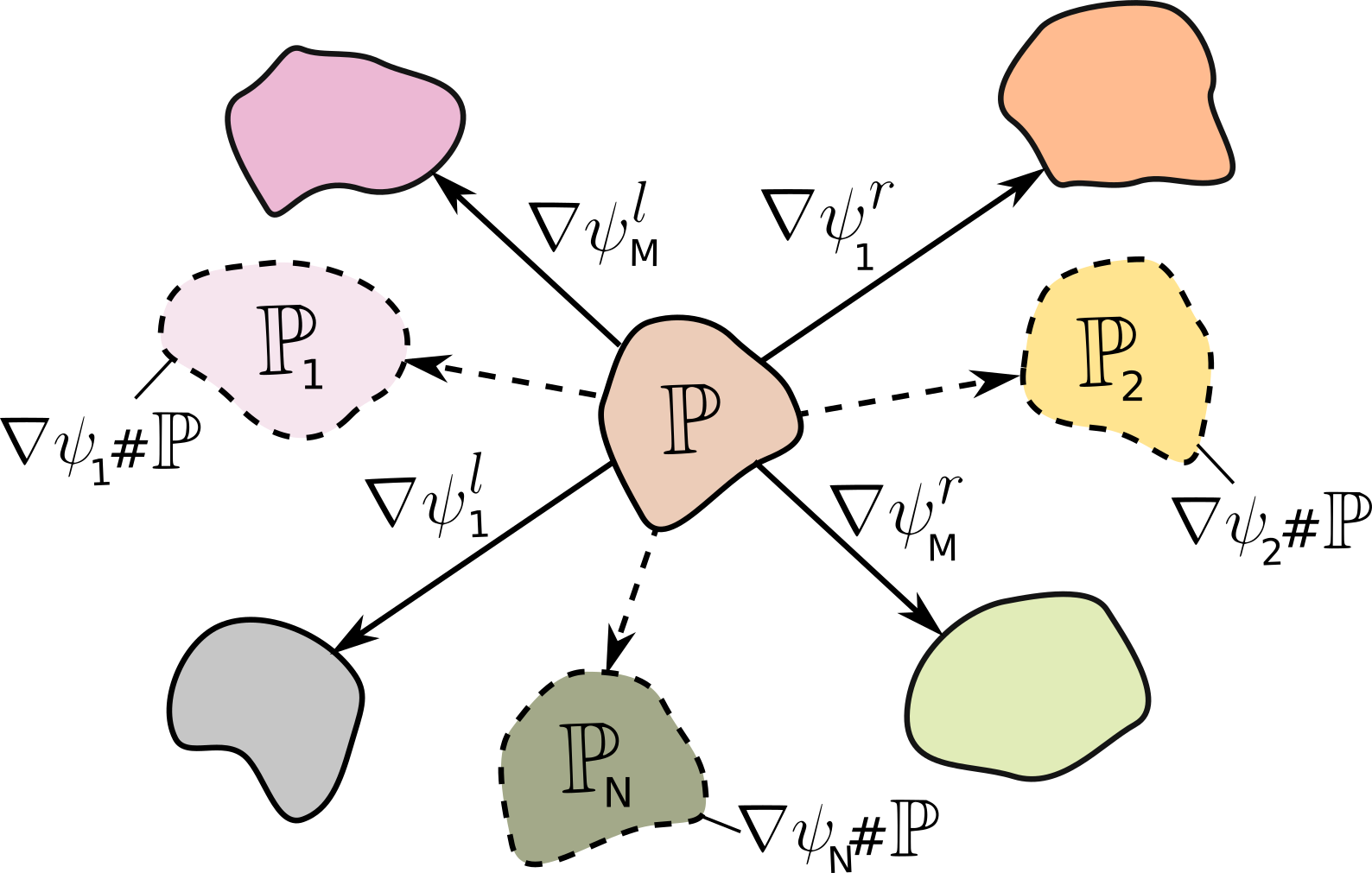}\vspace{3mm}
        
        \caption{\centering Construction of $N$ congruent $\psi_{n} $ as convex combinations of $M$ congruent pairs $(\psi_{m}^{l},\psi_{m}^{r})$, \protect\linebreak see Lemma \ref{lemma-cong-n-tuple}.}
        \label{fig:ntuple-construction}
     \end{subfigure}
    \caption{\centering Construction of tuples of congruent functions\protect\linebreak and production of measures with known $\mathbb{W}_{2}$ barycenter (\wasyparagraph\ref{sec-ave-celeba}).}
    \vspace{-5mm}
\end{figure*}

\begin{lemma}[Constructing congruent pairs]
\label{lemma-conj-pair}
Let $\psi$ be a strongly convex and $L$-smooth (for some $L>0$) function. Let $\beta\in(0,1)$. Define $\beta$-left and $\beta$-right functions of $\psi$ by 
\begin{equation}\psi^{l}\stackrel{\text{def}}{=}\overline{\beta\frac{\|\cdot\|^{2}}{2}+(1-\beta)\psi}\qquad\text{and}\qquad\psi^{r}\stackrel{\text{def}}{=}\overline{(1-\beta)\frac{\|\cdot\|^{2}}{2}+\beta\overline{\psi}}.\label{left-right-def}\end{equation}
Then $\beta\psi^{l}(x)+(1-\beta)\psi^{r}(x)=\frac{\|x\|^{2}}{2}$ for $x\in\mathbb{R}^{D}$, i.e., convex functions $\psi^{l},\psi^{r}$ are congruent w.r.t. weights ${(\beta,1-\beta)}$. Besides, for all $x\in\mathbb{R}^{D}$ the gradient $y^{l}\stackrel{\text{def}}{=}\nabla\psi^{l}(x)$ can be computed via solving $\beta$-strongly concave optimization:
\begin{equation}y^{l}=\argmax_{y\in\mathbb{R}}\bigg(\langle x,y\rangle-\beta\frac{\|y\|^{2}}{2}-(1-\beta)\psi(y)\bigg).
\label{y-left-opt}
\end{equation}
In turn, the value $y^{r}\stackrel{\text{def}}{=}\nabla\psi^{r}(x)$ is given by $y^{r}=\nabla\psi(y^{l})$.
\label{lemma-cong-pair}
\end{lemma}
The proof is given in Appendix \ref{sec-proofs}. We visualize the idea of our Lemma \ref{lemma-cong-pair} in Figure \ref{fig:pair-construction}. Thanks to Lemma \ref{lemma-cong-pair}, any analytically known convex $\psi$, e.g., an ICNN, can be used to produce a congruent pair $\psi^{l}$, $\psi^{r}$. To compute the gradient maps, optimization \eqref{y-left-opt} can be solved by convex optimization tools with $\nabla\psi$ computed by automatic differentiation.

\begin{lemma}[Constructing $N$ congruent functions.]
\label{lemma-conj-n}
Let $\psi_{1}^{0},\dots,\psi_{M}^{0}$ be convex functions, $\beta_{1},\dots,\beta_{M}\in (0,1)$ and $\psi^{l}_{m},\psi^{r}_{m}$ be $\beta_{m}$-left, $\beta_{m}$-right functions for $\psi_{m}^{0}$ respectively. Let $\gamma^{l},\gamma^{r}\in\mathbb{R}^{N\times M}$ be two rectangular matrices with non-negative elements and the sum of elements in each column equals to $1$. Let $w_{1},\dots,w_{M}>0$ satisfy $\sum_{m=1}^{M}w_{m}=1$. For $n=1,\dots,N$ define
\begin{equation}
    \psi_{n}(x)\!\stackrel{\text{def}}{=}\! \frac{\sum_{m=1}^{M}\!w_{m}\big[\beta_{m}\gamma^{l}_{nm} \psi_{m}^{l}(x)\!+\!(1\!-\!\beta_{m})\gamma^{r}_{nm} \psi_{m}^{r}(x)\big]}{\sum_{m=1}^{M}\!w_{m}\big[\beta_{m}\gamma^{l}_{nm}\!+\!(1\!-\!\beta_{m})\gamma^{r}_{nm}\big]}.
    \label{conv-comb-cong}
\end{equation}
Then $\psi_{1},\dots,\psi_{N}$ are congruent w.r.t. weights $\alpha_{n}\stackrel{\text{def}}{=}\sum_{m=1}^{M}w_{m}\big[\beta_{m}\gamma^{l}_{nm}+(1-\beta_{m})\gamma^{r}_{nm}\big].$
\label{lemma-cong-n-tuple}
\vspace{-2mm}
\end{lemma}
We prove Lemma \ref{lemma-conj-n} in Appendix \ref{sec-proofs}. We visualize the idea of our Lemma \ref{lemma-cong-n-tuple} in Figure \ref{fig:ntuple-construction}. The lemma provides an elegant way to create  $N\geq 2$ congruent functions from convex linear combinations of functions in given congruent pairs $(\psi_{m}^{l},\psi_{m}^{r})$. Gradients $\nabla\psi_{n}$ of these functions are respective linear combinations of gradients $\nabla\psi_{m}^{l}$ and $\nabla\psi_{m}^{r}$.

\textbf{Dataset creation.} We use CelebA $64\times64$ faces dataset \citep{liu2015faceattributes} as the basis for our Ave, celeba! dataset. We assume that CelebA dataset is an empirical sample from the continuous measure ${\mathbb{P}_{\text{Celeba}}\in\mathcal{P}_{2,\text{ac}}^{+}(\mathbb{R}^{3\times 64\times 64})}$ which we put to be the barycenter in our design, i.e., ${\overline{\mathbb{P}}=\mathbb{P}_{\text{Celeba}}}$. We construct diffirentiable congruent $\psi_{n}$ with bijective gradients that produce ${\mathbb{P}_{n}=\nabla\psi_{n}\sharp\overline{\mathbb{P}}\in\mathcal{P}_{2,\text{ac}}^{+}(\mathbb{R}^{3\times 64\times 64})}$ whose unique barycenter is $\mathbb{P}_{\text{Celeba}}$. In Lemma \ref{lemma-cong-n-tuple}, we set $N=3$, $M=2$, $\beta_{1}=\beta_{2}=\frac{1}{2}$, $w_{1}=w_{2}=\frac{1}{2}$ and 
\vspace{-2mm}\begin{equation}
    (\gamma^{l})^{\top}=\begin{pmatrix}
    1 & 0 & 0\\
    0 & 1 & 0
    \end{pmatrix},\qquad
    (\gamma^{r})^{\top}=\begin{pmatrix}
    0 & 1 & 0\\
    0 & 0 & 1
    \end{pmatrix}
    \nonumber\vspace{-0.3mm}\vspace{-2mm}
\end{equation}
which yields weights $(\alpha_{1},\alpha_{2},\alpha_{3})=(\frac{1}{4}, \frac{1}{2}, \frac{1}{4})$ We choose the constants above manually to make sure the final produced measures $\mathbb{P}_{n}$ are visually distinguishable. We use $\psi_{m}^{0}(x)=\text{ICNN}_{m}\big(s_{m}(\sigma_{m}(d_{m}(x)))\big)+\lambda\frac{\|x\|^{2}}{2}$ as convex functions, where ICNNs have ConvICNN64 architecture \citep[Appendix B.1]{korotin2021neural}, $\sigma_{1},\sigma_{2}$ are random permutations of pixels and channels, $s_{1}, s_{2}$ are axis-wise random reflections, $\lambda=\frac{1}{100}$. In both functions, $d_{m}$ is a de-colorization transform which sets R, G, B channels of each pixel to $(\frac{7}{10}\text{R}\!+\!\frac{1}{25}\text{G}\!+\!\frac{13}{50}\text{B})$ for $\psi_{1}^{0}$ and $\frac{1}{3}(\text{R}\!+\!\text{G}\!+\!\text{B})$ for $\psi_{2}^{0}$. The weights of ICNNs are initialized by the pre-trained potentials of $\mathbb{W}_{2}^{2}$ "Early"\ transport benchmark which map blurry faces to the clean ones \citep[\wasyparagraph4.1]{korotin2021neural}. All the implementation details are given in Appendix \ref{sec-ave-celeba-creation}.

\begin{figure*}[!t]
\centering
\includegraphics[width=0.98\linewidth]{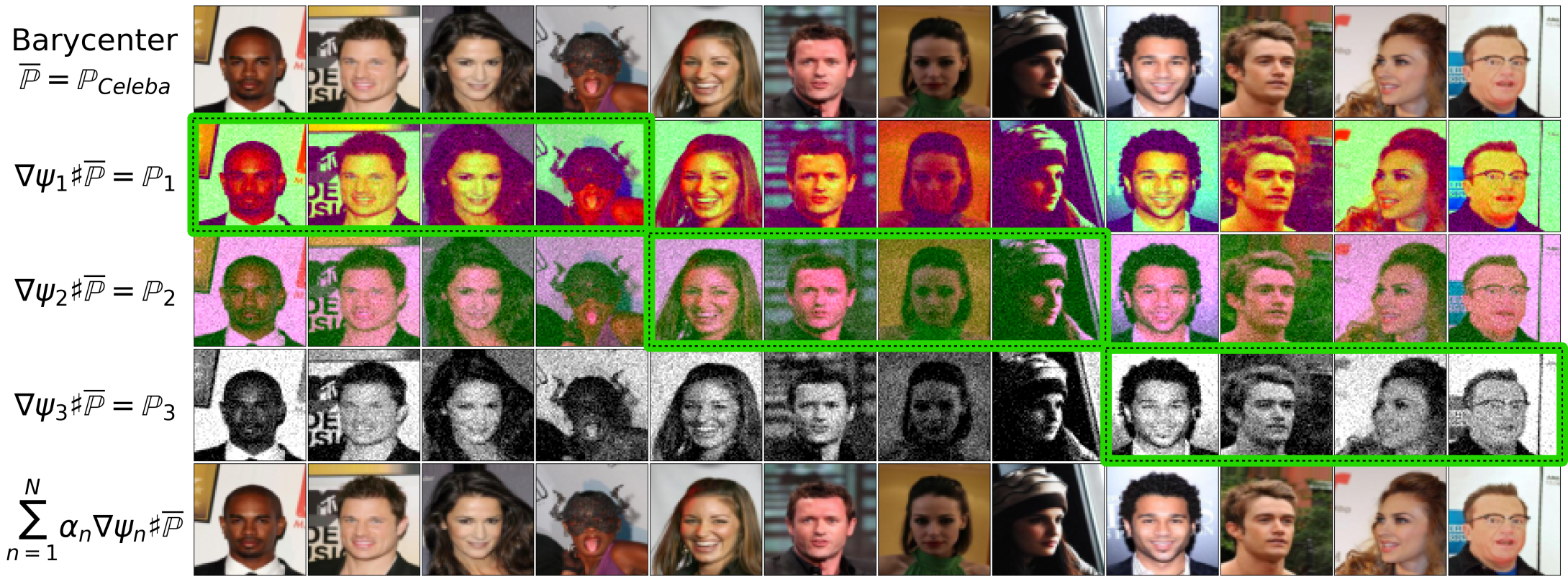}
\caption{The production of \underline{\textbf{Ave, celeba! dataset}}. The 1st line shows images ${x\sim\mathbb{P}_{\text{Celeba}}}$. Each of 3 next lines shows OT maps ${\nabla\psi_{n}(x)\sim\nabla\psi_{n}\sharp\mathbb{P}_{\text{Celeba}}=\mathbb{P}_{n}}$ to constructed measures $\mathbb{P}_{n}$. Their barycenter w.r.t. $(\alpha_{1},\alpha_{2},\alpha_{3})=(\frac{1}{4}, \frac{1}{2}, \frac{1}{4})$ is $\mathbb{P}_{\text{Celeba}}$. The last line shows congruence of $\psi_{n}$, i.e., ${\sum_{n=1}^{N}\alpha_{n}\nabla\psi_{n}(x)\equiv x}$. Samples \textbf{in green boxes} are included to dataset.}
\label{fig:ave-celeba}
\vspace{-4.5mm}
\end{figure*}

Finally, to create \textit{Ave, celeba!} dataset, we randomly split the images dataset into 3 equal parts containing $\approx 67$K samples, and map each part to respective measure ${\mathbb{P}_{n}=\nabla\psi_{n}\sharp \mathbb{P}_{\text{Celeba}}}$ by $\nabla\psi_{n}$. Resulting $3\times 67$K samples form the dataset consisting of $3$ parts. We show the samples in Figure \ref{fig:ave-celeba}. The samples from the respective parts are in green boxes.

\vspace{-2.5mm}\section{Evaluation}
\label{sec-evaluation}

\vspace{-2.5mm}The code\footnote{\url{https://github.com/iamalexkorotin/WassersteinIterativeNetworks}} is written on the PyTorch and includes the script for producing Ave, celeba! dataset. The experiments are conducted on 4$\times$GPU GTX 1080ti. The details are given in Appendix \ref{sec-exp-details}.

\vspace{-1.8mm}\subsection{Evaluation on Ave, celeba! Dataset}
\label{sec-exp-ave-celeba}

\begin{figure*}[!t]
\vspace{-2mm}
\centering
\begin{subfigure}[b]{0.48\textwidth}
     \centering
     \includegraphics[width=0.95\linewidth]{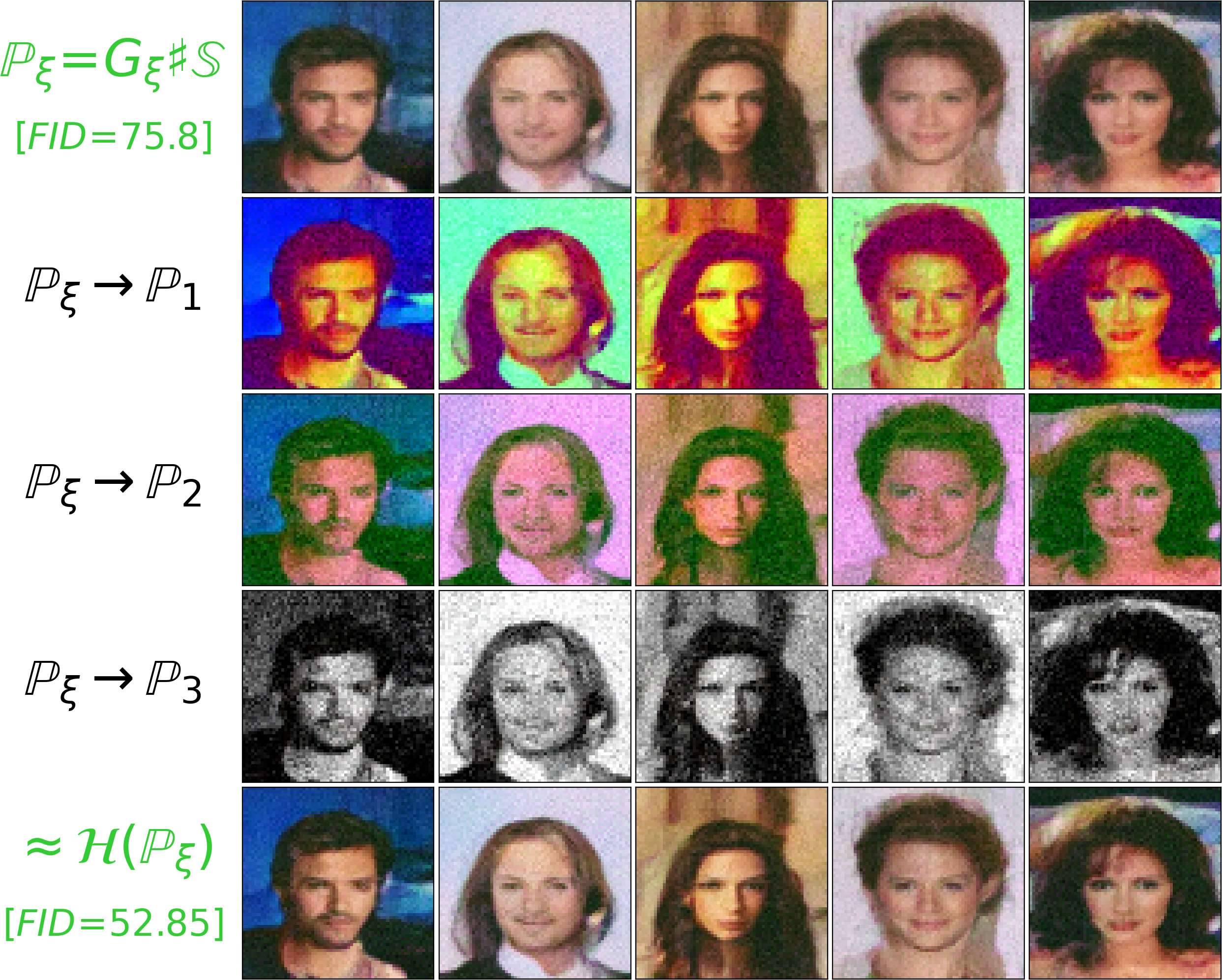}
    \caption{{\color{LimeGreen}\underline{\textbf{Our}}} Algorithm \ref{algorithm-win}.}
    \label{fig:ave-celeba-win}
\end{subfigure}\hspace{1mm}\begin{subfigure}[b]{0.48\textwidth}
 \centering
 \includegraphics[width=0.95\linewidth]{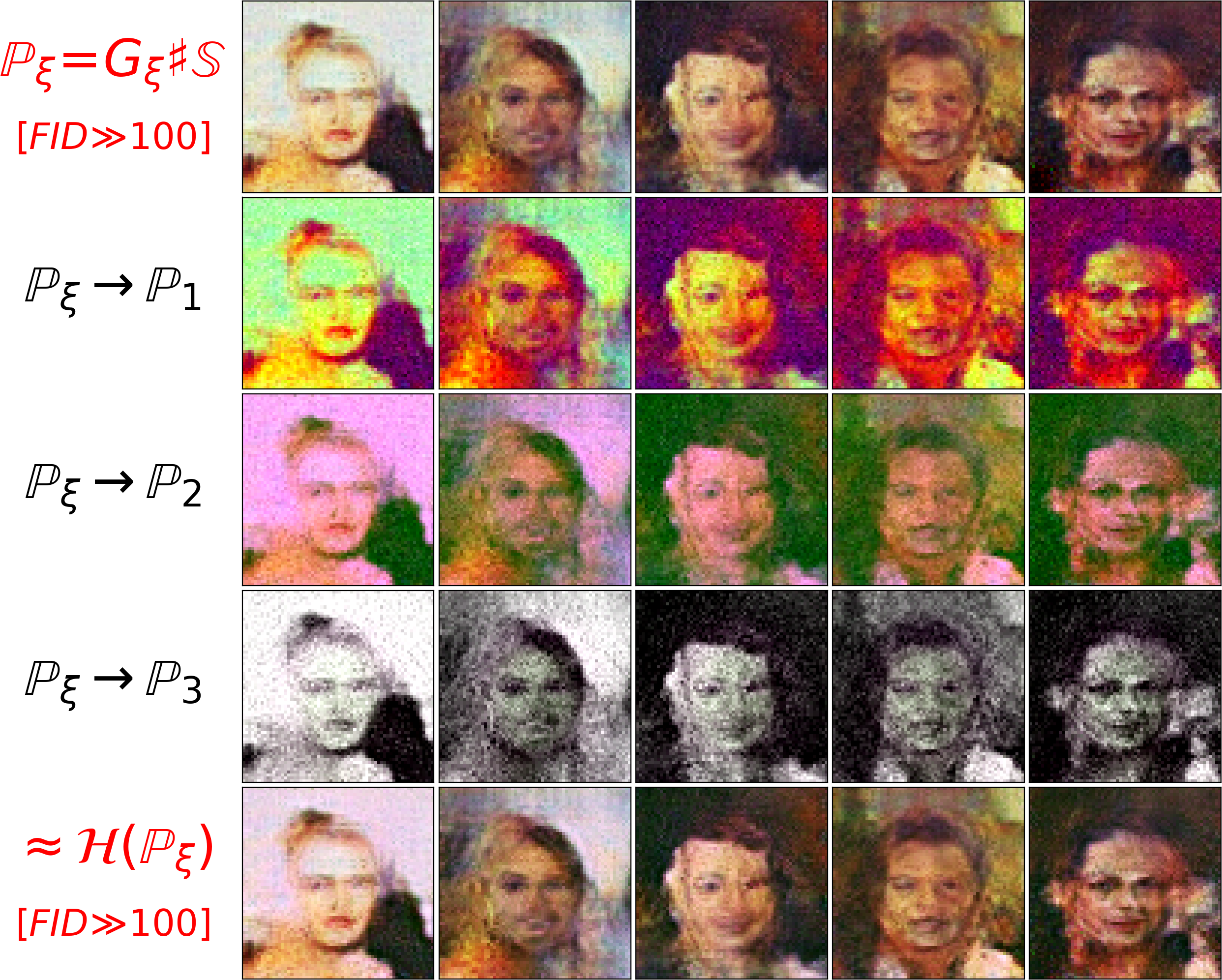}
\caption{{\color{red}\underline{\textbf{Competitive}}} $[\text{SC}\mathbb{W}_{2}\text{B}]$ algorithm.}
\label{fig:ave-celeba-icnn}
\end{subfigure}
\caption{The barycenter and maps to input measures estimated by barycenter algorithms. The 1st line shows generated samples $\mathbb{P}_{\xi}=G_{\xi}\sharp \mathbb{S}\approx \mathbb{P}_{\text{Celeba}}$. Lines 2-4 show maps $\widehat{T}_{\mathbb{P}_{\xi}\rightarrow\mathbb{P}_{n}}$. The last line shows the average map $\sum_{n=1}^{N}\alpha_{n}\widehat{T}_{\mathbb{P}_{\xi}\rightarrow\mathbb{P}_{n}}$.}
\label{fig:ave-celeba-gen}
\vspace{-3mm}
\end{figure*}

% \begin{wraptable}{r}{6.6cm}
% \vspace{-4mm}
% \centering
% \footnotesize
% \begin{tabular}{|c|c|c|}
% \hline
% \multicolumn{2}{|c|}{\textit{Method}} & \textit{FID}$\downarrow$  \\ \hline
% \multirow{2}{*}{\shortstack[c]{$[\text{SC}\mathbb{W}_{2}B]$}} & $G_{\xi}(z)$ &  \color{red}{156.3}   \\ \cline{2-3}
%  & $\sum_{n=1}^{N}\alpha_{n}\widehat{T}_{\mathbb{P}_{\xi}\rightarrow\mathbb{P}_{n}}\big(G_{\xi}(z)\big)$ & \color{red}{152.1}    \\
% \hline
% \multirow{2}{*}{\textbf{Ours}} & $G_{\xi}(z)$ &  {\color{LimeGreen}{75.8}}   \\ \cline{2-3}
%  & $\sum_{n=1}^{N}\alpha_{n}\widehat{T}_{\mathbb{P}_{\xi}\rightarrow\mathbb{P}_{n}}\big(G_{\xi}(z)\big)$ & {\color{LimeGreen}{52.85}}    \\ \hline
% \end{tabular}
% \vspace{-1mm}
% \caption{\centering FID scores of images \protect\linebreak from the learned barycenter.}
% \vspace{-5mm}
% \label{table-fid-ave-celeba}
% \end{wraptable}
\vspace{-1.5mm}We evaluate our \textit{iterative} algorithm \ref{algorithm-win} and a recent \underline{state-of-the-art} \textit{variational} $[\text{SC}\mathbb{W}_{2}\text{B}]$ by \cite{fan2020scalable} on Ave, celeba! dataset. Both algorithms use a generative model $\mathbb{P}_{\xi}=G_{\xi}\sharp\mathbb{S}$ for the barycenter and yield approximate maps $\widehat{T}_{\mathbb{P}_{\xi}\rightarrow\mathbb{P}_{n}}$ to input measures. In our case, the maps are neural networks $T_{\theta_{n}}$, while in $[\text{SC}\mathbb{W}_{2}\text{B}]$ they are gradients of ICNNs. The barycenters of Ave, celeba! fitted by our algorithm and $[\text{SC}\mathbb{W}_{2}\text{B}]$ are shown in Figures \ref{fig:ave-celeba-win} and \ref{fig:ave-celeba-icnn} respectively. Recall the ground truth barycenter is $\mathbb{P}_{\text{Celeba}}$. Thus, for quantitative evaluation we use FID score \citep{heusel2017gans} computed on 200K generated samples w.r.t. the original CelebA dataset, see Table \ref{table-fid-ave-celeba}. Our method \textit{drastically} outperforms $[\text{SC}\mathbb{W}_{2}\text{B}]$. Presumably, this is due to the latter using ICNNs which do not provide sufficient performance.

\begin{figure*}[!t]
    \centering
    \begin{subfigure}[b]{0.33\textwidth}
         \centering
         \includegraphics[width=0.95\linewidth]{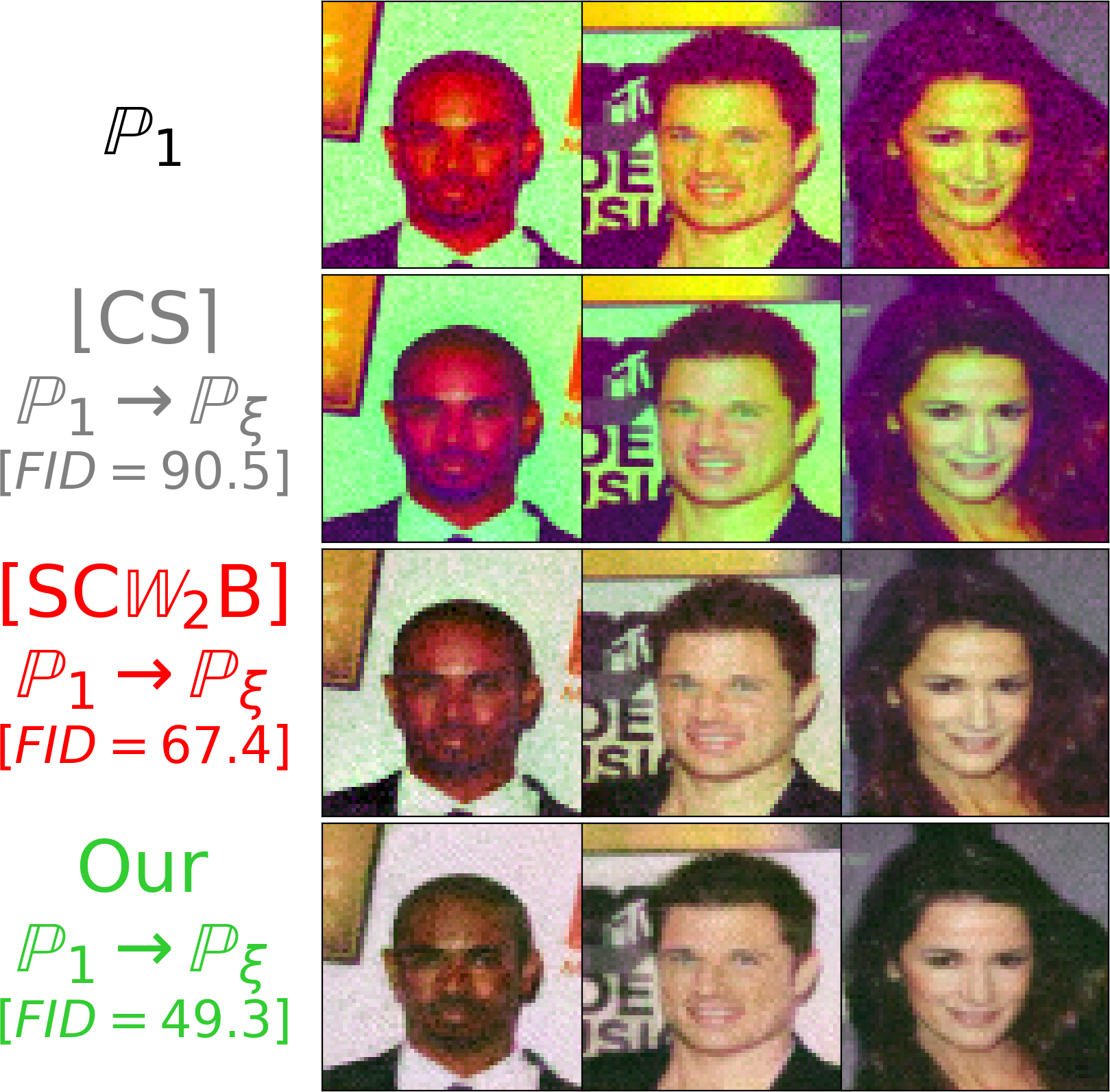}
        \caption{Maps from $\mathbb{P}_{1}$ to the barycenter.}
        \label{fig:celeba-win}
    \end{subfigure}\hfill
     \begin{subfigure}[b]{0.33\textwidth}
        \centering
        \includegraphics[width=0.95\linewidth]{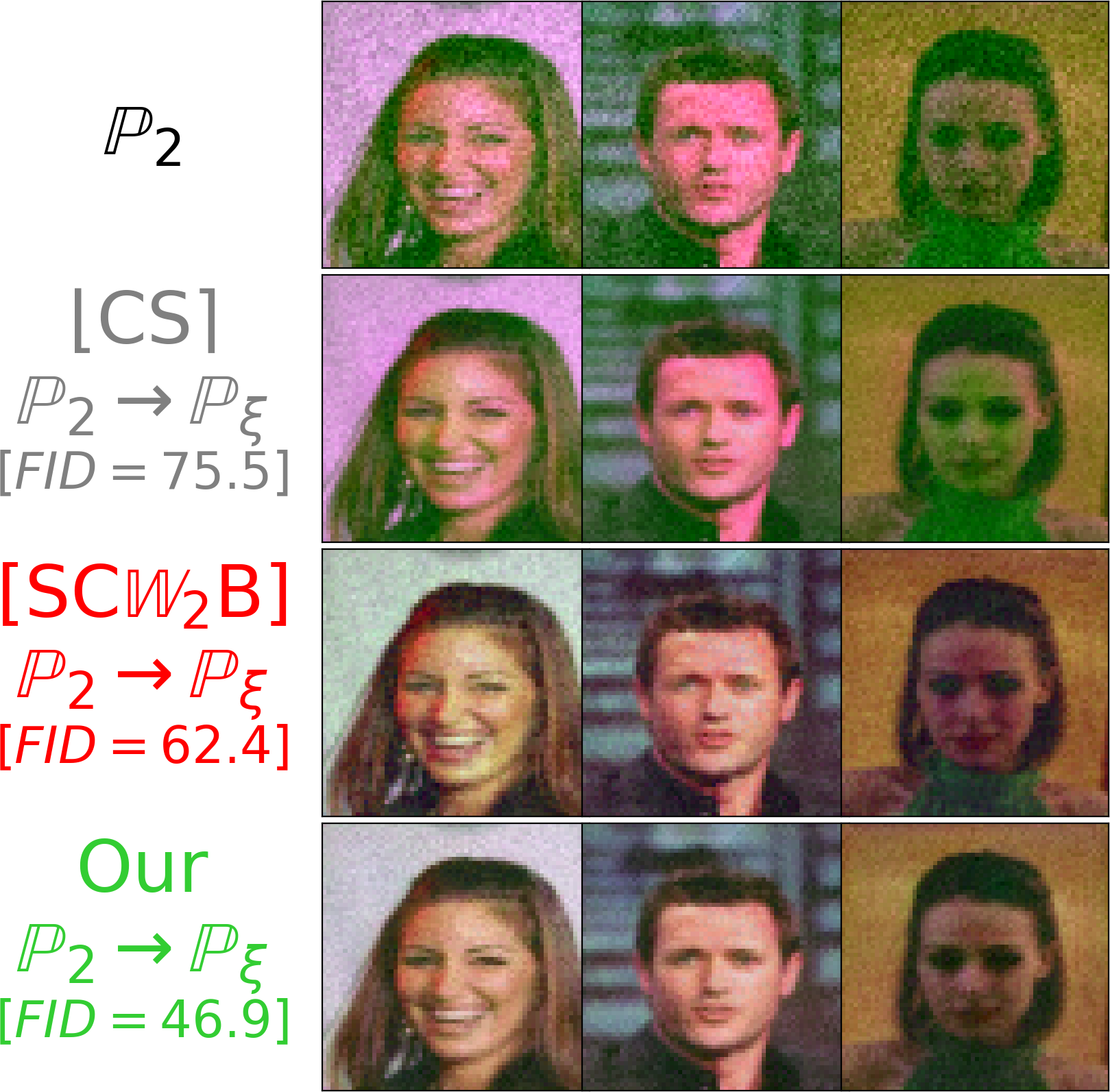}
        \caption{Maps from $\mathbb{P}_{2}$ to the barycenter.}
        \label{fig:celeba-win-fixed}
     \end{subfigure}
     \begin{subfigure}[b]{0.33\textwidth}
        \centering
        \includegraphics[width=0.95\linewidth]{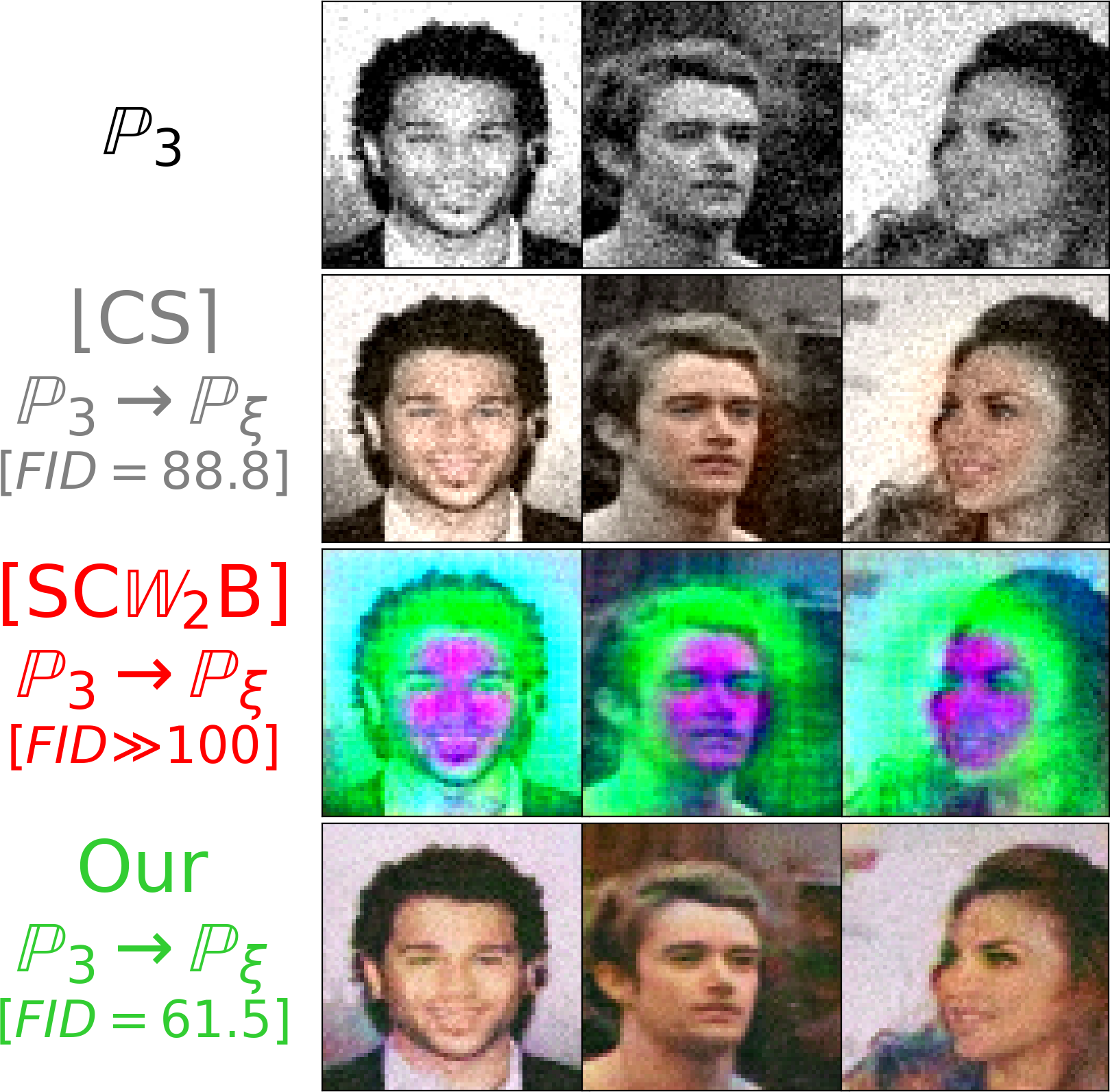}
        \caption{Maps from $\mathbb{P}_{3}$ to the barycenter.}
        \label{fig:celeba-win-fixed}
     \end{subfigure}
\vspace{-4.6mm}
    \caption{Maps from inputs $\mathbb{P}_{n}$ to the barycenter $\overline{\mathbb{P}}$ estimated by the barycenter algorithms in view. For comparison with the original barycenter images, the faces are the same as in Figure \ref{fig:ave-celeba}.}
    \label{fig:ave-celeba-inverse-maps}
\vspace{-3mm}
\end{figure*}

% \begin{wraptable}{r}{7.6cm} \vspace{-3mm}
% \centering
% \footnotesize
% \begin{tabular}{|c|c|c|c|c|}
% \hline
% \multicolumn{2}{|c|}{\multirow{2}{*}{\textit{Method}}} & \multicolumn{3}{|c|}{\textit{FID}$\downarrow$}  \\  \cline{3-5}
% \multicolumn{2}{|c|}{} & $n=1$ & $n=2$ & $n=3$  \\
% \hline
% $\lfloor \text{CS}\rceil$ & $y+(\overline{\mu}-\mu_{n})$  &  {\color{gray}90.5} & {\color{gray}75.5} & {\color{gray}88.8}   \\ \hline
% $[\text{SC}\mathbb{W}_{2}B]$ & $\widehat{T}_{\mathbb{P}_{n}\rightarrow\mathbb{P}_{\xi}}(y)$ &  \color{red}{67.4} & \color{red}{62.4} & \color{red}{319.62} \\
% \cline{2-3}
% \hline
% \textbf{Ours} & $\widehat{T}_{\mathbb{P}_{n}\rightarrow\mathbb{P}_{\xi}}(y)$ &  {\color{LimeGreen}49.3} & {\color{LimeGreen}46.9} & {\color{LimeGreen}61.5} \\ \cline{2-3} \hline
% \end{tabular}
% \vspace{-1mm}
% \caption{\centering FID scores of images \protect\linebreak mapped from inputs $\mathbb{P}_{n}$.}
% \vspace{-3mm}
% \label{table-fid-ave-celeba-inv}\vspace{-2mm}
% % \cite{arjovsky2017wasserstein}
% \end{wraptable}
Additionally, we evaluate to which extent the algorithms allow to recover the inverse OT maps $T_{\mathbb{P}_{n}\rightarrow\mathbb{P}_{\xi}}$ from inputs $\mathbb{P}_{n}$ to the barycenter $\mathbb{P}_{\xi}\approx\overline{\mathbb{P}}$. In $[\text{SC}\mathbb{W}_{2}\text{B}]$, these maps are computed during training. Our algorithm does not compute them. Thus, we separately fit the inverse maps after main training by using $\lfloor\text{MM:R}\rceil$ solver between each input $\mathbb{P}_{n}$ and learned $\mathbb{P}_{\xi}$ (Algorithm \ref{algorithm-win-inverse} of Appendix \ref{algorithm-win-inverse}). The inverse maps are given in Figure \ref{fig:ave-celeba-inverse-maps}; their FID scores -- in Table \ref{table-fid-ave-celeba-inv}. Here we add an additional \textit{constant shift} $\lceil\text{CS}\rfloor$ baseline which simply shifts the mean of input $\mathbb{P}_{n}$ to the mean $\overline{\mu}$ of $\overline{\mathbb{P}}$. The vector $\overline{\mu}$ is given by $\sum_{n=1}^{N}\alpha_{n}\mu_{n}$, where $\mu_{n}$ is the mean of $\mathbb{P}_{n}$ \citep{alvarez2016fixed}. We estimate $\overline{\mu}$ from samples $y\sim\mathbb{P}_{n}$.
% \vspace{-2mm}

\begin{table}[!t]
   \begin{minipage}{.46\linewidth}
      \centering
        \scriptsize
        \begin{tabular}{|c|c|c|}
        \hline
        \multicolumn{2}{|c|}{\textit{Method}} & \textit{FID}$\downarrow$  \\ \hline
        \multirow{2}{*}{\shortstack[c]{$[\text{SC}\mathbb{W}_{2}B]$}} & $G_{\xi}(z)$ &  \color{red}{156.3}   \\ \cline{2-3}
         & $\sum_{n=1}^{N}\alpha_{n}\widehat{T}_{\mathbb{P}_{\xi}\rightarrow\mathbb{P}_{n}}\big(G_{\xi}(z)\big)$ & \color{red}{152.1}    \\
        \hline
        \multirow{2}{*}{\textbf{Ours}} & $G_{\xi}(z)$ &  {\color{LimeGreen}{75.8}}   \\ \cline{2-3}
         & $\sum_{n=1}^{N}\alpha_{n}\widehat{T}_{\mathbb{P}_{\xi}\rightarrow\mathbb{P}_{n}}\big(G_{\xi}(z)\big)$ & {\color{LimeGreen}{52.85}}    \\ \hline
        \end{tabular}
        \vspace{3mm}
        \caption{\centering FID scores of images \protect\linebreak from the learned barycenter.}
        \label{table-fid-ave-celeba}
    \end{minipage}%
    \begin{minipage}{.54\linewidth}
      \centering
        \scriptsize
\begin{tabular}{|c|c|c|c|c|}
\hline
\multicolumn{2}{|c|}{\multirow{2}{*}{\textit{Method}}} & \multicolumn{3}{|c|}{\textit{FID}$\downarrow$}  \\  \cline{3-5}
\multicolumn{2}{|c|}{} & $n=1$ & $n=2$ & $n=3$  \\
\hline
$\lfloor \text{CS}\rceil$ & $y+(\overline{\mu}-\mu_{n})$  &  {\color{gray}90.5} & {\color{gray}75.5} & {\color{gray}88.8}   \\ \hline
$[\text{SC}\mathbb{W}_{2}B]$ & $\widehat{T}_{\mathbb{P}_{n}\rightarrow\mathbb{P}_{\xi}}(y)$ &  \color{red}{67.4} & \color{red}{62.4} & \color{red}{319.62} \\
\cline{2-3}
\hline
\textbf{Ours} & $\widehat{T}_{\mathbb{P}_{n}\rightarrow\mathbb{P}_{\xi}}(y)$ &  {\color{LimeGreen}49.3} & {\color{LimeGreen}46.9} & {\color{LimeGreen}61.5} \\ \cline{2-3} \hline
\end{tabular}
\vspace{3mm}
\caption{\centering FID scores of images \protect\linebreak mapped from inputs $\mathbb{P}_{n}$.}
\label{table-fid-ave-celeba-inv}
    \end{minipage} 
    \vspace{-9mm}
\end{table}

\vspace{-2.5mm}\subsection{Additional Experimental Results}
\begin{figure*}[!t]
\vspace{-2.5mm}
\begin{subfigure}{0.3\linewidth}
\centering
\includegraphics[width=0.97\linewidth]{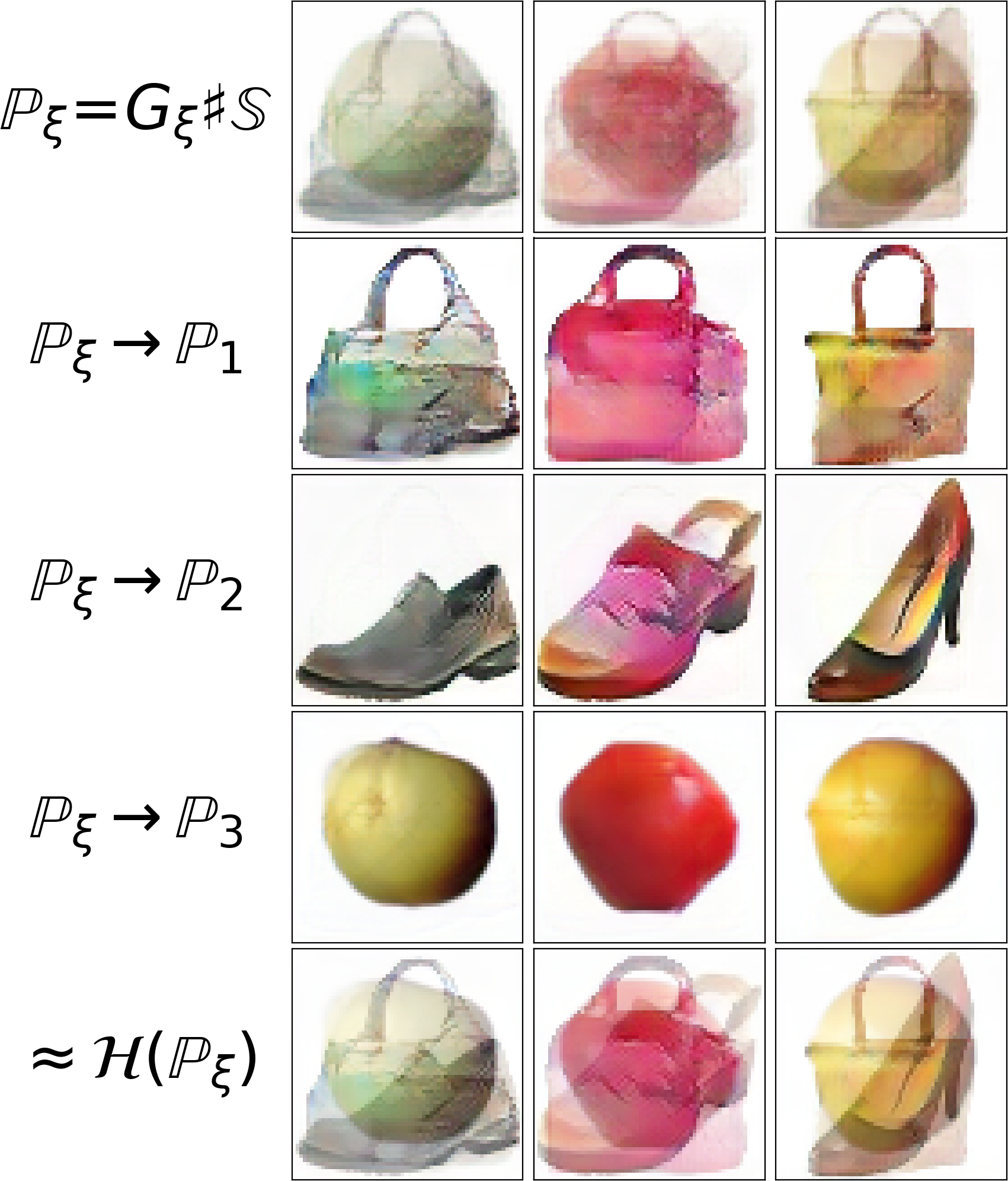}\caption{\centering Generated samples $\mathbb{P}_{\xi}\approx\overline{\mathbb{P}}$, fitted maps to each $\mathbb{P}_{n}$ and their average.}
\label{fig:fruit-generated}
\end{subfigure}
\begin{subfigure}{0.225\linewidth}
\centering
\includegraphics[width=0.97\linewidth]{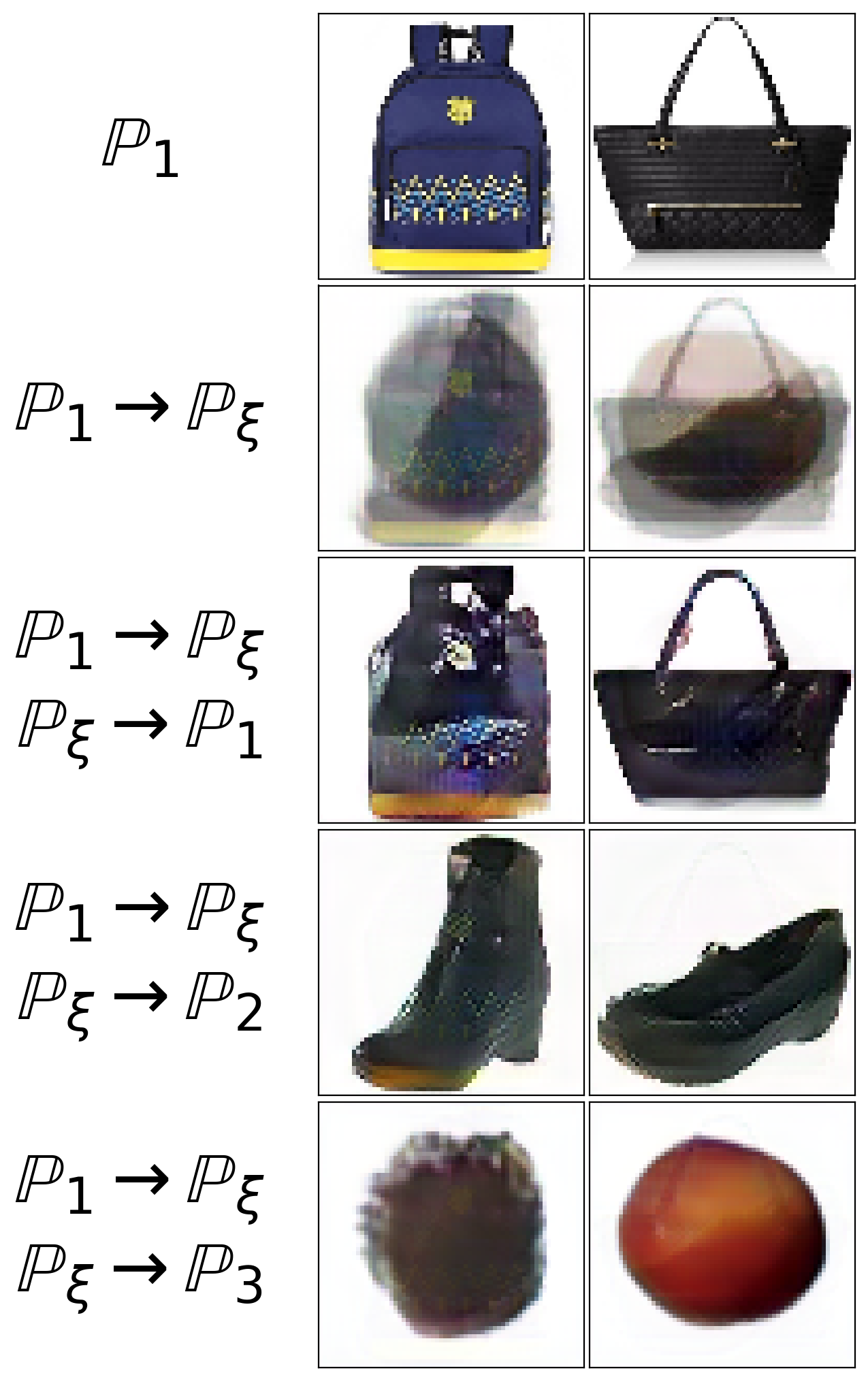}
\caption{\centering Samples $y\sim\mathbb{P}_{1}$ mapped through $\mathbb{P}_{\xi}$ to each $\mathbb{P}_{n}$.}
\label{fig:fruit-through1}
\end{subfigure}
\begin{subfigure}{0.225\linewidth}
\centering
\includegraphics[width=0.97\linewidth]{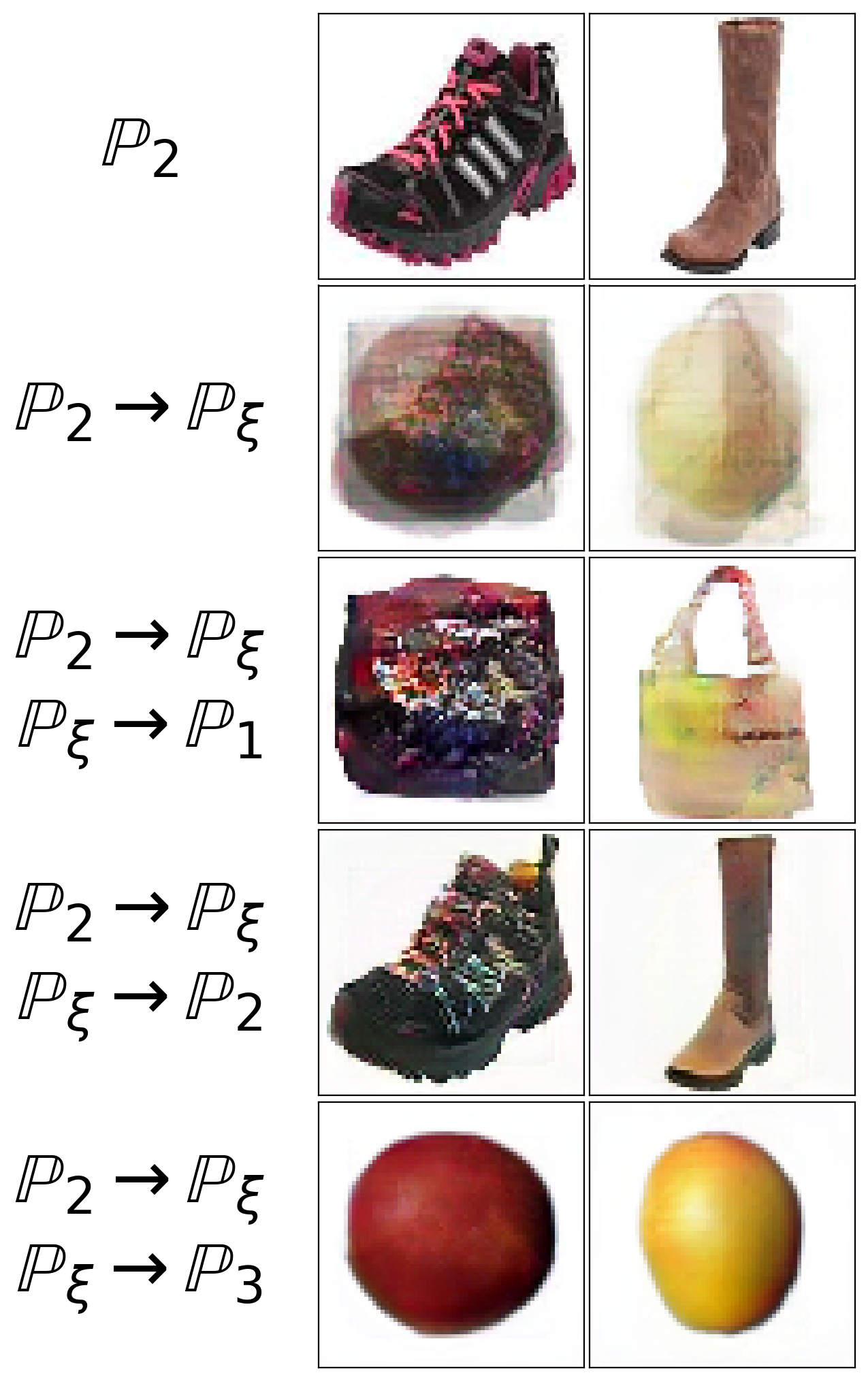}
\caption{\centering Samples $y\sim\mathbb{P}_{2}$ mapped through $\mathbb{P}_{\xi}$ to each $\mathbb{P}_{n}$.}
\label{fig:fruit-through2}
\end{subfigure}\begin{subfigure}{0.225\linewidth}
\centering
\includegraphics[width=0.97\linewidth]{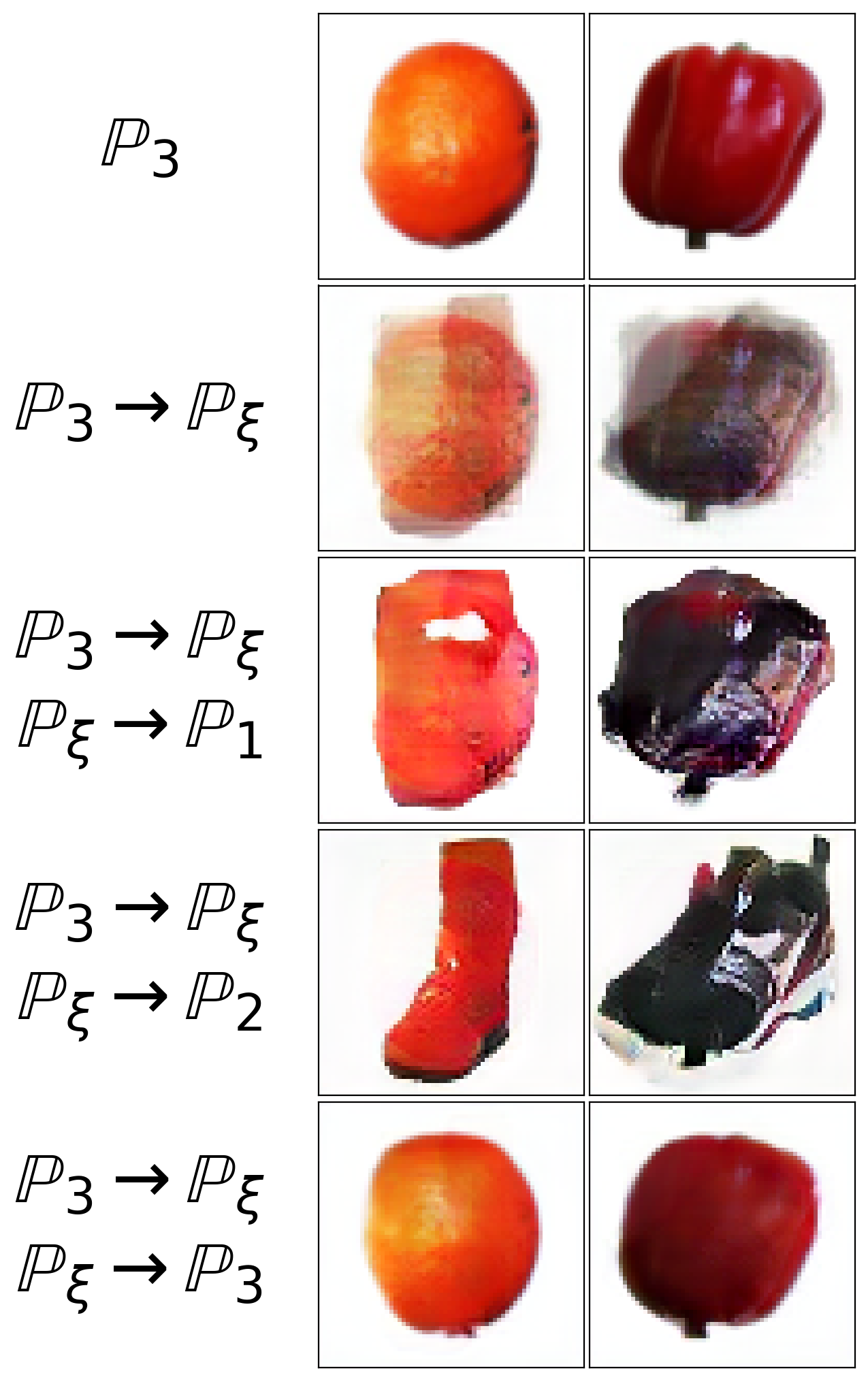}
\caption{\centering Samples $y\sim\mathbb{P}_{3}$ mapped through $\mathbb{P}_{\xi}$ to each $\mathbb{P}_{n}$.}
\label{fig:fruit-through3}
\end{subfigure}
\vspace{-1.6mm}
\caption{\centering The barycenter of Handbags, Shoes, Fruit ($64\times 64$) datasets fitted by {\color{LimeGreen}\underline{\textbf{our}}} Algorithm \ref{algorithm-win}. \protect\linebreak We give the results of {\color{Red}\underline{\textbf{competitive}}} [SC$\mathbb{W}_{2}$B] in Appendix \ref{sec-scwb-params}.}
\vspace*{-0.27in}
\label{fig:handbag-shoe-fruit}
\end{figure*}
\vspace{-1mm}\textbf{Different domains.} To stress-test our algorithm \ref{algorithm-win}, we compute the barycenters w.r.t. $(\alpha_{1},\alpha_{2},\alpha_{3})=(\frac{1}{3},\frac{1}{3},\frac{1}{3})$ of notably different datasets: 50K Shoes \citep{yu2014fine}, 138K Amazon Handbags and 90K Fruits \citep{murecsan2017fruit}. All the images are rescaled to $64\times 64$. The \underline{\textit{ground gruth barycenter is unknown}}, but one may imagine what it looks like. Due to \eqref{bar-condition}, each barycenter image is a \textit{pixel-wise average} of a shoe, a handbag and a fruit, which is supported by our result shown in Figure \ref{fig:handbag-shoe-fruit}. 
In the same figure, we also show the maps between datasets \textit{through} the barycenter (Figures \ref{fig:fruit-through1}, \ref{fig:fruit-through2}, \ref{fig:fruit-through3}): this allows generation of images in other categories with styles similar to a given image. 
For instance, in Figure \ref{fig:fruit-through3}, we generate an orange bag and an orange shoe by pushing the image of an orange through the barycenter.
We provide more examples in Figure \ref{fig:handbag-shoe-fruit-ext} of Appendix \ref{sec-extra-results}.

\vspace{-1mm}\textbf{Extra results.} In Appendix \ref{sec-toy-experiments}, we compute the barycenters of toy 2D dsitributions. In Appendix \ref{sec-gaussian-case}, we provide \textit{quantitative} results for computing barycenters in the Gaussian case.  In Appendix \ref{sec-generative-modeling}, we test how our algorithm works as a \textit{generative model} on the original CelebA dataset, i.e., when $N\!=\!1$ and $\mathbb{P}_{1}=\mathbb{P}_{\text{Celeba}}$. We show that in this case it achieves FID scores comparable to recent WGAN models. In Appendix \ref{sec-exp-mnist}, similar to \cite{fan2020scalable}, we compute barycenters of digit classes 0/1 of $32\times 32$ grayscale MNIST \citep{lecun-mnisthandwrittendigit-2010}. We also test our algorithm on FashionMNIST \citep{xiao2017fashion} \textbf{10 classes} dataset.\vspace{-2mm}

\vspace{-2.5mm}\section{Discussion}
\label{sec-discussion}

\vspace{-3mm}\textbf{Potential impact (algorithm).} We present a scalable barycenter algorithm based on fixed-point iterations with many application prospects.
For instance, in medical imaging, MRI is often acquired at multiple sites where the overlap of information (imaging, genetic, diagnosis) between any two sites is limited. Consequently, the data on each site may be biased and can cause generalizability and robustness issues when training models. The developed algorithm could help aggregate data from multiple sites and overcome the distributional shift issue across sites.

\vspace{-1mm}\textbf{Potential impact (dataset).} There is no high-dimensional dataset for the barycenter problem except for location-scattered cases (e.g. Gaussians), where the transport maps are always linear.
Hence our proposed dataset fills an important gap, thereby allowing quantitative evaluation of future related methods.  We expect our Ave, celeba! to become a standard dataset for evaluating continuous barycenter algorithms. In addition, we describe a generic recipe (\wasyparagraph\ref{sec-ave-celeba}) to produce new datasets.

\vspace{-1mm}\textbf{Limitations (algorithm).} In our algorithm, the evolving measure $\mathbb{P}_{\xi}$ is not guaranteed to be continuous, while it is continuous in the underlying fixed point approach. To enforce the absolute continuity of $\mathbb{P}_{\xi}=G_{\xi}\sharp\mathbb{S}$, one may use an invertible network \citep{etmann2020iunets} for $G_{\xi}$ and an absolutely continuous latent measure $\mathbb{S}$ in a latent space of dimension $H\!=\!D$.
However, our results suggest this is unnecessary in practice --- common GANs approaches also assume $H\!\ll\!D$. During the fixed-point iterations, the barycenter objective \eqref{w2-barycenter-def} decreases. However, there is no guarantee that the sequence of measures converges to a fixed point which is the barycenter. 
%Empirically, convergence to the barycenter appears to happen in our experiments.
Identifying the precise conditions on the input measures and the initial point is an important future direction. 
%Our algorithm is time and memory consuming but it is explained by the hardness of the barycenter problem itself. 
Besides, our algorithm does not recover inverse OT maps $T_{\mathbb{P}_{n}\rightarrow\overline{\mathbb{P}}}$; we compute them with an OT solver as a follow-up. To avoid this step, one may consider using invertible neural nets \citep{etmann2020iunets} to parametrize maps $T_{\theta_{n}}$ in our Algorithm \ref{algorithm-win}.
% However, it should be taken into account that they may be less expressive than 
%Studying this question is a promising avenue for the future work.
% One may extract them from dual potentials as $x-\nabla v_{\omega_{n}}(x)$, but such strategy might suffer from the gradient deviation \citep[\wasyparagraph 2]{korotin2021neural}.
% These statements serve as the challenge for our future research.
%  or to manually inverse maps $T_{\theta_{n}}$, e.g. by minimizing $\argmin_{x}\|T_{\theta_{n}}(x)-y\|^{2}$

\vspace{-1mm}\textbf{Limitations (dataset).} To create Ave, celeba! dataset (\wasyparagraph\ref{sec-ave-celeba}), we compose ICNNs with decolorization, random reflections and permutations to simulate degraded images. 
It is unclear how to produce other practically interesting effects via ICNNs. It also remains an open question on how to better hide the information of the barycenter image in the constructed marginal measures.
Studying these questions
% how to generate more realistic barycenter benchmark 
is an interesting future direction that can inspire benchmarking other OT problems.
%Besides, our Ave, celeba! contains $3\times 67$K samples from $3$ measures whose $\mathbb{W}_{2}$ barycenter is $\mathbb{P}_{\text{Celeba}}$. The barycenter estimate from samples might be biased from $\mathbb{P}_{\text{Celeba}}$ pointing to the necessity to analyse statistical properties of our approach.

\vspace{-1mm}\textsc{Acknowledgements}. {\small E. Burnaev was supported by the Russian Foundation for Basic Research grant 21-51-12005 NNIO\_a. A portion of this project was funded by the Skolkovo Institute of Science and Technology as part of the Skoltech NGP Program and funds were received by the Massachusetts Institute of Technology prior to September 1, 2022.  Neither Mr. Li, nor any other MIT personnel, contributed to any substantive or artistic alteration or enhancement of this publication after August 31, 2022.}

\bibliographystyle{plain}
\bibliography{references}

\begin{thebibliography}{10}

\bibitem{agueh2011barycenters}
Martial Agueh and Guillaume Carlier.
\newblock Barycenters in the {W}asserstein space.
\newblock {\em SIAM Journal on Mathematical Analysis}, 43(2):904--924, 2011.

\bibitem{altschuler2021averaging}
Jason~M Altschuler, Sinho Chewi, Patrik Gerber, and Austin~J Stromme.
\newblock Averaging on the bures-wasserstein manifold: dimension-free
  convergence of gradient descent.
\newblock {\em arXiv preprint arXiv:2106.08502}, 2021.

\bibitem{alvarez2016fixed}
Pedro~C {\'A}lvarez-Esteban, E~Del~Barrio, JA~Cuesta-Albertos, and
  C~Matr{\'a}n.
\newblock A fixed-point approach to barycenters in {W}asserstein space.
\newblock {\em Journal of Mathematical Analysis and Applications},
  441(2):744--762, 2016.

\bibitem{amos2017input}
Brandon Amos, Lei Xu, and J~Zico Kolter.
\newblock Input convex neural networks.
\newblock In {\em Proceedings of the 34th International Conference on Machine
  Learning-Volume 70}, pages 146--155. JMLR. org, 2017.

\bibitem{arjovsky2017wasserstein}
Martin Arjovsky, Soumith Chintala, and L{\'e}on Bottou.
\newblock Wasserstein {GAN}.
\newblock {\em arXiv preprint arXiv:1701.07875}, 2017.

\bibitem{barannikov2021representation}
Serguei Barannikov, Ilya Trofimov, Nikita Balabin, and Evgeny Burnaev.
\newblock Representation topology divergence: A method for comparing neural
  network representations.
\newblock In Kamalika Chaudhuri, Stefanie Jegelka, Le~Song, Csaba Szepesvari,
  Gang Niu, and Sivan Sabato, editors, {\em Proceedings of the 39th
  International Conference on Machine Learning}, volume 162 of {\em Proceedings
  of Machine Learning Research}, pages 1607--1626. PMLR, 17--23 Jul 2022.

\bibitem{barannikov2021manifold}
Serguei Barannikov, Ilya Trofimov, Grigorii Sotnikov, Ekaterina Trimbach,
  Alexander Korotin, Alexander Filippov, and Evgeny Burnaev.
\newblock Manifold topology divergence: a framework for comparing data
  manifolds.
\newblock {\em Advances in Neural Information Processing Systems},
  34:7294--7305, 2021.

\bibitem{bespalov2021data}
Iaroslav Bespalov, Nazar Buzun, Oleg Kachan, and Dmitry~V Dylov.
\newblock Data augmentation with manifold barycenters.
\newblock {\em arXiv preprint arXiv:2104.00925}, 2021.

\bibitem{bigot2019data}
J{\'e}r{\'e}mie Bigot, Elsa Cazelles, and Nicolas Papadakis.
\newblock Data-driven regularization of wasserstein barycenters with an
  application to multivariate density registration.
\newblock {\em Information and Inference: A Journal of the IMA}, 8(4):719--755,
  2019.

\bibitem{bonneel2015sliced}
Nicolas Bonneel, Julien Rabin, Gabriel Peyr{\'e}, and Hanspeter Pfister.
\newblock Sliced and radon wasserstein barycenters of measures.
\newblock {\em Journal of Mathematical Imaging and Vision}, 51(1):22--45, 2015.

\bibitem{brenier1991polar}
Yann Brenier.
\newblock Polar factorization and monotone rearrangement of vector-valued
  functions.
\newblock {\em Communications on pure and applied mathematics}, 44(4):375--417,
  1991.

\bibitem{chen2019gradual}
Yucheng Chen, Matus Telgarsky, Chao Zhang, Bolton Bailey, Daniel Hsu, and Jian
  Peng.
\newblock A gradual, semi-discrete approach to generative network training via
  explicit {W}asserstein minimization.
\newblock In {\em International Conference on Machine Learning}, pages
  1071--1080. PMLR, 2019.

\bibitem{chewi2020gradient}
Sinho Chewi, Tyler Maunu, Philippe Rigollet, and Austin~J Stromme.
\newblock Gradient descent algorithms for bures-wasserstein barycenters.
\newblock In {\em Conference on Learning Theory}, pages 1276--1304. PMLR, 2020.

\bibitem{chi2021variational}
Jinjin Chi, Zhiyao Yang, Jihong Ouyang, and Ximing Li.
\newblock Variational wasserstein barycenters with c-cyclical monotonicity.
\newblock {\em arXiv preprint arXiv:2110.11707}, 2021.

\bibitem{colombo2021automatic}
Pierre Colombo, Guillaume Staerman, Chloe Clavel, and Pablo Piantanida.
\newblock Automatic text evaluation through the lens of wasserstein
  barycenters, 2021.

\bibitem{daaloul2021sampling}
Chiheb Daaloul, Thibaut~Le Gouic, Jacques Liandrat, and Magali Tournus.
\newblock Sampling from the wasserstein barycenter.
\newblock {\em arXiv preprint arXiv:2105.01706}, 2021.

\bibitem{dognin2019wasserstein}
Pierre Dognin, Igor Melnyk, Youssef Mroueh, Jerret Ross, Cicero~Dos Santos, and
  Tom Sercu.
\newblock Wasserstein barycenter model ensembling.
\newblock {\em arXiv preprint arXiv:1902.04999}, 2019.

\bibitem{etmann2020iunets}
Christian Etmann, Rihuan Ke, and Carola-Bibiane Sch{\"o}nlieb.
\newblock iunets: learnable invertible up-and downsampling for large-scale
  inverse problems.
\newblock In {\em 2020 IEEE 30th International Workshop on Machine Learning for
  Signal Processing (MLSP)}, pages 1--6. IEEE, 2020.

\bibitem{fan2020scalable}
Jiaojiao Fan, Amirhossein Taghvaei, and Yongxin Chen.
\newblock Scalable computations of {W}asserstein barycenter via input convex
  neural networks.
\newblock {\em arXiv preprint arXiv:2007.04462}, 2020.

\bibitem{fenchel1949conjugate}
Werner Fenchel.
\newblock On conjugate convex functions.
\newblock {\em Canadian Journal of Mathematics}, 1(1):73--77, 1949.

\bibitem{genevay2016stochastic}
Aude Genevay, Marco Cuturi, Gabriel Peyr{\'e}, and Francis Bach.
\newblock Stochastic optimization for large-scale optimal transport.
\newblock In {\em Advances in neural information processing systems}, pages
  3440--3448, 2016.

\bibitem{genevay2017gan}
Aude Genevay, Gabriel Peyr{\'e}, and Marco Cuturi.
\newblock Gan and vae from an optimal transport point of view.
\newblock {\em arXiv preprint arXiv:1706.01807}, 2017.

\bibitem{heusel2017gans}
Martin Heusel, Hubert Ramsauer, Thomas Unterthiner, Bernhard Nessler, and Sepp
  Hochreiter.
\newblock {GAN}s trained by a two time-scale update rule converge to a local
  nash equilibrium.
\newblock In {\em Advances in neural information processing systems}, pages
  6626--6637, 2017.

\bibitem{inouye2021iterative}
David~I Inouye, Zeyu Zhou, Ziyu Gong, and Pradeep Ravikumar.
\newblock Iterative barycenter flows.
\newblock {\em arXiv preprint arXiv:2104.07232}, 2021.

\bibitem{kantorovitch1958translocation}
Leonid Kantorovitch.
\newblock On the translocation of masses.
\newblock {\em Management Science}, 5(1):1--4, 1958.

\bibitem{kingma2014adam}
Diederik~P Kingma and Jimmy Ba.
\newblock Adam: A method for stochastic optimization.
\newblock {\em arXiv preprint arXiv:1412.6980}, 2014.

\bibitem{koldasbayeva2022large}
Diana Koldasbayeva, Polina Tregubova, Dmitrii Shadrin, Mikhail Gasanov, and
  Maria Pukalchik.
\newblock Large-scale forecasting of heracleum sosnowskyi habitat suitability
  under the climate change on publicly available data.
\newblock {\em Scientific reports}, 12(1):1--11, 2022.

\bibitem{korotin2019wasserstein}
Alexander Korotin, Vage Egiazarian, Arip Asadulaev, Alexander Safin, and Evgeny
  Burnaev.
\newblock Wasserstein-2 generative networks.
\newblock In {\em International Conference on Learning Representations}, 2021.

\bibitem{korotin2021neural}
Alexander Korotin, Lingxiao Li, Aude Genevay, Justin~M Solomon, Alexander
  Filippov, and Evgeny Burnaev.
\newblock Do neural optimal transport solvers work? a continuous wasserstein-2
  benchmark.
\newblock {\em Advances in Neural Information Processing Systems},
  34:14593--14605, 2021.

\bibitem{korotin2021continuous}
Alexander Korotin, Lingxiao Li, Justin Solomon, and Evgeny Burnaev.
\newblock Continuous wasserstein-2 barycenter estimation without minimax
  optimization.
\newblock In {\em International Conference on Learning Representations}, 2021.

\bibitem{korotin2021mixability}
Alexander Korotin, Vladimir V’yugin, and Evgeny Burnaev.
\newblock Mixability of integral losses: A key to efficient online aggregation
  of functional and probabilistic forecasts.
\newblock {\em Pattern Recognition}, 120:108175, 2021.

\bibitem{lacombe2021learning}
Julien Lacombe, Julie Digne, Nicolas Courty, and Nicolas Bonneel.
\newblock Learning to generate wasserstein barycenters.
\newblock {\em arXiv preprint arXiv:2102.12178}, 2021.

\bibitem{lecun-mnisthandwrittendigit-2010}
Yann LeCun and Corinna Cortes.
\newblock {MNIST} handwritten digit database.
\newblock 2010.

\bibitem{li2020continuous}
Lingxiao Li, Aude Genevay, Mikhail Yurochkin, and Justin Solomon.
\newblock Continuous regularized {W}asserstein barycenters.
\newblock {\em arXiv preprint arXiv:2008.12534}, 2020.

\bibitem{liu2019wasserstein}
Huidong Liu, Xianfeng Gu, and Dimitris Samaras.
\newblock Wasserstein {GAN} with quadratic transport cost.
\newblock In {\em Proceedings of the IEEE International Conference on Computer
  Vision}, pages 4832--4841, 2019.

\bibitem{liu2015faceattributes}
Ziwei Liu, Ping Luo, Xiaogang Wang, and Xiaoou Tang.
\newblock Deep learning face attributes in the wild.
\newblock In {\em Proceedings of International Conference on Computer Vision
  (ICCV)}, December 2015.

\bibitem{lu2020large}
Guansong Lu, Zhiming Zhou, Jian Shen, Cheng Chen, Weinan Zhang, and Yong Yu.
\newblock Large-scale optimal transport via adversarial training with
  cycle-consistency.
\newblock {\em arXiv preprint arXiv:2003.06635}, 2020.

\bibitem{lucic2018gans}
Mario Lucic, Karol Kurach, Marcin Michalski, Sylvain Gelly, and Olivier
  Bousquet.
\newblock Are {GAN}s created equal? a large-scale study.
\newblock In {\em Advances in neural information processing systems}, pages
  700--709, 2018.

\bibitem{lyu2021barycenteric}
Boyang Lyu, Thuan Nguyen, Prakash Ishwar, Matthias Scheutz, and Shuchin Aeron.
\newblock Barycenteric distribution alignment and manifold-restricted
  invertibility for domain generalization, 2021.

\bibitem{makkuva2019optimal}
Ashok~Vardhan Makkuva, Amirhossein Taghvaei, Sewoong Oh, and Jason~D Lee.
\newblock Optimal transport mapping via input convex neural networks.
\newblock {\em arXiv preprint arXiv:1908.10962}, 2019.

\bibitem{metelli2019propagating}
Alberto~Maria Metelli, Amarildo Likmeta, and Marcello Restelli.
\newblock Propagating uncertainty in reinforcement learning via wasserstein
  barycenters.
\newblock In {\em 33rd Conference on Neural Information Processing Systems,
  NeurIPS 2019}, pages 4335--4347. Curran Associates, Inc., 2019.

\bibitem{mokrov2021large}
Petr Mokrov, Alexander Korotin, Lingxiao Li, Aude Genevay, Justin~M Solomon,
  and Evgeny Burnaev.
\newblock Large-scale wasserstein gradient flows.
\newblock {\em Advances in Neural Information Processing Systems},
  34:15243--15256, 2021.

\bibitem{montesuma2021wasserstein}
Eduardo~Fernandes Montesuma and Fred Maurice~Ngole Mboula.
\newblock Wasserstein barycenter for multi-source domain adaptation.
\newblock In {\em Proceedings of the IEEE/CVF Conference on Computer Vision and
  Pattern Recognition}, pages 16785--16793, 2021.

\bibitem{mroueh2019wasserstein}
Youssef Mroueh.
\newblock Wasserstein style transfer.
\newblock {\em arXiv preprint arXiv:1905.12828}, 2019.

\bibitem{murecsan2017fruit}
Horea Mure{\c{s}}an and Mihai Oltean.
\newblock Fruit recognition from images using deep learning.
\newblock {\em arXiv preprint arXiv:1712.00580}, 2017.

\bibitem{nhan2019threeplayer}
Quan~Hoang Nhan~Dam, Trung Le, Tu~Dinh Nguyen, Hung Bui, and Dinh Phung.
\newblock Threeplayer {W}asserstein {GAN} via amortised duality.
\newblock In {\em Proc. of the 28th Int. Joint Conf. on Artificial Intelligence
  (IJCAI)}, 2019.

\bibitem{paris2021online}
Quentin Paris.
\newblock Online learning with exponential weights in metric spaces.
\newblock {\em arXiv preprint arXiv:2103.14389}, 2021.

\bibitem{peyre2019computational}
Gabriel Peyr{\'e}, Marco Cuturi, et~al.
\newblock Computational optimal transport.
\newblock {\em Foundations and Trends{\textregistered} in Machine Learning},
  11(5-6):355--607, 2019.

\bibitem{rabin2014adaptive}
Julien Rabin, Sira Ferradans, and Nicolas Papadakis.
\newblock Adaptive color transfer with relaxed optimal transport.
\newblock In {\em 2014 IEEE International Conference on Image Processing
  (ICIP)}, pages 4852--4856. IEEE, 2014.

\bibitem{rabin2011wasserstein}
Julien Rabin, Gabriel Peyr{\'e}, Julie Delon, and Marc Bernot.
\newblock Wasserstein barycenter and its application to texture mixing.
\newblock In {\em International Conference on Scale Space and Variational
  Methods in Computer Vision}, pages 435--446. Springer, 2011.

\bibitem{rockafellar1976integral}
R~Tyrrell Rockafellar.
\newblock Integral functionals, normal integrands and measurable selections.
\newblock In {\em Nonlinear operators and the calculus of variations}, pages
  157--207. Springer, 1976.

\bibitem{ronneberger2015u}
Olaf Ronneberger, Philipp Fischer, and Thomas Brox.
\newblock U-net: Convolutional networks for biomedical image segmentation.
\newblock In {\em International Conference on Medical image computing and
  computer-assisted intervention}, pages 234--241. Springer, 2015.

\bibitem{rout2021generative}
Litu Rout, Alexander Korotin, and Evgeny Burnaev.
\newblock Generative modeling with optimal transport maps.
\newblock In {\em International Conference on Learning Representations}, 2021.

\bibitem{santambrogio2015optimal}
Filippo Santambrogio.
\newblock Optimal transport for applied mathematicians.
\newblock {\em Birk{\"a}user, NY}, 55(58-63):94, 2015.

\bibitem{seguy2017large}
Vivien Seguy, Bharath~Bhushan Damodaran, R{\'e}mi Flamary, Nicolas Courty,
  Antoine Rolet, and Mathieu Blondel.
\newblock Large-scale optimal transport and mapping estimation.
\newblock {\em arXiv preprint arXiv:1711.02283}, 2017.

\bibitem{simon2020barycenters}
Dror Simon and Aviad Aberdam.
\newblock Barycenters of natural images constrained wasserstein barycenters for
  image morphing.
\newblock In {\em Proceedings of the IEEE/CVF Conference on Computer Vision and
  Pattern Recognition}, pages 7910--7919, 2020.

\bibitem{simonyan2014very}
Karen Simonyan and Andrew Zisserman.
\newblock Very deep convolutional networks for large-scale image recognition.
\newblock {\em arXiv preprint arXiv:1409.1556}, 2014.

\bibitem{solomon2015convolutional}
Justin Solomon, Fernando De~Goes, Gabriel Peyr{\'e}, Marco Cuturi, Adrian
  Butscher, Andy Nguyen, Tao Du, and Leonidas Guibas.
\newblock Convolutional {W}asserstein distances: Efficient optimal
  transportation on geometric domains.
\newblock {\em ACM Transactions on Graphics (TOG)}, 34(4):1--11, 2015.

\bibitem{srivastava2015wasp}
Sanvesh Srivastava, Volkan Cevher, Quoc Dinh, and David Dunson.
\newblock Wasp: Scalable bayes via barycenters of subset posteriors.
\newblock In {\em Artificial Intelligence and Statistics}, pages 912--920,
  2015.

\bibitem{srivastava2018scalable}
Sanvesh Srivastava, Cheng Li, and David~B Dunson.
\newblock Scalable bayes via barycenter in {W}asserstein space.
\newblock {\em The Journal of Machine Learning Research}, 19(1):312--346, 2018.

\bibitem{staudt2022uniqueness}
Thomas Staudt, Shayan Hundrieser, and Axel Munk.
\newblock On the uniqueness of kantorovich potentials.
\newblock {\em arXiv preprint arXiv:2201.08316}, 2022.

\bibitem{taghvaei20192}
Amirhossein Taghvaei and Amin Jalali.
\newblock 2-{W}asserstein approximation via restricted convex potentials with
  application to improved training for {GAN}s.
\newblock {\em arXiv preprint arXiv:1902.07197}, 2019.

\bibitem{vidal2019progressive}
Jules Vidal, Joseph Budin, and Julien Tierny.
\newblock Progressive wasserstein barycenters of persistence diagrams.
\newblock {\em IEEE transactions on visualization and computer graphics},
  26(1):151--161, 2019.

\bibitem{villani2003topics}
C{\'e}dric Villani.
\newblock {\em Topics in optimal transportation}.
\newblock Number~58. American Mathematical Soc., 2003.

\bibitem{villani2008optimal}
C{\'e}dric Villani.
\newblock {\em Optimal transport: old and new}, volume 338.
\newblock Springer Science \& Business Media, 2008.

\bibitem{xiao2017fashion}
Han Xiao, Kashif Rasul, and Roland Vollgraf.
\newblock Fashion-mnist: a novel image dataset for benchmarking machine
  learning algorithms.
\newblock {\em arXiv preprint arXiv:1708.07747}, 2017.

\bibitem{xie2019scalable}
Yujia Xie, Minshuo Chen, Haoming Jiang, Tuo Zhao, and Hongyuan Zha.
\newblock On scalable and efficient computation of large scale optimal
  transport.
\newblock volume~97 of {\em Proceedings of Machine Learning Research}, pages
  6882--6892, Long Beach, California, USA, 09--15 Jun 2019. PMLR.

\bibitem{yu2014fine}
Aron Yu and Kristen Grauman.
\newblock Fine-grained visual comparisons with local learning.
\newblock In {\em Proceedings of the IEEE Conference on Computer Vision and
  Pattern Recognition}, pages 192--199, 2014.

\end{thebibliography}

%%%%%%%%%%%%%%%%%%%%%%%%%%%%%%%%%%%%%%%%%%%%%%%%%%%%%%%%%%%%

\newpage
\section*{Checklist}
\begin{enumerate}

\item For all authors...
\begin{enumerate}
  \item Do the main claims made in the abstract and introduction accurately reflect the paper's contributions and scope? \newline\answerYes{See \wasyparagraph\ref{sec-intro}.}
  \item Did you describe the limitations of your work?
  \newline\answerYes{See \wasyparagraph\ref{sec-discussion}.}
  \item Did you discuss any potential negative societal impacts of your work?
\newline\answerNA{}
% The potential societal impact is the same as that of any related work in the field of generative modeling. Namely, the advancement in the field may affect some jobs, e.g., those which are related to image processing and design.
  \item Have you read the ethics review guidelines and ensured that your paper conforms to them?
    \answerYes{}
\end{enumerate}

\item If you are including theoretical results...
\begin{enumerate}
  \item Did you state the full set of assumptions of all theoretical results?
    \newline\answerYes{All the assumptions are stated in the main text.}
        \item Did you include complete proofs of all theoretical results?
    \newline\answerYes{All the proofs are given in the appendices.}
\end{enumerate}

\item If you ran experiments...
\begin{enumerate}
  \item Did you include the code, data, and instructions needed to reproduce the main experimental results (either in the supplemental material or as a URL)?
\newline\answerYes{The code and the instructions are included in the supplementary material. The datasets that we use are publicly available.}
  \item Did you specify all the training details (e.g., data splits, hyperparameters, how they were chosen)?
    \newline\answerYes{See \wasyparagraph\ref{sec-evaluation} and the supplementary material (appendices + code).}
    \item Did you report error bars (e.g., with respect to the random seed after running experiments multiple times)?
    \newline\answerNo{Due to the well-known high computational complexity of learning generative models, most experiments (both with our method and alternatives) were conducted only once.}
    \item Did you include the total amount of compute and the type of resources used (e.g., type of GPUs, internal cluster, or cloud provider)?
    \newline\answerYes{See the discussion in \wasyparagraph\ref{sec-evaluation} and Appendices.}
\end{enumerate}

\item If you are using existing assets (e.g., code, data, models) or curating/releasing new assets...
\begin{enumerate}
  \item If your work uses existing assets, did you cite the creators?
  \newline\answerYes{See the discussion in section \wasyparagraph\ref{sec-evaluation}}.
  \item Did you mention the license of the assets?
    \newline\answerNo{We refer to the datasets' public pages.}
  \item Did you include any new assets either in the supplemental material or as a URL?
  \newline\answerYes{See the supplementary material}
  \item Did you discuss whether and how consent was obtained from people whose data you're using/curating?
  \newline\answerNA{}
  \item Did you discuss whether the data you are using/curating contains personally identifiable information or offensive content?
  \newline\answerNo{For CelebA faces dataset, we refer to the original authors publication.}
\end{enumerate}

\item If you used crowdsourcing or conducted research with human subjects...
\begin{enumerate}
  \item Did you include the full text of instructions given to participants and screenshots, if applicable?
  \newline\answerNA{}
  \item Did you describe any potential participant risks, with links to Institutional Review Board (IRB) approvals, if applicable?
  \newline\answerNA{}
  \item Did you include the estimated hourly wage paid to participants and the total amount spent on participant compensation?
  \newline\answerNA{}
\end{enumerate}

\end{enumerate}

%%%%%%%%%%%%%%%%%%%%%%%%%%%%%%%%%%%%%%%%%%%%%%%%%%%%%%%%%%%%

\newpage
\appendix

\section{Proofs}
\label{sec-proofs}
First, we recall basic properties of convex conjugate functions that we rely on in our proofs. Let ${\psi:\mathbb{R}^{D}\rightarrow\mathbb{R}}$ be a convex function and $\overline{\psi}$ be its convex conjugate. From the definition of $\overline{\psi}$, we obtain
$$\psi(x)+\overline{\psi}(y)\geq \langle x,y\rangle$$
for all $x,y\in\mathbb{R}^{D}.$
Assume that $\psi$ is differentiable and has an invertible gradient $\nabla\psi:\mathbb{R}^{D}\rightarrow\mathbb{R}^{D}$. The latter condition holds, e.g., for strongly convex functions. From the convexity of $\psi$, we derive
$${x=\argmax_{x\in\mathbb{R}^{D}}[\langle x,y\rangle - \psi(x)]\Leftrightarrow y=\nabla \psi(x)\Leftrightarrow x=(\nabla \psi)^{-1}(y)},$$
which yields
$${\overline{\psi}(y)=\langle (\nabla \psi)^{-1}(x), x\rangle-\psi\big((\nabla \psi)^{-1}(x))}.$$
In particular, the strict equality $\psi(x)+\overline{\psi}(y)= \langle x,y\rangle$ holds if and only if $y=\nabla \psi(x)$. By applying the same logic to $\overline{\psi}$, we obtain $(\nabla \overline{\psi})^{-1}=\nabla \psi$ and $(\nabla \psi)^{-1}=\nabla \overline{\psi}$, i.e., the gradients of conjugate functions are mutually inverse.

\subsection{Proof of Lemma 1}
\begin{proof} For each $n=1,2,\dots,N$ we perform the following evaluation:
\begin{eqnarray}
\frac{\partial}{\partial\xi}\mathbb{W}_{2}^{2}(G_{\xi}\sharp\mathbb{S},\mathbb{P}_{n})=\int_{z} \mathbf{J}_{\xi} G_\xi(z)^T\nabla u_{n}^{*}\big(G_{\xi}(z)\big)d\mathbb{S}(z),
\label{grad-summand-semifinal}
\\
\int_{z} \mathbf{J}_{\xi} G_\xi(z)^T\big(G_{\xi}(z)-T_{\mathbb{P}_{\xi}\rightarrow\mathbb{P}_{n}}\big(G_{\xi}(z)\big)\bigg)d\mathbb{S}(z),
\label{grad-summand-final}
\end{eqnarray}
where $u_{n}^{*}$ is the optimal dual potential for $\mathbb{P}_{\xi}=G_{\xi}\sharp\mathbb{S}$ and $\mathbb{P}_{n}$. Equation \eqref{grad-summand-semifinal} follows from \citep[Equation 3]{genevay2017gan}. Equation \eqref{grad-summand-final} follows from the property $\nabla u_{n}^{*}(x)=x-T_{\mathbb{P}_{\xi}\rightarrow\mathbb{P}_{n}}(x)$ connecting dual potentials and OT maps.
% \begin{eqnarray}
% \frac{\partial}{\partial\xi}\mathbb{W}_{2}^{2}(G_{\xi}\sharp\mathbb{S},\mathbb{P}_{n})=
% \frac{\partial}{\partial\xi}\min_{T: T\sharp\mathbb{P}_{\xi}=\mathbb{P}_{n}}\int \frac{1}{2}\|x-T(x)\|^{2}d\mathbb{P}_{\xi}(x)=
% \label{envelope-w-grad}
% \\
% \frac{\partial}{\partial\xi}\int \frac{1}{2}\|x-\texttt{StopGrad}\big(T_{\mathbb{P}_{\xi_0}\rightarrow\mathbb{P}_{n}}(x)\big)\|^{2}d\mathbb{P}_{\xi}(x)=
% \nonumber
% \\
% \frac{\partial}{\partial\xi}\int \frac{1}{2}\|x-T_{\mathbb{P}_{\xi_0}\rightarrow\mathbb{P}_{n}}(x)\|^{2}d\big[G_{\xi}\sharp\mathbb{S}\big](x)=
% \label{envelope-w-grad-used}
% \\
% \frac{\partial}{\partial\xi}\int \frac{1}{2}\|G_{\xi}(z)-T_{\mathbb{P}_{\xi_0}\rightarrow\mathbb{P}_{n}}\big(G_{\xi}(z)\big)\|^{2}d\mathbb{S}(z)=
% \label{envelope-w-grad-done}
% \\
% \int_{z} \mathbf{J}_{\xi} G_\xi(z)^T\big(G_{\xi}(z)-T_{\mathbb{P}_{\xi_0}\rightarrow\mathbb{P}_{n}}\big(G_{\xi}(z)\big)d\mathbb{S}(z),
% \label{grad-summand-final}
% \end{eqnarray}
% In \eqref{envelope-w-grad}, we use the primal form \eqref{ot-primal-form-monge} of $\mathbb{W}_{2}$. In transition to \eqref{envelope-w-grad-used}, we use the envelope theorem \citep{milgrom2002envelope} and substitute the OT map $T_{\mathbb{P}_{\xi_0}\rightarrow\mathbb{P}_{n}}$.
% In \eqref{envelope-w-grad-done}, we apply the change of variables for $x=G_{\xi}(z)$. In \eqref{grad-summand-final}, we use $\mathbf{J}_{\xi} G_\xi(z)^T$ to denote the transpose of the Jacobian matrix of $G_\xi(z)$ w.r.t.\ parameters $\xi$. 
% \lingxiao{(16) seems to miss a derivative of T due to chain rule?}
We sum \eqref{grad-summand-final} for $n=1,\dots,N$ w.r.t. weights $\alpha_{n}$ with $\xi=\xi_{0}$ and obtain
\begin{eqnarray}\frac{\partial}{\partial\xi}\sum_{n=1}^{N}\alpha_{n}\mathbb{W}_{2}^{2}(G_{\xi}\sharp\mathbb{S},\mathbb{P}_{n})=
\int_{z} \mathbf{J}_{\xi} G_{\xi_0}(z)^T\bigg(G_{\xi_0}(z)-\sum_{n=1}^{N}\alpha_{n}T_{\mathbb{P}_{\xi_0}\rightarrow\mathbb{P}_{n}}\big(G_{\xi_0}(z)\big)\bigg)d\mathbb{S}(z).
\label{right-hand-side-derivative}\end{eqnarray}
Note that \eqref{right-hand-side-derivative} exactly matches the derivative of the left-hand side of \eqref{equivalence-grads} evaluated at $\xi=\xi_0$.
\end{proof}

\subsection{Proof of Lemma 2}

\begin{proof}
First, we prove the congruence, i.e., $\beta\psi^{l}(x)+(1-\beta)\psi^{r}(x)=\frac{\|x\|^{2}}{2}$ for all $x\in\mathbb{R}^{D}$.
\begin{eqnarray}\beta\psi^{l}(x)+(1-\beta)\psi^{r}(x)=
\nonumber
\\
\beta\max_{y_{1}\in\mathbb{R}^{D}}\big[\langle x,y_{1}\rangle-\overline{\psi^{l}}(y_{1})\big]+(1-\beta)\max_{y_{2}\in\mathbb{R}^{D}}\big[\langle x,y_{2}\rangle-\overline{\psi^{r}}(y_{2})\big]=
\label{cong-sum}
\\
\beta\max_{y_{1}\in\mathbb{R}^{D}}\big[\langle x,y_{1}\rangle-\beta\frac{\|y_{1}\|^{2}}{2}-(1-\beta)\psi(y_{1})\big]+
\nonumber
\\
(1-\beta)\max_{y_{2}\in\mathbb{R}^{D}}\big[\langle x,y_{2}\rangle-(1-\beta)\frac{\|y_{2}\|^{2}}{2}-\beta\overline{\psi}(x)\big]=
\nonumber
\\
\max_{y_{1},y_{2}\in\mathbb{R}^{D}}\big[\langle x,\beta y_{1}+(1-\beta)y_{2}\rangle-\beta^2\frac{\|y_{1}\|^{2}}{2}-(1-\beta)^2\frac{\|y_{2}\|^{2}}{2}-\beta(1-\beta)(\psi(y_{1})+\overline{\psi}(y_{2}))\big]\leq
\label{conj-upper-bound}
\\
\max_{y_{1},y_{2}\in\mathbb{R}^{D}}\big[\langle x,\beta y_{1}+(1-\beta)y_{2}\rangle-\beta^2\frac{\|y_{1}\|^{2}}{2}-(1-\beta)^2\frac{\|y_{2}\|^{2}}{2}-\beta(1-\beta)\langle y_{1},y_{2}\rangle\big]=
\nonumber
\\
\max_{y_{1},y_{2}\in\mathbb{R}^{D}}\frac{\|x\|^{2}}{2}-\frac{1}{2}\|x-(\beta y_{1}+(1-\beta)y_{2})\|^{2}\leq\frac{\|x\|^{2}}{2}.
\label{cong-final-bound}
\end{eqnarray}
First, we substitute $(y_{1},y_{2})=(y^{l},\nabla \psi(y^{l}))$. For this pair, $x=\nabla\overline{\psi^{l}}(y^{l})=\beta y^{l}+(1-\beta)\nabla\psi(y^{l})$, which results in $x=\beta y_{1}+(1-\beta)y_{2}$. Moreover, since $y_{2}=\nabla \psi(y_{1})$, we have $\psi(y_{1})+\overline{\psi}(y_{2})=\langle y_{1},y_{2}\rangle$.
As the consequence, both inequalities \eqref{conj-upper-bound} and \eqref{cong-final-bound} turn to strict equalities yielding congruence of $\psi^{l},\psi^{r}$. From \eqref{left-right-def}, the smoothness and strong convexity of $\psi$ imply that $\psi^{l}$ and $\psi^{r}$ are smooth. Consequently, $\overline{\psi^{l}}$ and $\overline{\psi^{l}}$ are strongly convex. Thus, the maximizer of \eqref{cong-sum} is unique. We know the maximum of \eqref{cong-sum} is attained at $(y_{1},y_{2})=(\nabla\psi^{l}(x),\nabla\psi^{r}(x))=(y^{l},y^{r})$. We conclude $(y^{l},y^{r})=(y^{l},\nabla \psi(y^{l}))$, i.e., $y^{r}=\nabla \psi(y^{l})$. Finally, $y^{l}=\nabla\psi^{l}(x)\Leftrightarrow x=\nabla\overline{\psi^{l}}(y^{l})\Leftrightarrow y^{l}=\max\limits_{y\in\mathbb{R}^{D}}\big[\langle x,y\rangle-\overline{\psi^{l}}(y)\big]$, which matches \eqref{y-left-opt}.
\end{proof}

\subsection{Proof of Lemma 3}
\begin{proof}First, we check that $\sum_{n=1}^{N}\alpha_{n}$ indeed equals 1:
\begin{eqnarray}
\sum_{n=1}^{N}\alpha_{n}=\sum_{n=1}^{N}\sum_{m=1}^{M}w_{m}\big[\beta_{m}\gamma^{l}_{nm}+(1-\beta_{m})\gamma^{r}_{nm}\big]=
\nonumber
\\
\sum_{m=1}^{M}\big[w_{m}\beta_{m}\underbrace{\sum_{n=1}^{N}\gamma^{l}_{nm}}_{=1}\big]+\sum_{m=1}^{M}\big[w_{m}(1-\beta_{m})\underbrace{\sum_{n=1}^{N}\gamma^{r}_{nm}}_{=1}\big]=
\nonumber
\\
\sum_{m=1}^{M}w_{m}\beta_{m}+\sum_{m=1}^{M}w_{m}(1-\beta_{m})=
\sum_{m=1}^{M}w_{m}\big(\beta_{m}+(1-\beta_{m})\big)=\sum_{m=1}^{M}w_{m}=1.
\end{eqnarray}
Next, we check that $\psi_{1},\dots,\psi_{N}$ are congruent w.r.t. weights $\alpha_{1},\dots,\alpha_{N}$:
\begin{eqnarray}
\sum_{n=1}^{N}\alpha_{n}\psi_{n}(x)=\sum_{n=1}^{N}\sum_{m=1}^{M}w_{m}\big[\beta_{m}\gamma^{l}_{nm}\cdot \psi_{m}^{l}(x)+(1-\beta_{m})\gamma^{r}_{nm}\cdot \psi_{m}^{r}(x)\big]=
\nonumber
\\
\sum_{m=1}^{M}\big[w_{m}\beta_{m}\psi_{m}^{l}(x)\underbrace{\sum_{n=1}^{N}\gamma^{l}_{nm}}_{=1}\big]+\sum_{m=1}^{M}\big[w_{m}(1-\beta_{m})\psi_{m}^{r}(x)\underbrace{\sum_{n=1}^{N}\gamma^{r}_{nm}}_{=1}\big]=
\nonumber
\\
\sum_{m=1}^{M}\big[w_{m}\underbrace{\big(\beta_{m}\psi_{m}^{l}(x)+(1-\beta_{m})\psi^{r}(x)\big]}_{=\frac{\|x\|^{2}}{2}}=\sum_{m=1}^{M}w_{m}\frac{\|x\|^{2}}{2}=\frac{\|x\|^{2}}{2}.
\nonumber
\end{eqnarray}
\end{proof}

\section{Experimental Details}
\label{sec-exp-details}

\subsection{Ave, celeba! Dataset Creation}
\label{sec-ave-celeba-creation}
The initialization of random permutations $\sigma_{m}$ and reflections $s_{m}$ (for $m=1,2$) as well as the random split of CelebA dataset into 3 parts (each containing $\approx 67K$ images) are \textit{hardcoded} in our provided script for producing Ave, celeba! dataset. To initialize ICNN$_{m}$ (for $m=1,2$), we use use ConvICNN64 \citep[Appendix B.1]{korotin2021neural} checkpoints \texttt{Early\_v1\_conj.pt}, \texttt{Early\_v2\_conj.pt} from the official Wasserstein-2 benchmark repository\footnote{\url{https://github.com/iamalexkorotin/Wasserstein2Benchmark}}.

We rescale Celeba images to $64\times 64$ by using \texttt{imresize} from  \texttt{scipy.misc}. To create empirical samples from input distributions $\mathbb{P}_{n}$ by using the rescaled CelebA dataset, we compute the gradient maps $\nabla\psi_{n}(x)$ ($n=1,2,3$) in Lemma \ref{lemma-cong-n-tuple} for images $x$ in the CelebA dataset. This computation implies computing gradient maps $\nabla\psi_{m}^{l}(x)$ and $\nabla\psi_{m}^{r}(x)$ for each base function $\psi_{m}^{0}$ ($m=1,2$) and summing them with respective coefficients \eqref{conv-comb-cong}. Following our Lemma \ref{lemma-cong-pair}, we compute $y^{l}_{m}\stackrel{\text{def}}{=}\nabla\psi_{m}^{l}(x)$ by solving a concave  optimization problem \eqref{y-left-opt} over the space of images. We solve this problem with the gradient descent. We use Adam optimizer \citep{kingma2014adam} with default betas,  $lr=2\cdot 10^{-2}$ and do $1000$ gradient steps. To speed up the computation, we simultaneously solve the problem for a batch of $256$ images $x$ from CelebA dataset. Then we compute $y^{r}\stackrel{\text{def}}{=}\nabla\psi_{m}^{r}(x)$ as $y^{r}=\nabla\psi_{m}(y^{l})$ (Lemma \ref{lemma-cong-pair}).

\textbf{Computational complexity.} Producing Ave, celeba! takes about $1,5$ days on a GPU GTX 1080 ti.

% \textbf{URL to download Ave, Celeba! dataset} (\textit{anonymous} Google Drive):\\ {\small\url{https://drive.google.com/file/d/1hJ83ZXcLbQMQund8OrCDJxWxWDCpTq-n/view?usp=sharing}}

\subsection{Hyperparameters (Algorithm \ref{algorithm-win}, Main Training)}
\label{sec-main-algorithm-appendix}

We provide the hyperparameters of all the experiments with algorithm \ref{algorithm-win} in Table \ref{table-params}. The column \textbf{total iters} shows the sum of gradient steps over generator $G_{\xi}$ and each of $N$ potentials $v_{\omega_{n}}$ in OT solvers.

\textbf{Optimization.} We use Adam optimizer with the default betas. During training, we decrease the learning rates of the generator $G_{\xi}$ and each potential $v_{\omega_{n}}$ every 10K steps of their optimizers. In the Gaussian case, we use a single GPU GTX 1080ti. In all other cases we split the batch over 4$\times$GPU GTX 1080ti (\texttt{nn.DataParallel} in PyTorch).

\textbf{Neural Network Architectures.} In the Gaussian case, we use In the evaluation in the Gaussian case, we use sequential fully-connected neural networks with ReLU activations for the generator $G_{\xi}:\mathbb{R}^{D}\rightarrow\mathbb{R}^{D}$, potentials $v_{\omega_{n}}:\mathbb{R}^{D}\rightarrow\mathbb{R}$ and transport maps ${T_{\theta_{n}}:\mathbb{R}^{D}\rightarrow\mathbb{R}^{D}}$. For all the networks the sizes of hidden layers are:
$$[\max(100, 2D), \max(100, 2D), \max(100, 2D)].$$
Working with images, we use the ResNet\footnote{\url{https://github.com/harryliew/WGAN-QC}} generator and discriminator architectures of WGAN-QC \citep{liu2019wasserstein} for our generator $G_{\xi}$ and potentials $v_{\omega_{n}}$ respectively. As the maps $T_{\theta_{n}}$, we use U-Net \footnote{\url{https://github.com/milesial/Pytorch-UNet}} \citep{ronneberger2015u}.

\textbf{Generator regression loss.} In the Gaussian case and experiments with grayscale images (MNIST, FashionMNIST), we use mean squared loss for generator regression. In other experiments, we use the perceptual mean squared loss based on the features of the pre-trained VGG-16 network \citep{simonyan2014very}. The loss is hardcoded in the implementation.

\textbf{Data pre-processing.} In all experiments with images we normalize them to $[-1,1]$. We rescale MNIST and FashionMNIST images to $32\times 32$. In all other cases, we rescale images to $64\times 64$. Note that Fruit360 dataset originally contains $114\times 114$ images; before rescaling, we add white color padding to make the images have the size $128\times 128$. Working with Ave, celeba! dataset, we additionally shift each subset $\mathbb{P}_{n}$ by $(\overline{\mu}-\mu_{n})$, i.e., we train the models on the $\lceil \text{CS}\rfloor$ baseline. This helps the models to avoid learning the shift.

\textbf{Computational complexity}. The most challenging experiments (Ave, celeba! and Handbags, Shoes, Fruit) take about 2-3 days to converge on $4\times$GPU GTX 1080 ti. Other experiments converge faster.

\begin{table}[!h]
% \centering
\tiny
\hspace{-11mm}\begin{tabular}{|c|c|c|c|c|c|c|c|c|c|c|c|c|c|c|c|}\hline
\textbf{Experiment} & D & H & $N$ & $G_{\xi}$ & $v_{\omega_{n}}$ & $T_{\theta_{n}}$ & $k_{G}$ & $k_{v}$ & $k_{T}$ & $lr_{G}$ & $lr_{v}$ & $lr_{T}$ & $\ell$ & \makecell{\textbf{Total}\\\textbf{iters}} & \makecell{\textbf{Batch}\\\textbf{size}} \\ \hline
Toy 2D  & 2 & 2 & 3 & \multicolumn{2}{|c|}{MLP}  & MLP  & \multirow{7}{*}{50} & \multirow{8}{*}{50} & 10 & $1\cdot 10^{-4}$ & $1\cdot 10^{-3}$ & $1\cdot 10^{-3}$ & \multirow{4}{*}{MSE} & 12K & 1024\\ \cline{1-4}
Gaussians  & 2-128 & 2-128 & 4 & \multicolumn{2}{|c|}{MLP}  & MLP  & & & 10 & $1\cdot 10^{-4}$ & $1\cdot 10^{-3}$ & $1\cdot 10^{-3}$ &  & 12K & 1024\\ \cline{0-6}\cline{16-16}
MNIST 0/1 & \multirow{2}{*}{1024} & \multirow{2}{*}{16} & 2 & \multicolumn{2}{|c|}{\multirow{6}{*}{ResNet}}   & \multirow{6}{*}{UNet} & & & 15 & $1\cdot 10^{-4}$ & $1\cdot 10^{-4}$ & $1\cdot 10^{-4}$ & & 60K & \multirow{6}{*}{64}\\ \cline{0-0}
FashionMNIST &  &  & 10 & \multicolumn{2}{|c|}{} & & & & 10 & $1\cdot 10^{-4}$ & $1\cdot 10^{-4}$ & $1\cdot 10^{-4}$ & & 100K & \\ \cline{0-0}\cline{1-3}\cline{14-14}
Bags, Shoes, Fruit & \multirow{4}{*}{12288} & \multirow{4}{*}{128}  & 3 & \multicolumn{2}{|c|}{} & & & & 10 & $3\cdot 10^{-4}$ & $3\cdot 10^{-4}$ & $3\cdot 10^{-4}$ & $\multirow{4}{*}{VGG}$ & 36K & \\ \cline{0-0}
Ave, celeba! &  &  & 3 & \multicolumn{2}{|c|}{} & & & & 10 & $3\cdot 10^{-4}$ & $3\cdot 10^{-4}$ & $3\cdot 10^{-4}$ & & 60K & \\ \cline{0-0}
Celeba &  &   & 1 & \multicolumn{2}{|c|}{} & & & & 15 & $1\cdot 10^{-4}$ & $1\cdot 10^{-4}$ & $1\cdot 10^{-4}$ & & 80K & \\ \cline{0-0}\cline{8-8}
Celeba (fixed $G$) &  &  & 1 & \multicolumn{2}{|c|}{} & & 0 & & 15 & $1\cdot 10^{-4}$ & $1\cdot 10^{-4}$ & $1\cdot 10^{-4}$ & & 120K & \\
\hline
\end{tabular}
\vspace{2mm}
\caption{Hyperparameters that we use in the experiments with our algorithm \ref{algorithm-win}.}
\label{table-params}
\end{table}

\subsection{Hyperparameters (Algorithm \ref{algorithm-win-inverse}, Learning Maps to the Barycenter)}

After using the main algorithm \ref{algorithm-win} to train $G_{\xi}$, we use algorithm \ref{algorithm-win-inverse} to extract the inverse optimal maps $\mathbb{P}_{n}\rightarrow\mathbb{P}_{\xi}$. We detail the hyperparameters in Table \ref{table-params-2} below. In all the cases we use Adam optimizer with the default betas. The column \textbf{total iters} show the number of update steps for each $v^{\text{inv}}_{\omega_{n}'}$.

\begin{table}[!h]
\centering
\scriptsize
\begin{tabular}{|c|c|c|c|c|c|c|c|c|c|}\hline
\textbf{Experiment} & D & $N$ & $v_{\omega_{n}}$ & $T_{\theta_{n}}$ & $k_{T}$ & $lr_{v}$ & $lr_{T}$ & \makecell{\textbf{Total}\\\textbf{iters}} & \makecell{\textbf{Batch}\\\textbf{size}} \\ \hline
Toy 2D & 2 & 2 & MLP & MLP & 10 & $1\cdot 10^{-3}$ & $1\cdot 10^{-3}$ & 10k & 1024\\\hline
MNIST 0/1 & \multirow{2}{*}{1024} & 2 & \multirow{4}{*}{ResNet} & \multirow{4}{*}{UNet} & \multirow{4}{*}{10} & \multirow{4}{*}{$1\cdot 10^{-4}$} & \multirow{4}{*}{$1\cdot 10^{-4}$} & 4k & \multirow{4}{*}{64}\\ \cline{0-0}
FashionMNIST  &  & 10  & & & & & & 4k & \\ \cline{0-1}
Bags, Shoes, Fruit & \multirow{2}{*}{12288} & 3   & & & & & & 20K & \\\cline{0-0}
Ave, celeba! &  & 3  & & & & & & 12K & \\\hline
\end{tabular}
\vspace{1mm}
\caption{Hyperparameters that we use in the experiments with algorithm \ref{algorithm-win-inverse}}
\label{table-params-2}
\end{table}

\begin{algorithm}[t!]
\SetInd{0.5em}{0.3em}
    {
        \SetAlgorithmName{Algorithm}{empty}{Empty}
        \SetKwInOut{Input}{Input}
        \SetKwInOut{Output}{Output}
        \Input{latent $\mathbb{S}$ and input
        $\mathbb{P}_{1},\dots,\mathbb{P}_{N}$ measures;\ pretrained generator $G_{\xi}:\mathbb{R}^{H}\rightarrow\mathbb{R}^{D}$ satisfying $G_{\xi}\sharp\mathbb{S}\approx\overline{\mathbb{P}}$;\\
        mapping networks $T^{\text{inv}}_{\theta_{1}'},\dots,T^{\text{inv}}_{\theta_{N}'}:\mathbb{R}^{D}\rightarrow\mathbb{R}^{D}$; potentials $v^{\text{inv}}_{\omega_{1}'},\dots,v_{\omega_{N}'}^{\text{inv}}:\mathbb{R}^{D}\rightarrow\mathbb{R}$;\\
        number of inner iterations for training transport maps: $K_{T}$;\\
        }
        \Output{OT maps satisfying $T_{\theta_{n}'}^{\text{inv}}\sharp \mathbb{P}_{n}\approx \mathbb{P}_{\xi}=(G_{\xi}\sharp\mathbb{S})\approx \overline{\mathbb{P}}$\;
        }
        
        \Repeat{not converged}{
            \For{$n = 1,2, \dots, N$}{
                Sample batches $Z\!\sim\! \mathbb{S}$, $Y\!\sim\! \mathbb{P}_{n}$; $X\!\leftarrow\! G_{\xi}(Z)$\;
                $\mathcal{L}_{v}\leftarrow \frac{1}{|Y|}\sum\limits_{y\in Y}v^{\text{inv}}_{\omega_{n}'}\big(T^{\text{inv}}_{\theta_{n}'}(y)\big)-\frac{1}{|X|}\sum\limits_{x\in X}v^{\text{inv}}_{\omega_{n}'}\big(x\big)$\;
                Update $\omega_{n}'$ by using $\frac{\partial \mathcal{L}_{v}}{\partial \omega_{n}'}$\;
                
                \For{$k_{T} = 1,2, \dots, K_{T}$}{
                    Sample batch $Y\!\sim\! \mathbb{P}_{n}$\;
                    ${\mathcal{L}_{T}\!=\!\frac{1}{|Y|}\!\sum\limits_{y\in Y}\!\big[\frac{1}{2}\|y\!-\!T^{\text{inv}}_{\theta_{n}'}(y)\|^{2}\!-\!v^{\text{inv}}_{\omega_{n}'}\!\big(T^{\text{inv}}_{\theta_{n}'}(y)\big)\!\big]}$\;
                Update $\theta_{n}'$ by using $\frac{\partial \mathcal{L}_{T}}{\partial \theta_{n}'}$\;
                }
            }
        }
        
        \caption{Learning maps from input measures to the learned barycenter $\mathbb{P}_{\xi}\approx\overline{\mathbb{P}}$ with $ \lceil\text{MM:R}\rceil$ OT solver.}
        \label{algorithm-win-inverse}
    }
\end{algorithm}

\subsection{Hyperparameters of competitive [SC$\mathbb{W}_{2}$B] algorithm}
\label{sec-scwb-params}

\begin{figure*}[!h]
\begin{subfigure}{0.3\linewidth}
\centering
\includegraphics[width=0.97\linewidth]{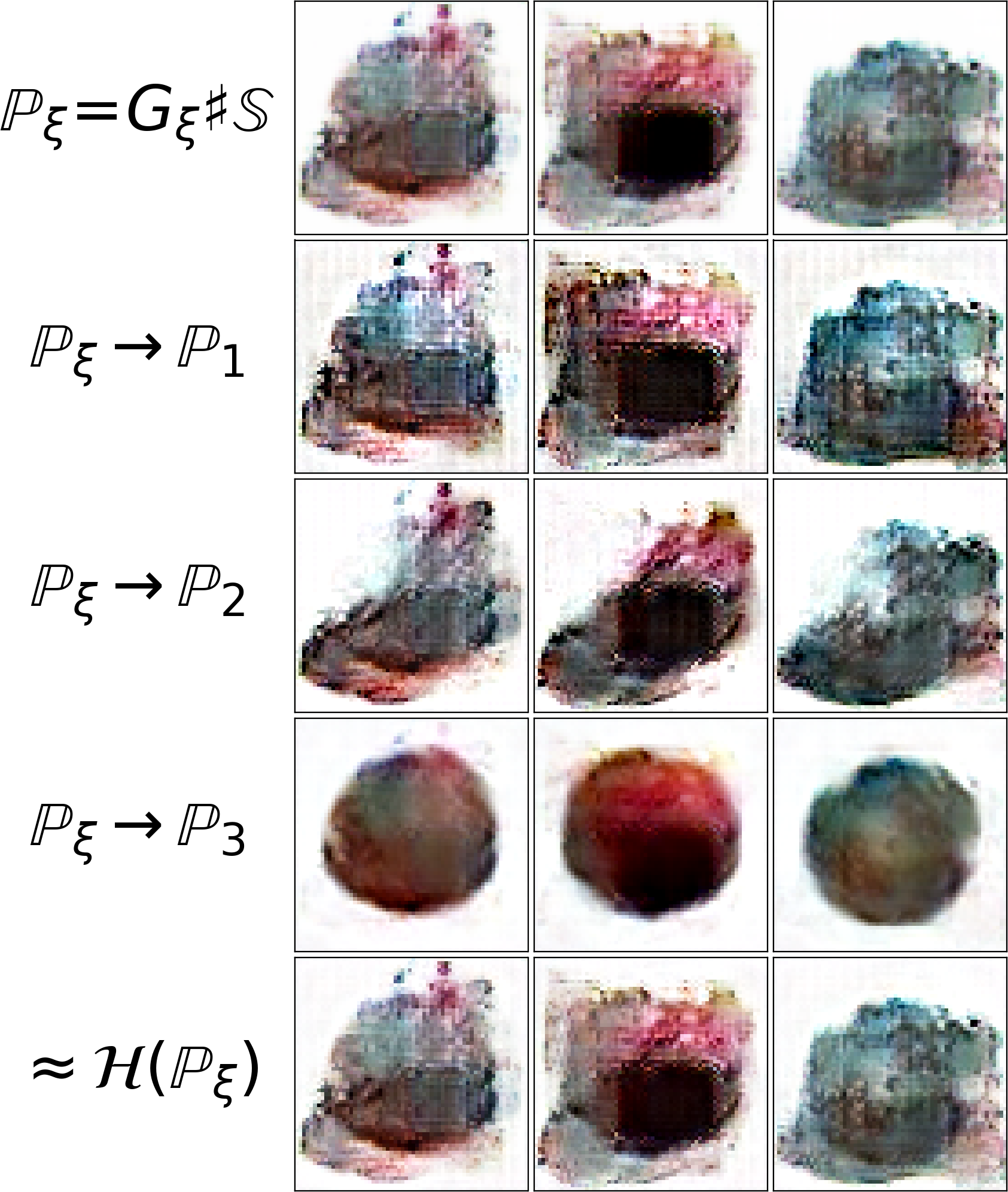}\caption{\centering Generated samples $\mathbb{P}_{\xi}\approx\overline{\mathbb{P}}$, fitted maps to each $\mathbb{P}_{n}$ and their average.}
\label{fig:fruit-generated-icnn}
\end{subfigure}
\begin{subfigure}{0.225\linewidth}
\centering
\includegraphics[width=0.97\linewidth]{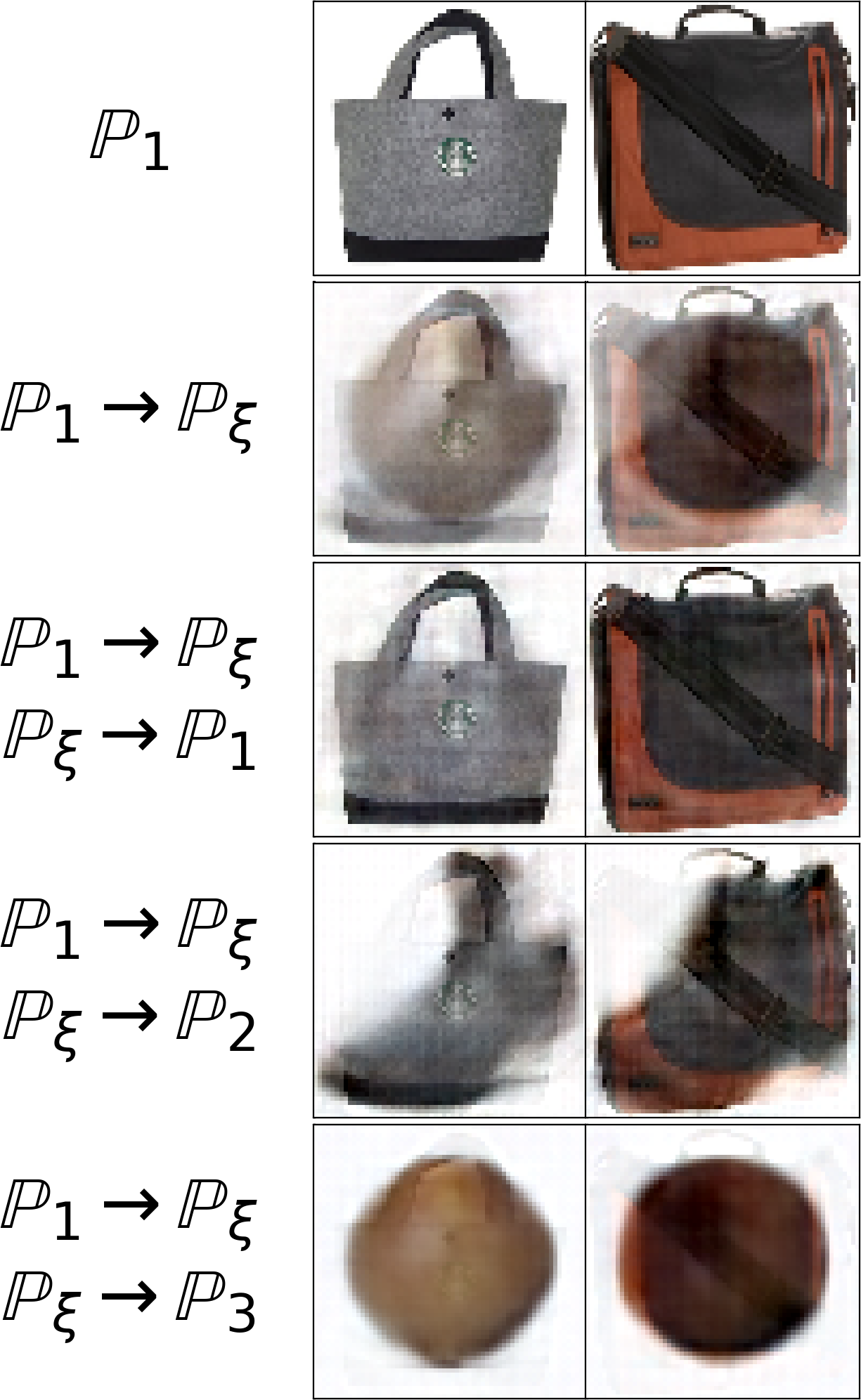}
\caption{\centering Samples $y\sim\mathbb{P}_{1}$ mapped through $\mathbb{P}_{\xi}$ to each $\mathbb{P}_{n}$.}
\label{fig:fruit-through1-icnn}
\end{subfigure}
\begin{subfigure}{0.225\linewidth}
\centering
\includegraphics[width=0.97\linewidth]{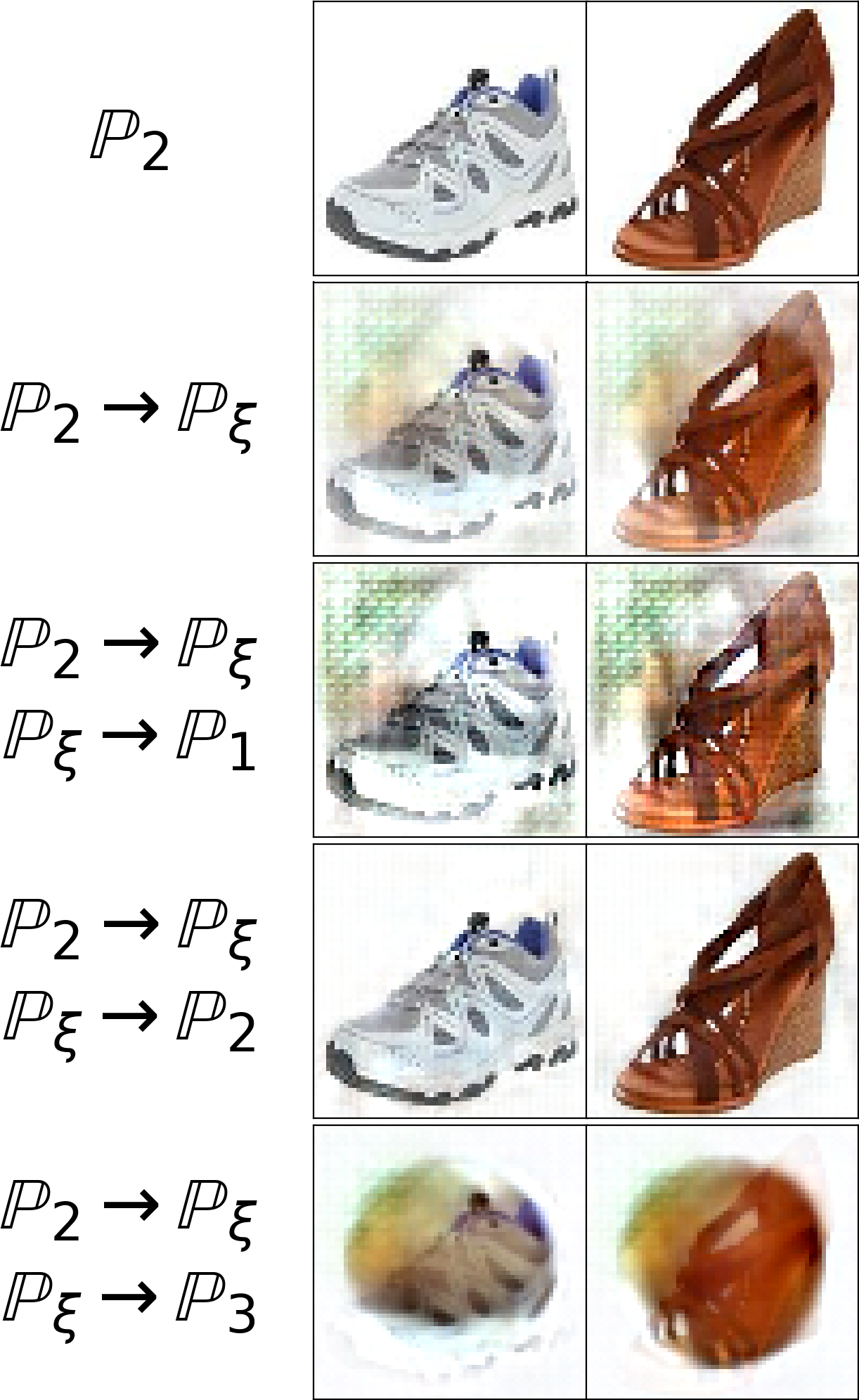}
\caption{\centering Samples $y\sim\mathbb{P}_{2}$ mapped through $\mathbb{P}_{\xi}$ to each $\mathbb{P}_{n}$.}
\label{fig:fruit-through2-icnn}
\end{subfigure}
\begin{subfigure}{0.225\linewidth}
\centering
\includegraphics[width=0.97\linewidth]{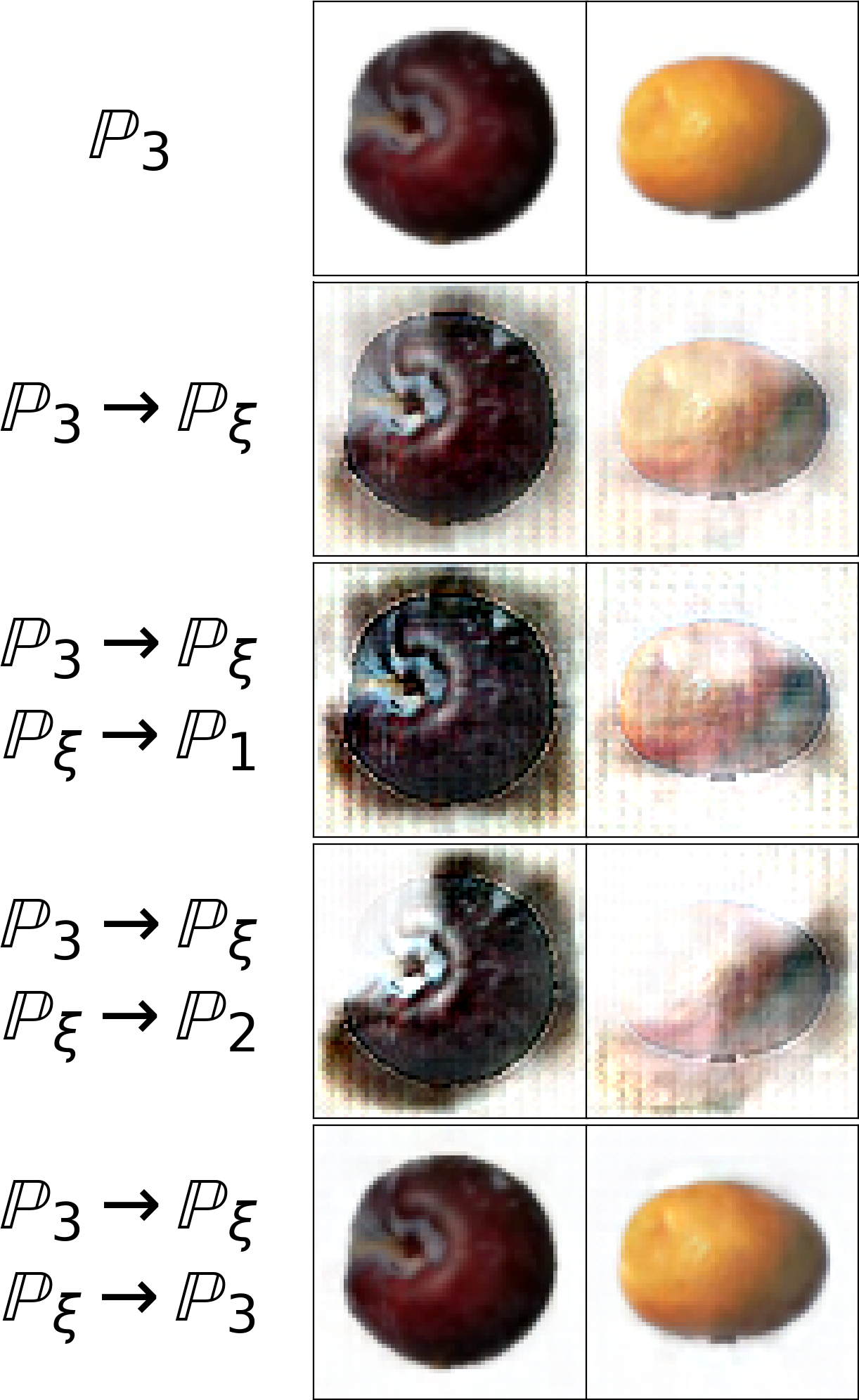}
\caption{\centering Samples $y\sim\mathbb{P}_{3}$ mapped through $\mathbb{P}_{\xi}$ to each $\mathbb{P}_{n}$.}
\label{fig:fruit-through3-icnn}
\end{subfigure}
\caption{The barycenter of Handbags, Shoes, Fruit ($64\times 64$) datasets fitted by {\color{Red}\underline{\textbf{competitive}}} [SC$\mathbb{W}_{2}$B].}
\vspace*{-0.2in}
\label{fig:handbag-shoe-fruit-icnn}
\end{figure*}

On Ave, celeba! we use \citep[Algorithm 1]{fan2020scalable} with $k_{3}=50000$, $k_{2}=k_{1}=10$.\footnote{We also tried training their ICNN-based algorithm in our \textbf{iterative} manner, i.e., by performing multiple regression updates of the generator instead of the single \textbf{variational} update. This provided the same results.} The optimizer, the learning rates and the generator network are the same as in our algorithm. However, for the potentials (OT solver), we use ICNN architecture as it is required by their method. We use ConvICNN64 \citep[Appendix B.1]{korotin2021neural} architecture. For handbags, shoes, fruit (Figure \ref{fig:handbag-shoe-fruit-icnn}), the parameters are the same.

\section{Additional Experimental Results}
\label{sec-exp-extra}

\subsection{Toy Experiments}
\label{sec-toy-experiments}
\begin{figure*}[!h]
\hspace{-5mm}\begin{subfigure}{1.05\linewidth}
\centering
\includegraphics[width=0.99\linewidth]{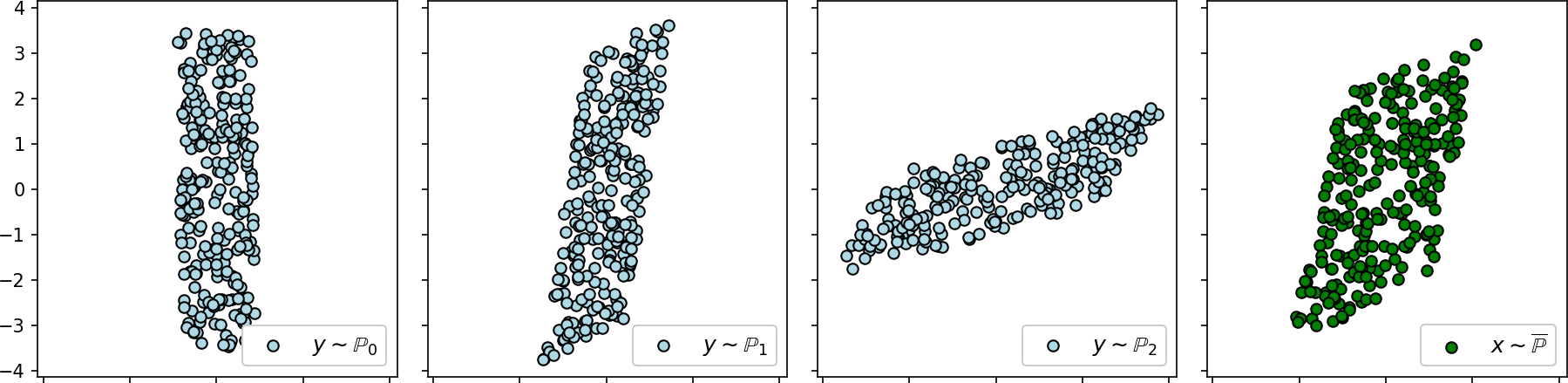}\caption{\centering Input measures $\mathbb{P}_{\xi}$ and their {\color{ForestGreen}\underline{ground truth barycenter}} $\overline{\mathbb{P}}$ w.r.t. weights $\alpha_{1}=\alpha_{2}=\alpha_{3}=\frac{1}{3}$.}
\label{fig:toy-rectangles-1}
\end{subfigure}\vspace{2mm}\newline

\hspace{-5mm}\begin{subfigure}{1.05\linewidth}
\centering
\includegraphics[width=0.99\linewidth]{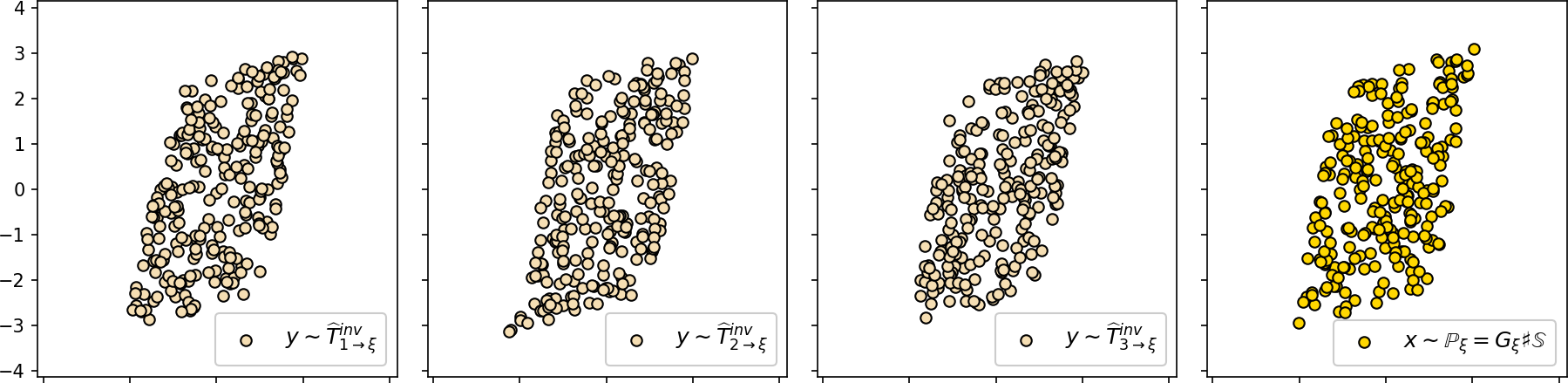}\caption{\centering Learned maps $\mathbb{P}_{n}\rightarrow\mathbb{P}_{\xi}$ from the input measures and the {\color{Goldenrod}\underline{generated barycenter}} $\mathbb{P}_{\xi}=G_{\xi}\sharp\mathbb{S}$.}
\label{fig:toy-rectangles-2}
\end{subfigure}\vspace{2mm}\newline

\hspace{-5mm}\begin{subfigure}{1.05\linewidth}
\centering
\includegraphics[width=0.99\linewidth]{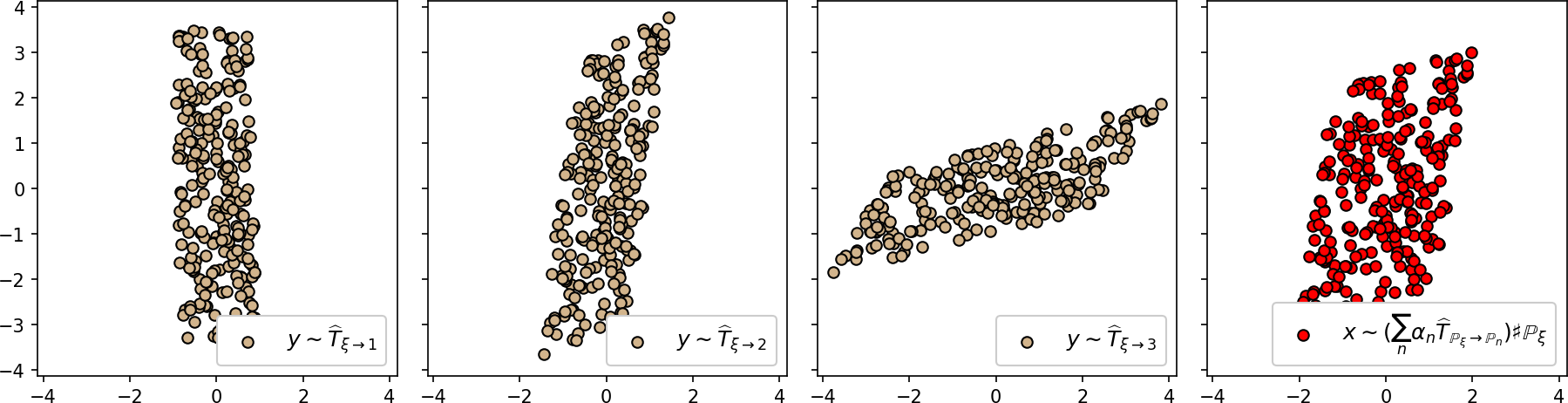}\caption{\centering Learned maps $\mathbb{P}_{\xi}\rightarrow\mathbb{P}_{n}$ from the generated barycenter $\mathbb{P}_{\xi}=G_{\xi}\sharp\mathbb{S}$ to the input measures $\mathbb{P}_{n}$\protect\linebreak and their {\color{red}\underline{weighted average map}} $\sum_{n=1}^{N}\big[\alpha_{n}\widehat{T}_{\mathbb{P}_{\xi}\rightarrow\mathbb{P}_{n}}\big]\sharp\mathbb{P}_{\xi}$.}
\label{fig:toy-rectangles-3}
\end{subfigure}
\caption{\centering  The results of applying our algorithm to compute the barycenter of a 2D\protect\linebreak location-scatter population produced by a rectangle.}
\label{fig:toy-rectangles}
\end{figure*}

\begin{figure*}[!h]
\hspace{-5mm}\begin{subfigure}{1.05\linewidth}
\centering
\includegraphics[width=0.99\linewidth]{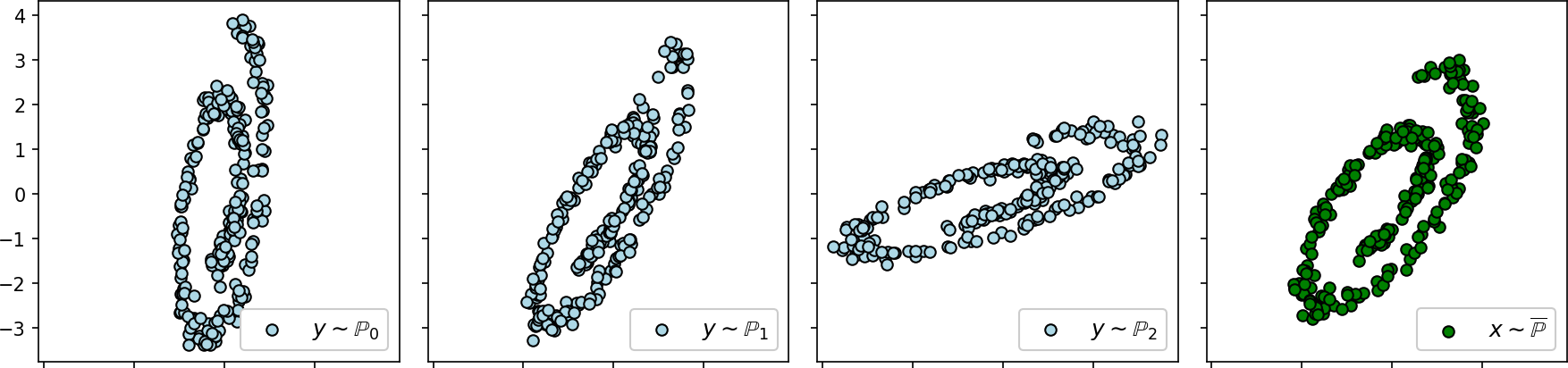}\caption{\centering Input measures $\mathbb{P}_{\xi}$ and their {\color{ForestGreen}\underline{ground truth barycenter}} $\overline{\mathbb{P}}$ w.r.t. weights $\alpha_{1}=\alpha_{2}=\alpha_{3}=\frac{1}{3}$.}
\label{fig:toy-swiss-1}
\end{subfigure}\vspace{2mm}\newline

\hspace{-5mm}\begin{subfigure}{1.05\linewidth}
\centering
\includegraphics[width=0.99\linewidth]{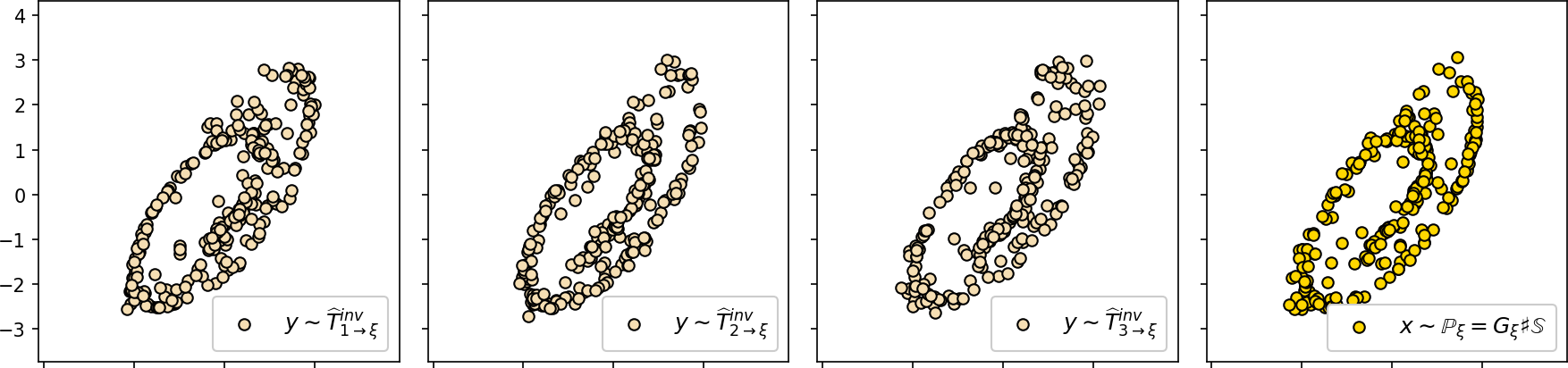}\caption{\centering Learned maps $\mathbb{P}_{n}\rightarrow\mathbb{P}_{\xi}$ from the input measures and the {\color{Goldenrod}\underline{generated barycenter}} $\mathbb{P}_{\xi}=G_{\xi}\sharp\mathbb{S}$.}
\label{fig:toy-swiss-2}
\end{subfigure}\vspace{2mm}\newline

\hspace{-5mm}\begin{subfigure}{1.05\linewidth}
\centering
\includegraphics[width=0.99\linewidth]{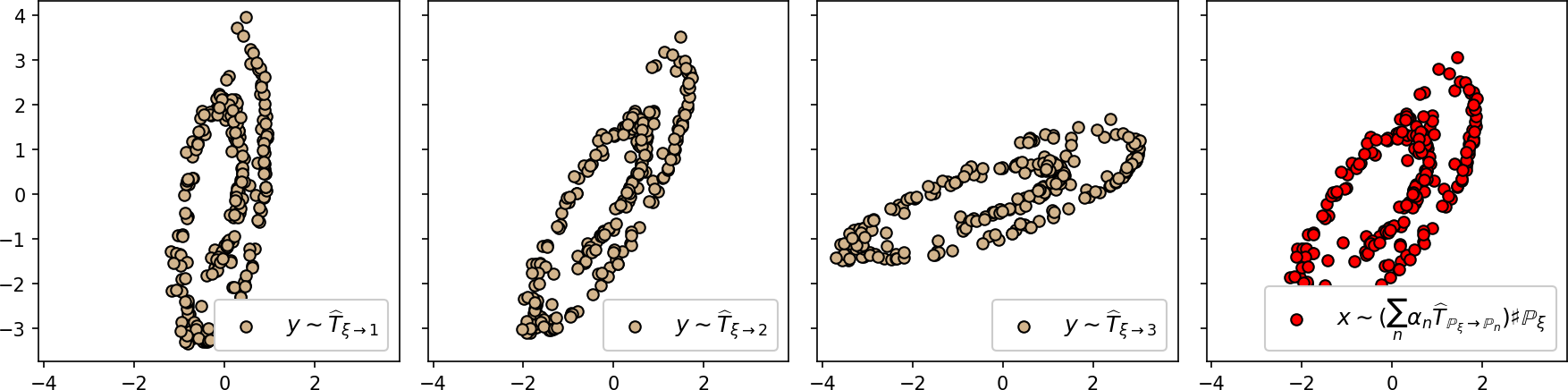}\caption{\centering Learned maps $\mathbb{P}_{\xi}\rightarrow\mathbb{P}_{n}$ from the generated barycenter $\mathbb{P}_{\xi}=G_{\xi}\sharp\mathbb{S}$ to the input measures $\mathbb{P}_{n}$\protect\linebreak and their {\color{red}\underline{weighted average map}} $\sum_{n=1}^{N}\big[\alpha_{n}\widehat{T}_{\mathbb{P}_{\xi}\rightarrow\mathbb{P}_{n}}\big]\sharp\mathbb{P}_{\xi}$.}
\label{fig:toy-swiss-3}
\end{subfigure}
\caption{\centering  The results of applying our algorithm to compute the barycenter of a 2D\protect\linebreak location-scatter population produced by a Swiss roll.}
\label{fig:toy-swiss}
\end{figure*}

In this section, we provide examples of barycenters computed by our Algorithm for 2D location-scatter cases. To produce the location-scatter population of distributions and compute their ground truth barycenters, we employ the publicly available code\footnote{\url{http://github.com/iamalexkorotin/Wasserstein2Barycenters}} of $[\text{C}\mathbb{W}_{2}\text{B}]$ paper \cite{korotin2021continuous}. The hyper-parameters of our Algorithm \ref{algorithm-win} (learning the barycenter and maps to input measures) and Algorithm \ref{algorithm-win-inverse} ($\lceil \text{MM:R}\rfloor$ solver, learning the inverse maps) are given in Tables \ref{table-params} and \ref{table-params-2}, respectively. For evaluation, we consider two location-scatter populations produced by a rectangle and a swiss-roll respectively \cite[\wasyparagraph 5]{korotin2021continuous}. The computed barycenters and maps to/from the input distributions are shown in Figures \ref{fig:toy-rectangles}, \ref{fig:toy-swiss}.

% \lingxiao{Add some summarizing sentence such as: from Figure X and Y, we see that our algorithm produce visually convincing barycenters similar to that of CW2B.}

% \subsection{Toy Examples}
% \label{sec-exp-toy}

\subsection{Location-Scatter Case}
\label{sec-gaussian-case}

Similar to \cite{korotin2021continuous,fan2020scalable}, we consider \textbf{location-scatter} cases for which the true barycenter can be computed \citep[\S4]{alvarez2016fixed}. Let $\mathbb{P}_{0}\in\mathcal{P}_{2,\text{ac}}(\mathbb{R}^{D})$ and define the following location-scatter family of distributions
$\mathcal{F}(\mathbb{P}_{0})=\{f_{S,u}\sharp \mathbb{P}_{0}\mbox{ }\vert\mbox{ } S\in\mathcal{M}^{+}_{D\times D}, u\in\mathbb{R}^{D}\},$
where $f_{S,u}:\mathbb{R}^{D}\rightarrow\mathbb{R}^{D}$ is a linear map $f_{S,u}(x)=Sx+u$ with positive definite matrix $S\in\mathcal{M}^{+}_{D\times D}$.
When $\{\mathbb{P}_{n}\}\subset\mathcal{F}(\mathbb{P}_{0})$, their barycenter $\overline{\mathbb{P}}$ is also an element of $\mathcal{F}(\mathbb{P}_{0})$ and can be computed via fixed-point iterations \citep{alvarez2016fixed}. We use ${N=4}$ measures with weights ${(\alpha_{1},\dots,\alpha_{4})=(\frac{1}{10},\frac{2}{10},\frac{3}{10},\frac{4}{10})}$. We consider two choices for $\mathbb{P}_{0}$: the $D$-dimensional standard \textbf{Gaussian} and the \textbf{uniform} distribution on $[-\sqrt{3},+\sqrt{3}]^{D}$. 
By using the publicly available code of \cite{korotin2021continuous}, we construct $\mathbb{P}_{n}$ as $f_{S_{n}^{T}\Lambda S_{n},0}\sharp\mathbb{P}_{0}\in\mathcal{F}(\mathbb{P}_{0})$, where $S_{n}$ is a random rotation matrix and $\Lambda$ is diagonal with entries $[\frac{1}{2}b^0,\frac{1}{2}b^{1},\dots,2]$ where $b=\sqrt[D-1]{4}$. We quantify the generated barycenter $G_{\xi}\sharp\mathbb{S}$ with the Bures-Wasserstein Unexplained Variance Percentage \citep[\wasyparagraph5]{korotin2021continuous}: $$\text{B}\mathbb{W}_{2}^{2}\text{-UVP}(G_{\xi}\sharp\mathbb{S},\overline{\mathbb{P}})=100\cdot \text{B}\mathbb{W}_{2}^{2}(G_{\xi}\sharp\mathbb{S},\overline{\mathbb{P}})/\big[\frac{1}{2}\text{Var}(\overline{\mathbb{P}})\big]\%,$$
where $\text{B}\mathbb{W}_{2}^{2}(\mathbb{P},\mathbb{Q})=\mathbb{W}_{2}^{2}\big(\mathcal{N}(\mu_{\mathbb{P}},\Sigma_{\mathbb{P}}),\mathcal{N}(\mu_{\mathbb{Q}},\Sigma_{\mathbb{Q}})\big)$ is the Bures-Wasserstein metric and $\mu_{\mathbb{P}}$, $\Sigma_{P}$ denote mean and covariance of $\mathbb{P}$. The metric admits the closed form \citep{chewi2020gradient}. For the trivial baseline prediction $G_{\xi_0}(z)\equiv \mu_{\overline{\mathbb{P}}}\equiv\sum_{n=1}^{N}\alpha_{n}\mu_{\mathbb{P}_{n}}$ the metric value is $100\%$. We denote this baseline as $ \lfloor\text{C}\rceil.$

\begin{table}[!h]
\scriptsize
\centering
\hspace{-15mm}\begin{tabular}{|c|c|c|c|c|c|c|c|c|}
\hline
\textbf{Method}                & \textbf{D=2}    & \textbf{4}    & \textbf{8}    & \textbf{16}    & \textbf{32}    & \textbf{64}          & \textbf{128}                 \\ \hline
$ \lfloor\text{C}\rceil$ & 100 & 100 & 100 & 100 & 100 & 100 & 100\\ \hline
[SC$\mathbb{W}_{2}$B] & 0.07 & 0.09 & 0.16 & 0.28  & 0.43  & 0.59        & 1.28         \\ \hline
\textbf{Ours} & 0.01 & 0.02 & 0.01 & 0.08  & 0.11  & 0.23 & 0.38  \\ \hline
\end{tabular}\hspace{2mm}
\begin{tabular}{|c|c|c|c|c|c|c|c|c|}
\hline
\textbf{Method}                & \textbf{D=2}    & \textbf{4}    & \textbf{8}    & \textbf{16}    & \textbf{32}    & \textbf{64}          & \textbf{128}                 \\ \hline
$ \lfloor\text{C}\rceil$ & 100 & 100 & 100 & 100 & 100 & 100 & 100\\ \hline
[SC$\mathbb{W}_{2}$B] & 0.12 & 0.10 & 0.19 & 0.29 & 0.46 & 0.6 & 1.38        \\ \hline
\textbf{Ours} & 0.04 & 0.06 & 0.06 & 0.08  & 0.11 & 0.27 & 0.46  \\ \hline
\end{tabular}\hspace{-18mm}
% \vspace{-.1in}
\vspace{1mm}
\caption{\centering Comparison of B$\mathbb{W}_{2}^{2}$-UVP$\downarrow$ (\%) in the location-scatter cases:\protect\linebreak ${\mathbb{P}_{0}=\mathcal{N}(0, I_{D})}$ on the \textbf{left} and ${\mathbb{P}_{0}=\text{Uniform}\big([-\sqrt{3},+\sqrt{3}]^{D}}\big)$ on the \textbf{right}.}
% \vspace{-.20in}
\label{table-gaussians}
\end{table}

The results of our algorithm \ref{algorithm-win} and $[\text{SC}\mathbb{W}_{2}\text{B}]$ adapted from \citep[Table 1]{korotin2021continuous} are given in Table \ref{table-gaussians}. Both algorithms work well in the location-scatter cases and provide $\text{B}\mathbb{W}_{2}\text{-UVP}<2\%$ in dimension 128.

\subsection{Generative Modeling}
\label{sec-generative-modeling}

Analogously to \cite{fan2020scalable}, we evaluate our algorithm when $N=1$. In this case, the minimizer of \eqref{w2-barycenter-def} is the measure $\mathbb{P}_{1}$ itself, i.e., $\overline{\mathbb{P}}=\mathbb{P}_{1}$. As the result, our algorithm \ref{algorithm-win} works as a usual generative model, i.e., it fits data $\mathbb{P}_{1}$ by a generator $G_{\xi}$. For experiments, we use CelebA $64\times 64$ dataset. Generated images $G_{\xi}(z)$ and $\widehat{T}_{\mathbb{P}_{\xi}\rightarrow\mathbb{P}_{1}}\big(G_{\xi}(z)\big)$ are shown in Figure \ref{fig:celeba-win}.

\begin{figure*}[!h]
    \centering
    \begin{subfigure}[b]{0.49\textwidth}
         \centering
         \includegraphics[width=0.99\linewidth]{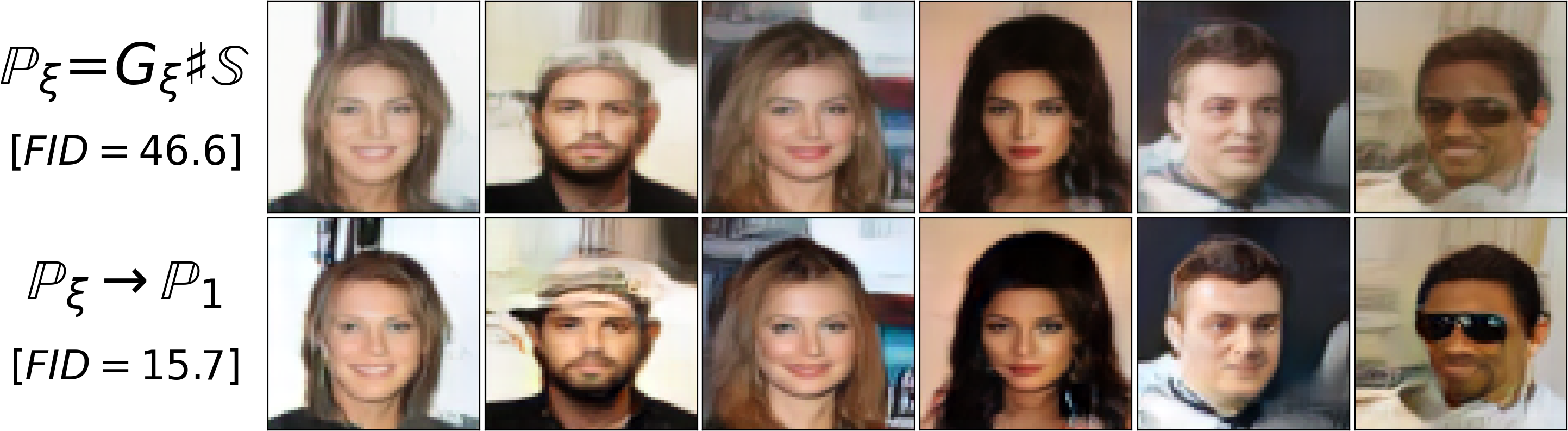}
        \caption{Generator $G_{\xi}$ training enabled ($K_{G}>0$).}
        \label{fig:celeba-win}
    \end{subfigure}\hfill
     \begin{subfigure}[b]{0.49\textwidth}
        \centering
        \includegraphics[width=0.99\linewidth]{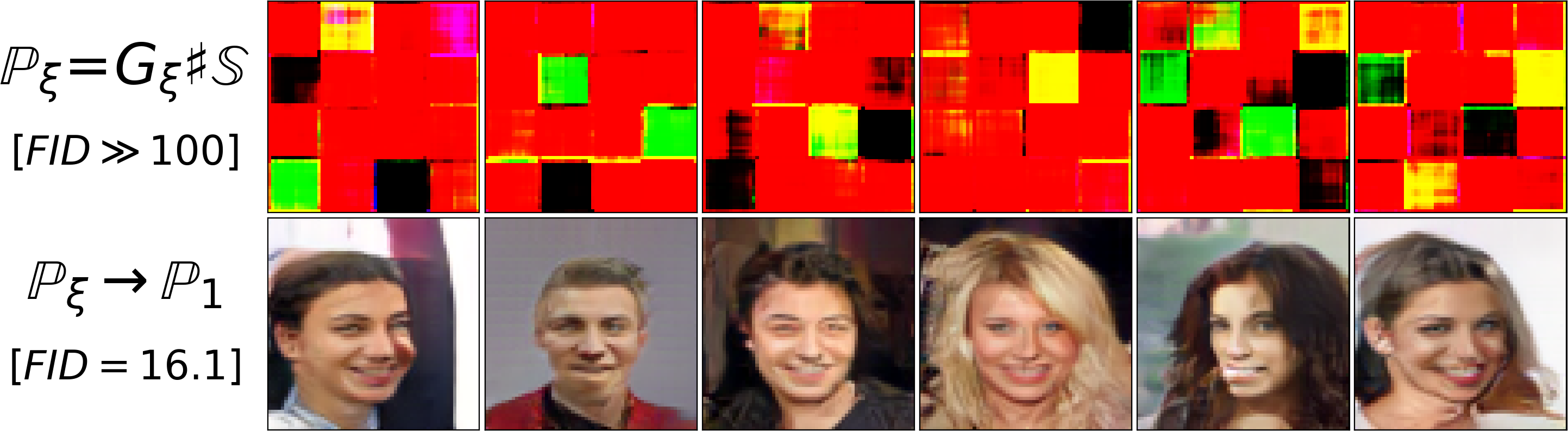}
        \caption{Generator $G_{\xi}$ training disabled ($K_{G}=0$).}
        \label{fig:celeba-win-fixed}
     \end{subfigure}
    \caption{Images generated by our algorithm \ref{algorithm-win} serving as a generative model. The 1st line shows samples from ${G_{\xi}\sharp \mathbb{S}\approx \mathbb{P}_{\text{1}}}$, the 2nd line shows estimated OT map $\widehat{T}_{\mathbb{P}_{\xi}\rightarrow\mathbb{P}_{1}}$ from $G_{\xi}\sharp \mathbb{S}$ to $\mathbb{P}_{1}$ which further improves generated images.}
\end{figure*}

In Table \ref{table-fid-celeba}, we provide FID for generated images. For comparison, we include FID for ICNN-based $[\text{SC}\mathbb{W}_{2}\text{B}]$, 
and WGAN-QC \citep{liu2019wasserstein}. FID scores are adapted from \citep[\wasyparagraph 4.5]{korotin2021neural}. Note that for $N=1$, $[\text{SC}\mathbb{W}_{2}\text{B}]$ is reduced to the OT solver by \cite{makkuva2019optimal} used as the loss for generative models, a setup tested in \citep[Figure 3a]{korotin2021neural}.
% Our method and $[\text{SC}\mathbb{W}_{2}\text{B}]$ also provide an approximate OT map $\widehat{T}_{\mathbb{P}_{\xi}\rightarrow\mathbb{P}_{1}}$
% (in case of $[\text{SC}\mathbb{W}_{2}\text{B}]$ it is a gradient of ICNN $\psi_{\theta_{1}}$)
% from generated images to data which further improves samples.
Serving as a generative model when $N=1$, our algorithm \ref{algorithm-win} performs comparably to WGAN-QC and \textit{drastically} outperforms ICNN-based $[\text{SC}\mathbb{W}_{2}\text{B}]$.

\begin{table}[!h]
\centering
\footnotesize
\vspace{-2mm}\begin{tabular}{|c|c|c|}
\hline
\multicolumn{2}{|c|}{\textit{Method}} & \textit{FID}$\downarrow$  \\ \hline
\multirow{2}{*}{\shortstack[c]{$[\text{SC}\mathbb{W}_{2}B]$}} & $G_{\xi}(z)$ &  \color{red}{90.2}   \\ \cline{2-3}
 & $\widehat{T}_{\mathbb{P}_{\xi}\rightarrow\mathbb{P}_{1}}\big(G_{\xi}(z)\big)$ & \color{red}{89.8}    \\
\hline
WGAN-QC & $G_{\xi}(z)$ & 14.4    \\
\hline
\multirow{2}{*}{\textbf{Ours}} & $G_{\xi}(z)$ &  46.6   \\ \cline{2-3}
 & $\widehat{T}_{\mathbb{P}_{\xi}\rightarrow\mathbb{P}_{1}}\big(G_{\xi}(z)\big)$ & 15.7    \\ \hline
\multirow{2}{*}{\textbf{Ours} (fixed $G_{\xi}$)} & $G_{\xi}(z)$ &  N/A   \\ \cline{2-3}
 & $\widehat{T}_{\mathbb{P}_{\xi}\rightarrow\mathbb{P}_{1}}\big(G_{\xi}(z)\big)$ & 16.1   \\
\hline
\end{tabular}
\vspace{1mm}
\caption{FID scores of generated faces.}
% \vspace{1mm}
\label{table-fid-celeba}
\end{table}

\textbf{Fixed generator.} For $N\!=\!1$, the fixed point approach \wasyparagraph\ref{sec-fixed-point} converges in only one step since operator $\mathcal{H}$ immediately maps $G_{\xi}\sharp\mathbb{S}$ to $\mathbb{P}_{1}$. As a result, in our algorithm \ref{algorithm-win}, \textit{exclusively} when $N=1$, we can fix generator $G_{\xi}$ and train only OT map $T_{\theta_{1}}$ from $G_{\xi}\sharp\mathbb{S}$ to data measure $\mathbb{P}_{1}$ and related potential $v_{\omega_{1}}$. As a sanity check, we conduct such an experiment with randomly initialized generator network $G_{\xi}$. The results are given in Figure \ref{fig:celeba-win-fixed}, the FID is included in Table \ref{table-fid-celeba}. Our algorithm performs well even \textit{without generator training} at all.

\subsection{Barycenters of MNIST Digits and FashionMNIST Classes}
\label{sec-exp-mnist}

\begin{figure*}[!h]
    \centering
    \begin{subfigure}[b]{0.98\textwidth}
        \includegraphics[width=0.975\linewidth]{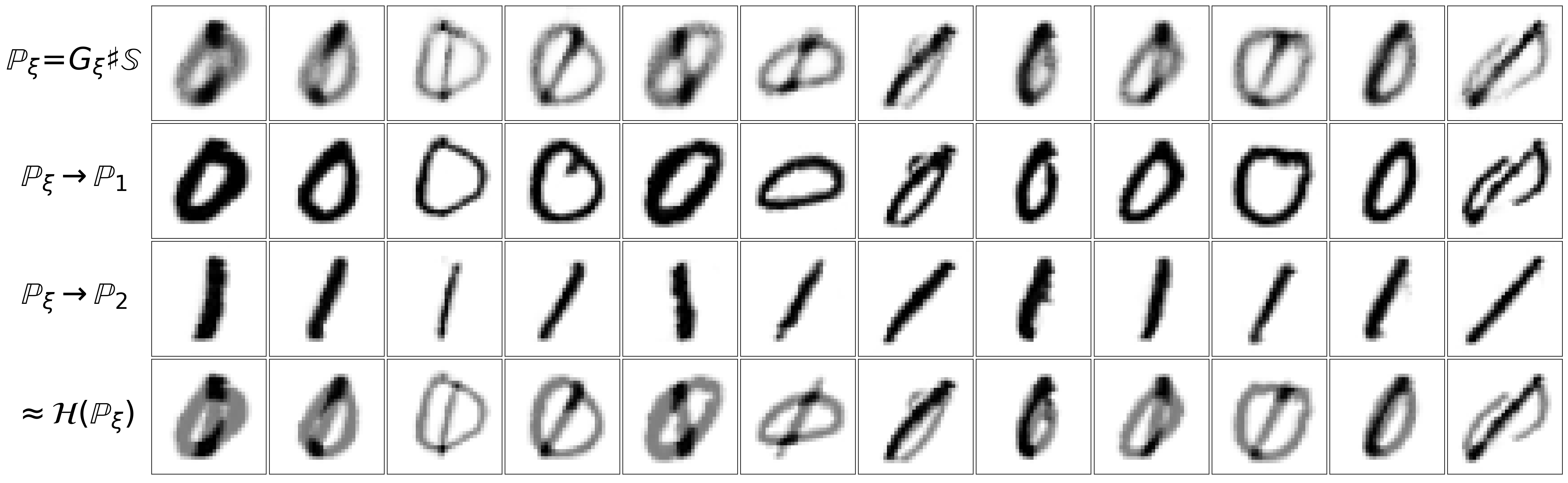}
        \caption{The barycenter $\mathbb{P}_{\xi}$ and maps to input measures $\mathbb{P}_{n}$.}
    \end{subfigure}\vspace{4mm}
    \begin{subfigure}[b]{0.48\textwidth}
         \centering
         \includegraphics[width=0.92\linewidth]{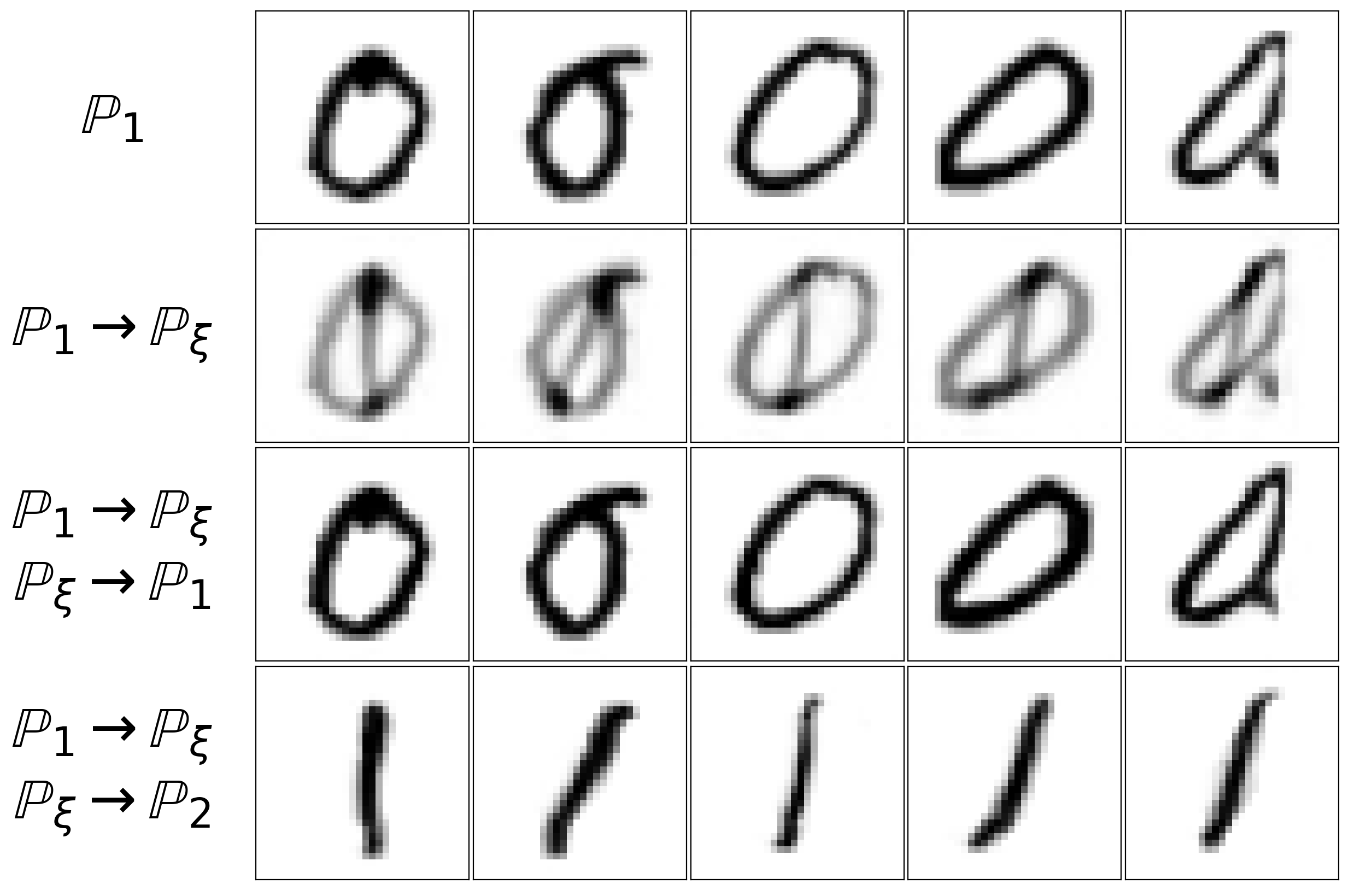}
        \caption{Samples from $\mathbb{P}_{1}$ mapped through $\mathbb{P}_{\xi}$ to each $\mathbb{P}_{n}$.}
    \end{subfigure}\hspace{3mm}
     \begin{subfigure}[b]{0.48\textwidth}
        \centering
        \includegraphics[width=0.92\linewidth]{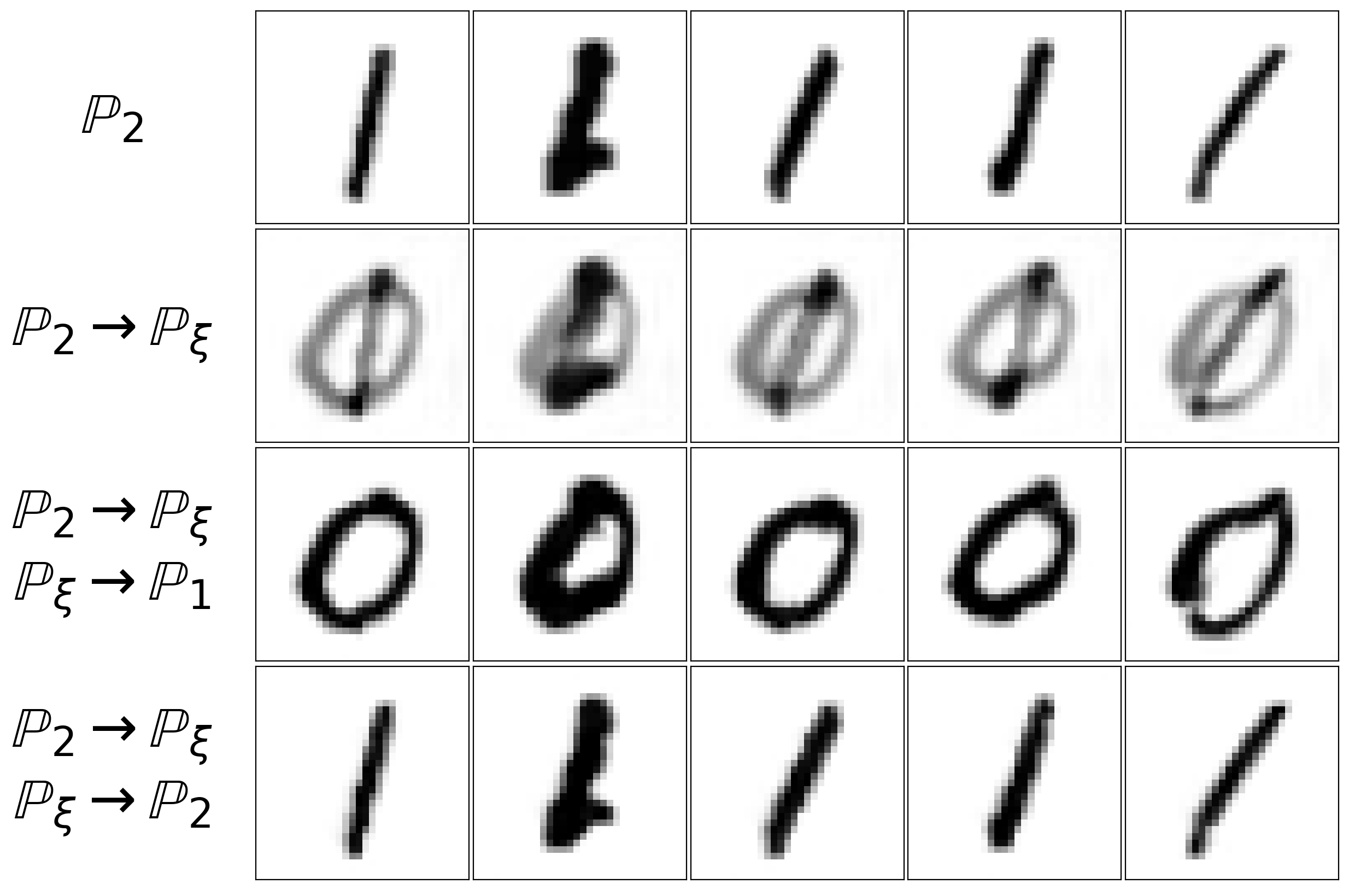}
        \caption{Samples from $\mathbb{P}_{2}$ mapped through $\mathbb{P}_{\xi}$ to each $\mathbb{P}_{n}$.}
     \end{subfigure}
    \caption{The barycenter of MNIST digit classes 0/1 learned by Algorithm \ref{algorithm-win}.}
    \label{fig:mnist-win}
\end{figure*}

\begin{figure}[!t]
\includegraphics[width=0.94\linewidth]{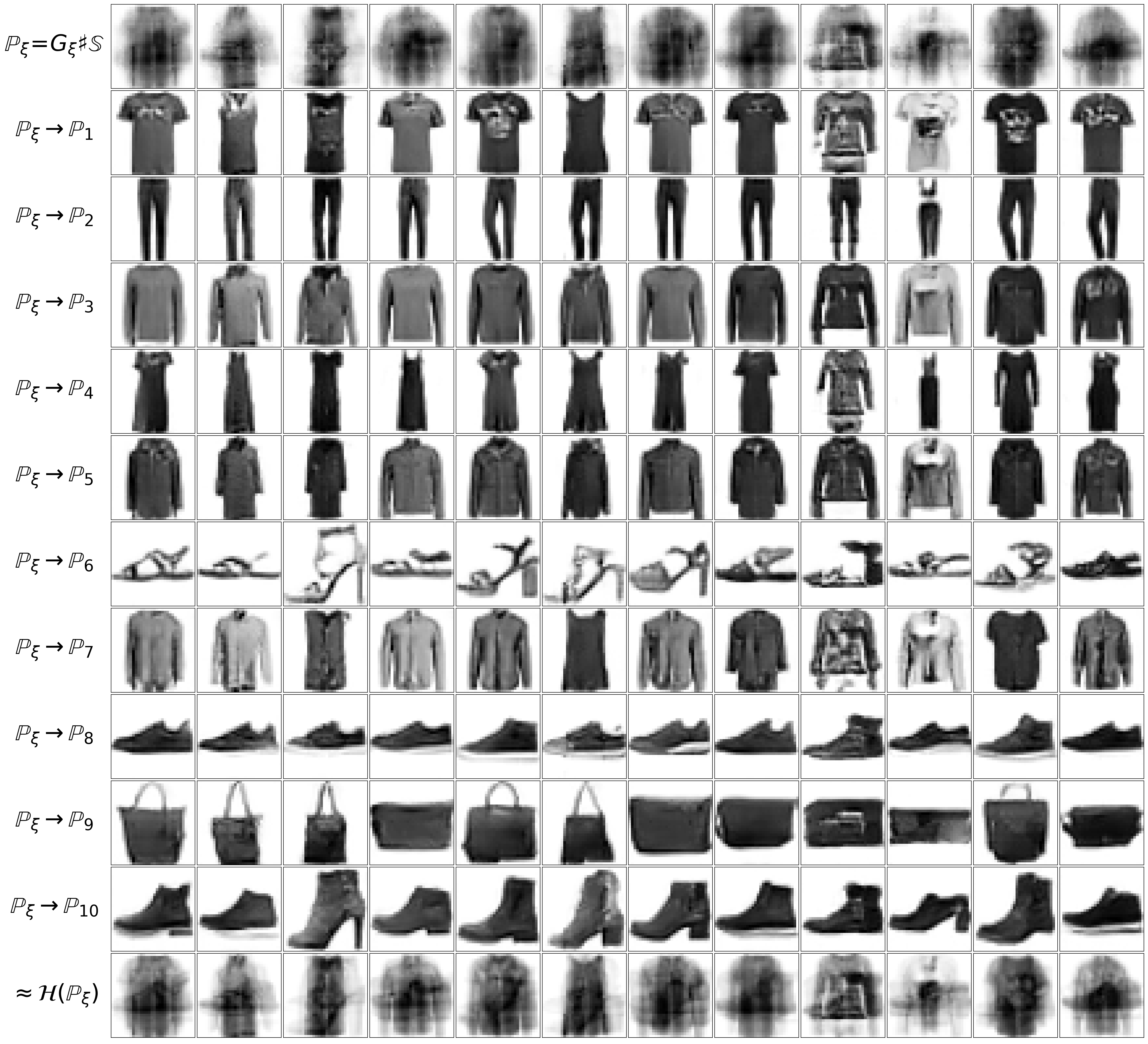}
\caption{The barycenter and maps to input measures estimated by our method on 10 FashionMNIST classes (32$\times$32). The 1st line shows generated samples from $\mathbb{P}_{\xi}=G_{\xi}\sharp \mathbb{S}\approx \overline{\mathbb{P}}$. Each of 10 next lines shows estimated optimal maps $\widehat{T}_{\mathbb{P}_{\xi}\rightarrow\mathbb{P}_{n}}$ to measures $\mathbb{P}_{n}$. The last line shows average $\big[\sum_{n=1}^{N}\alpha_{n}\widehat{T}_{\mathbb{P}_{\xi}\rightarrow\mathbb{P}_{n}}\big]\sharp\mathbb{P}_{\xi}$.}
\label{fig:fashionmnist-win}
\end{figure}

\begin{figure}[!t]
\includegraphics[width=0.94\linewidth]{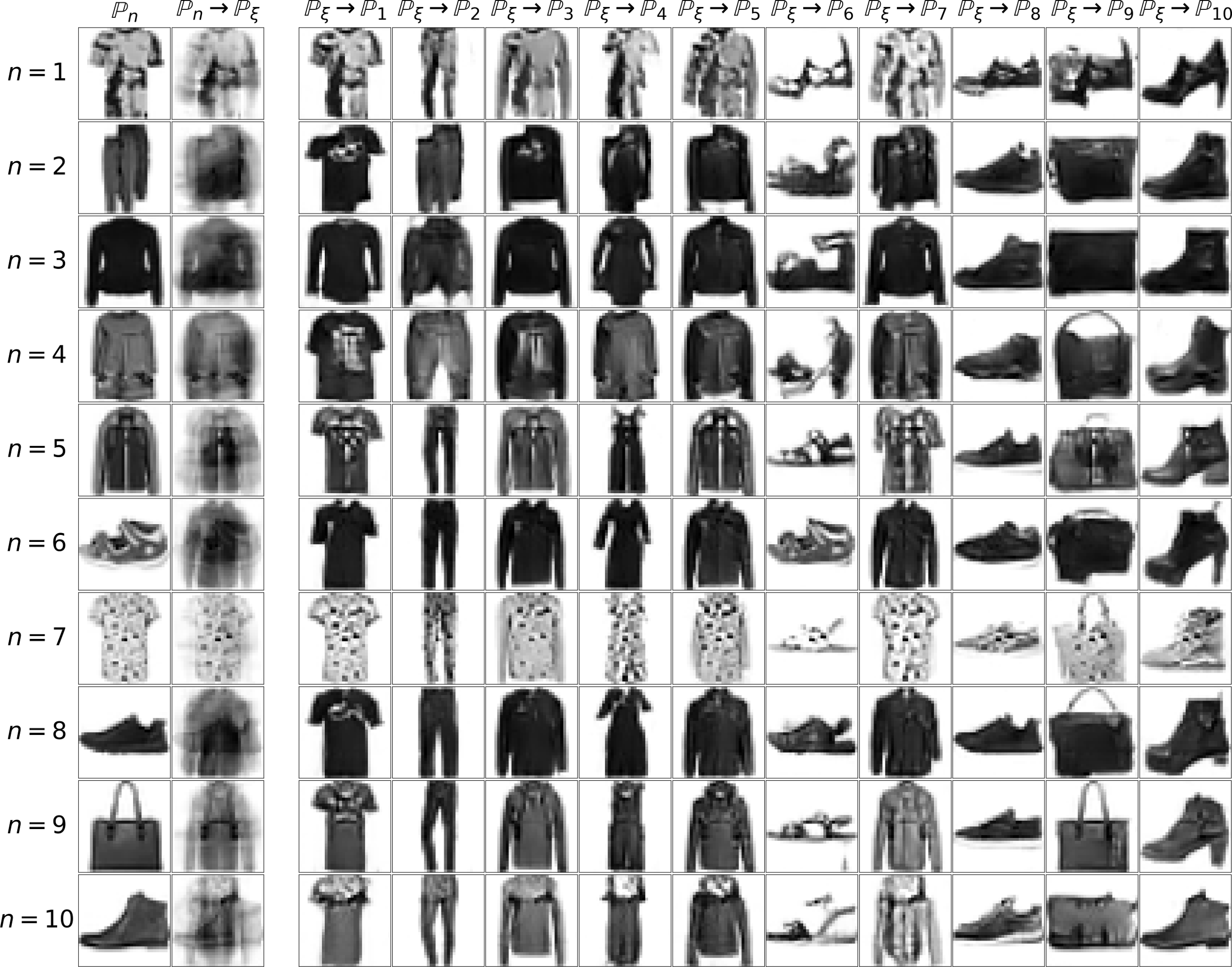}
\caption{Maps between FashionMNIST classes through the learned barycenter. The 1st images in each $n$-th column shows a sample from $\mathbb{P}_{n}$. The 2nd columns maps these samples to the barycenter. Each next column shows how the maps from the barycenter to the input classes $\mathbb{P}_{n}$.}
\label{fig:fashionmnist-win-through}
\end{figure}

Similar to \citep[Figure 6]{fan2020scalable}, we provide qualitative results of our algorithm applied to computing the barycenter of two MNIST classes of digits $0,1$. The barycenter w.r.t. weights $(\frac{1}{2},\frac{1}{2})$ computed by our algorithm is shown in Figure \ref{fig:mnist-win}. We also consider a more complex FashionMNIST \citep{xiao2017fashion} dataset. Here we compute the barycenter of 10 classes w.r.t. weights $(\frac{1}{10},\dots,\frac{1}{10})$. The results are given in Figures \ref{fig:fashionmnist-win} and Figure \ref{fig:fashionmnist-win-through}.

Due to \eqref{bar-condition}, each barycenter images are an average (in pixel space) of certain images  from the input measure. In all the Figures, the produced barycenter images satisfy this property. The maps to input measures are visually good. The approximate fixed point operator $\mathcal{H}(\mathbb{P}_{\xi})$ is almost the identity as expected (the method converged).
% Interestingly, it further improves generated barycenter samples $\mathbb{P}_{\xi}$.

\subsection{Additional Results}
\label{sec-extra-results}

\begin{figure*}[!t]
\centering
\begin{subfigure}[b]{0.99\linewidth}
\centering
\includegraphics[width=0.99\linewidth]{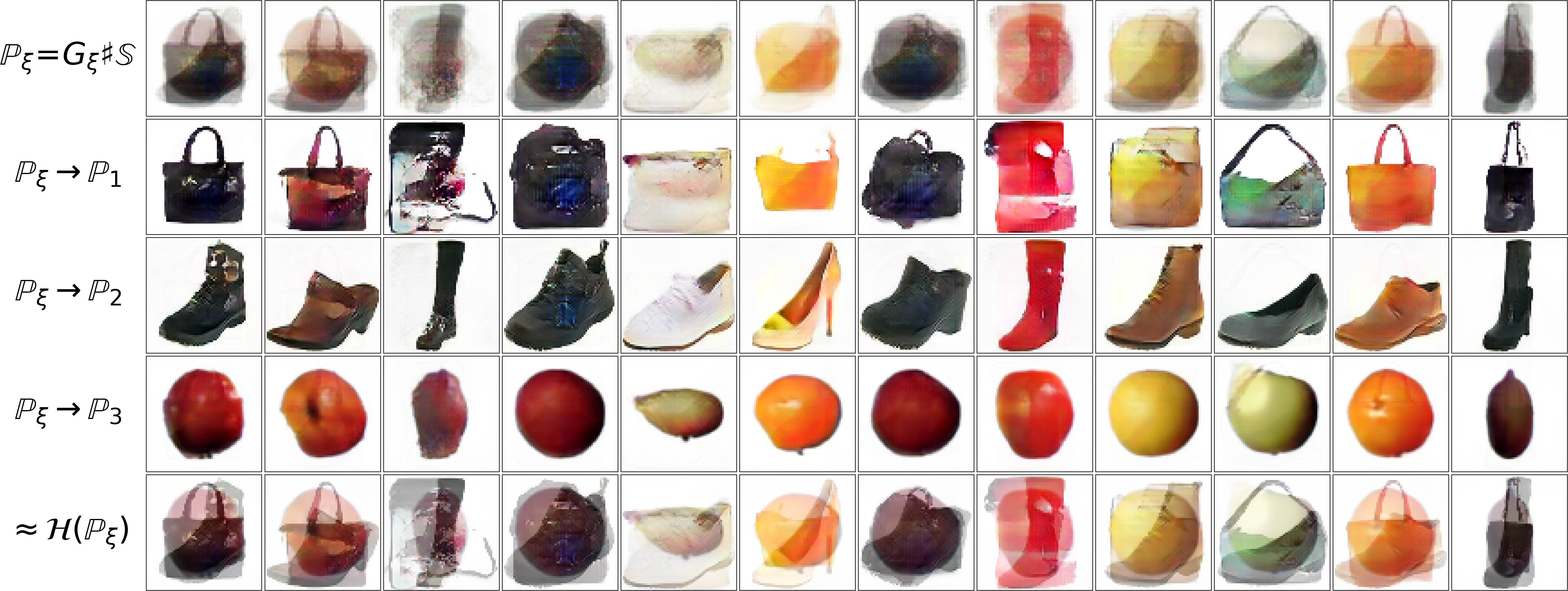}\caption{\centering Generated samples $\mathbb{P}_{\xi}\approx\overline{\mathbb{P}}$, fitted maps to each $\mathbb{P}_{n}$ and their average.}
\label{fig:fruit-generated}
\end{subfigure}\vspace{4mm}
\begin{subfigure}[b]{0.31\linewidth}
\centering
\includegraphics[width=0.99\linewidth]{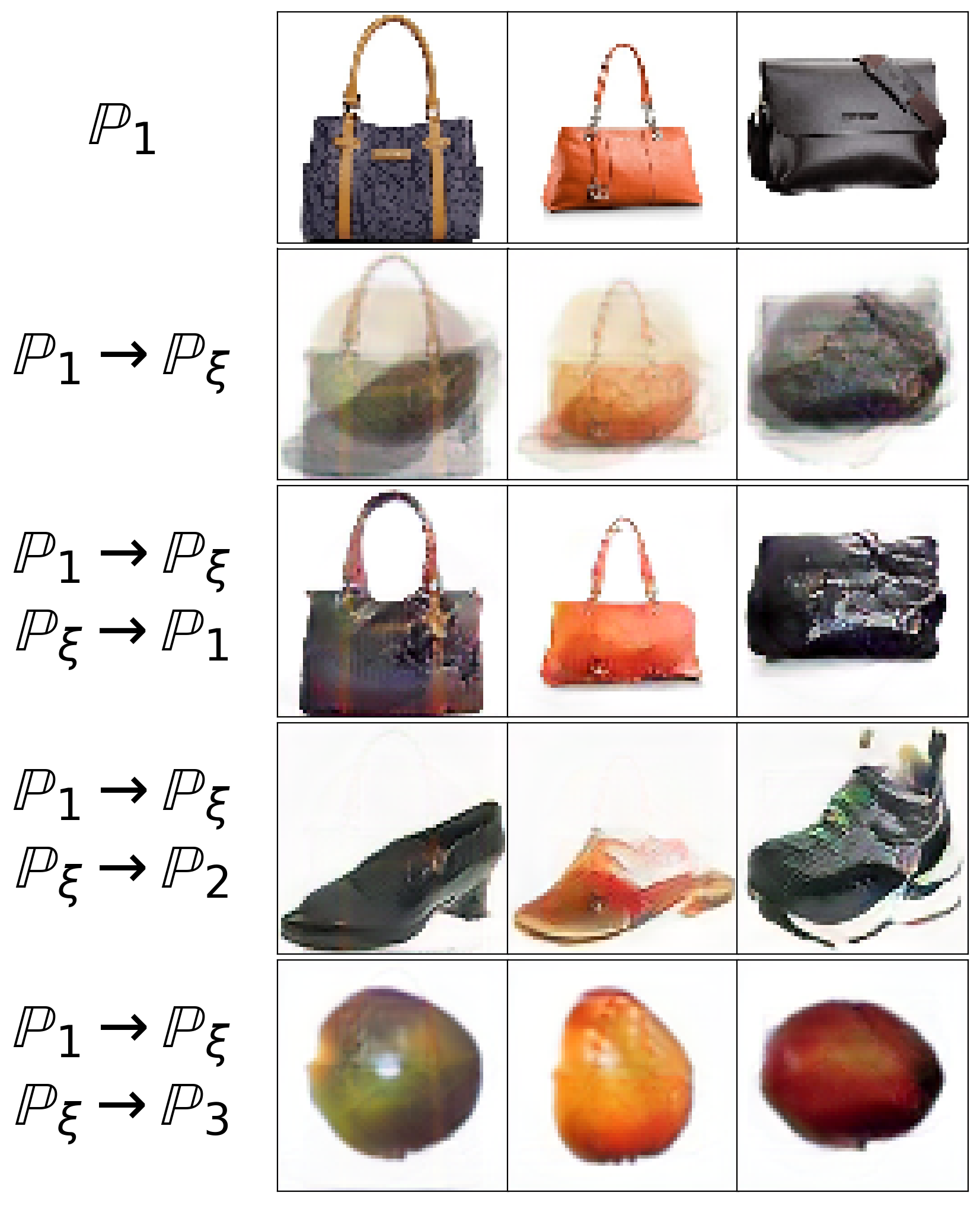}
\caption{\centering Samples $y\sim\mathbb{P}_{1}$ mapped through $\mathbb{P}_{\xi}$ to each $\mathbb{P}_{n}$.}
\label{fig:fruit-through1-ext}
\end{subfigure}
\begin{subfigure}[b]{0.31\linewidth}
\centering
\includegraphics[width=0.99\linewidth]{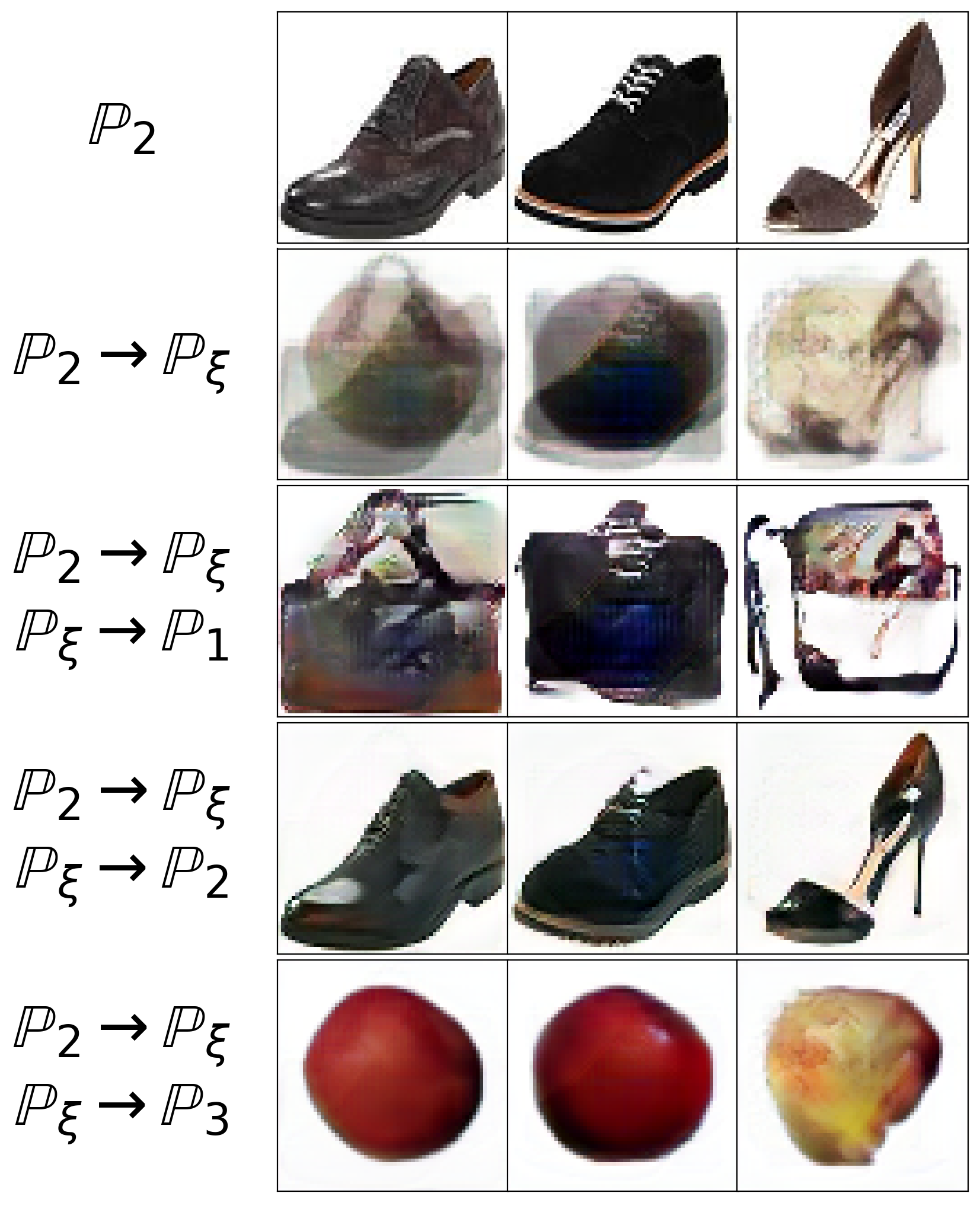}
\caption{\centering Samples $y\sim\mathbb{P}_{2}$ mapped through $\mathbb{P}_{\xi}$ to each $\mathbb{P}_{n}$.}
\label{fig:fruit-through2-ext}
\end{subfigure}
\begin{subfigure}[b]{0.31\linewidth}
\centering
\includegraphics[width=0.99\linewidth]{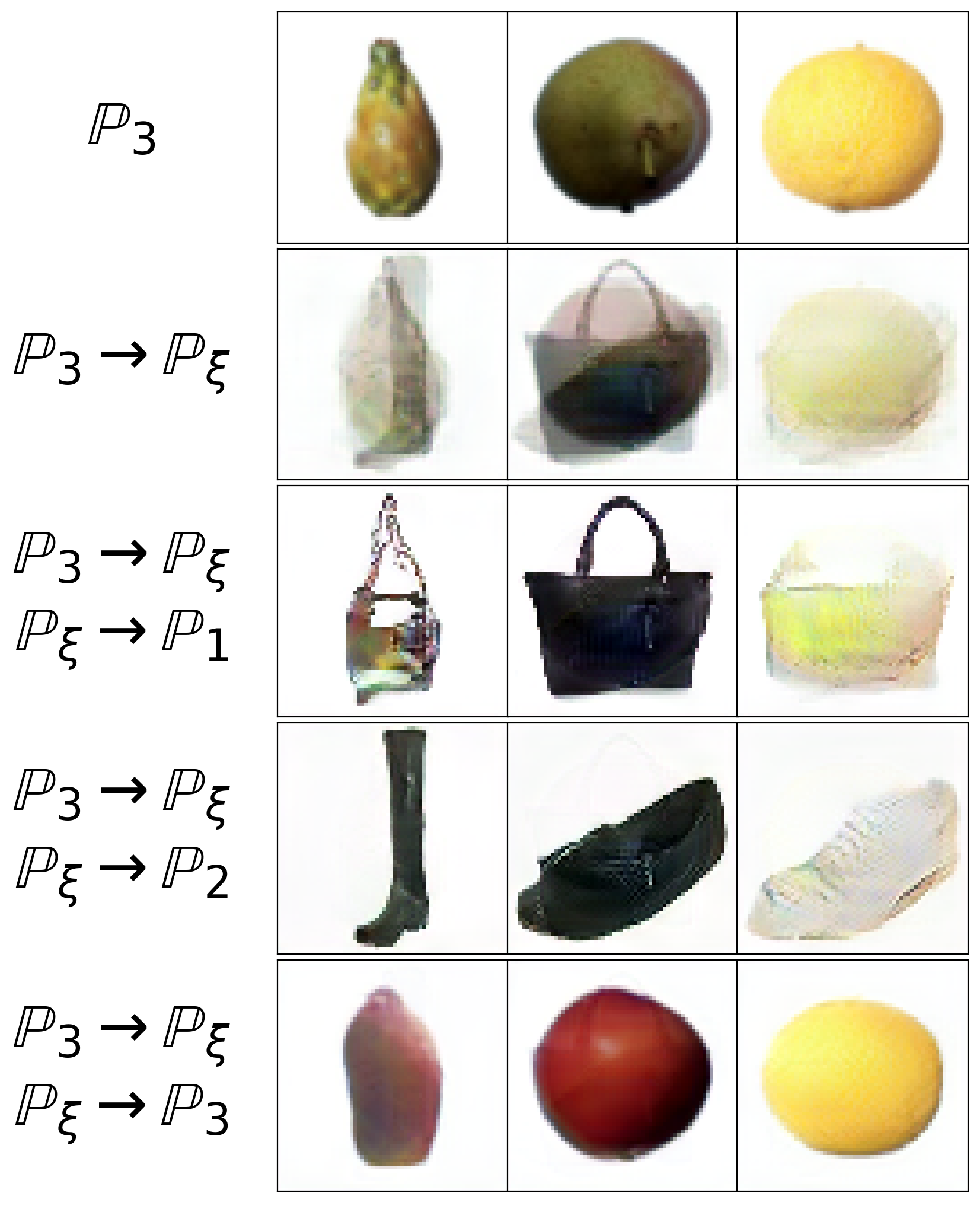}
\caption{\centering Samples $y\sim\mathbb{P}_{3}$ mapped through $\mathbb{P}_{\xi}$ to each $\mathbb{P}_{n}$.}
\label{fig:fruit-through3-ext}
\end{subfigure}
\caption{The barycenter of Handbags, Shoes, Fruit ($64\times 64$) datasets fitted by our algorithm \ref{algorithm-win}.}
\label{fig:handbag-shoe-fruit-ext}
\end{figure*}

\begin{figure*}[!t]
\begin{subfigure}[b]{0.33\linewidth}
\centering
\includegraphics[width=0.99\linewidth]{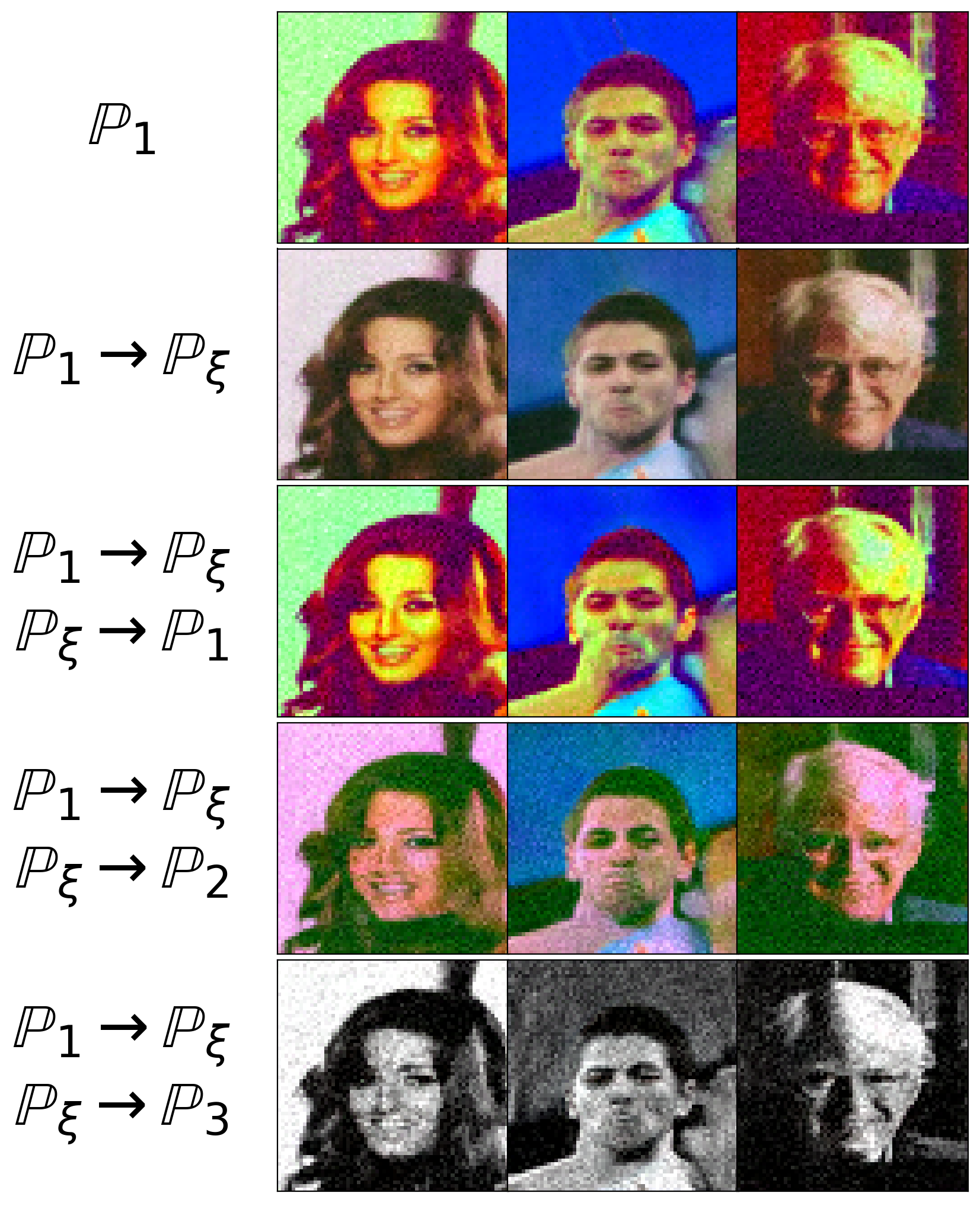}
\caption{\centering Samples $y\sim\mathbb{P}_{1}$ mapped through $\mathbb{P}_{\xi}$ to each $\mathbb{P}_{n}$.}
\end{subfigure}
\begin{subfigure}[b]{0.33\linewidth}
\centering
\includegraphics[width=0.99\linewidth]{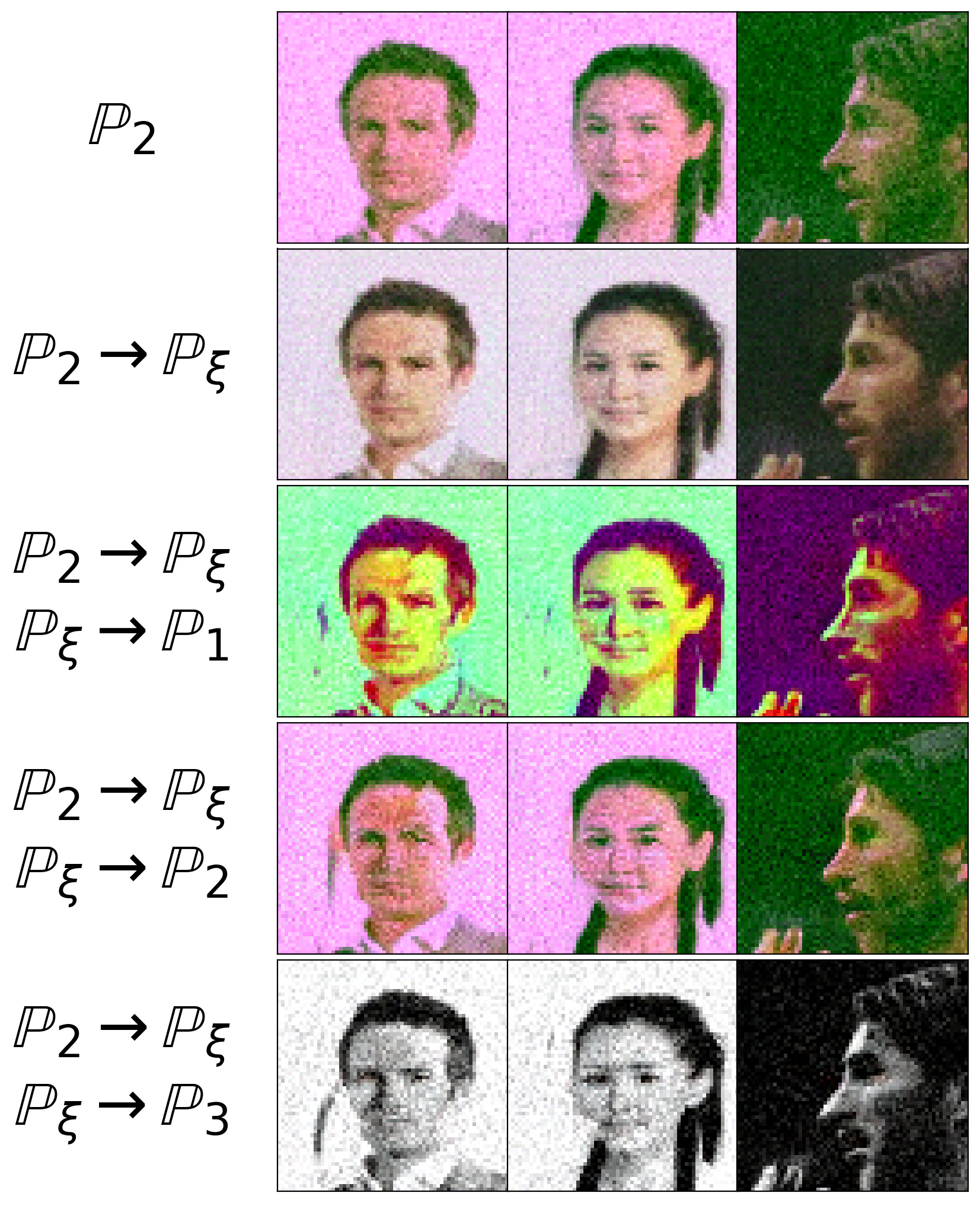}
\caption{\centering Samples $y\sim\mathbb{P}_{2}$ mapped through $\mathbb{P}_{\xi}$ to each $\mathbb{P}_{n}$.}
\end{subfigure}
\begin{subfigure}[b]{0.33\linewidth}
\centering
\includegraphics[width=0.99\linewidth]{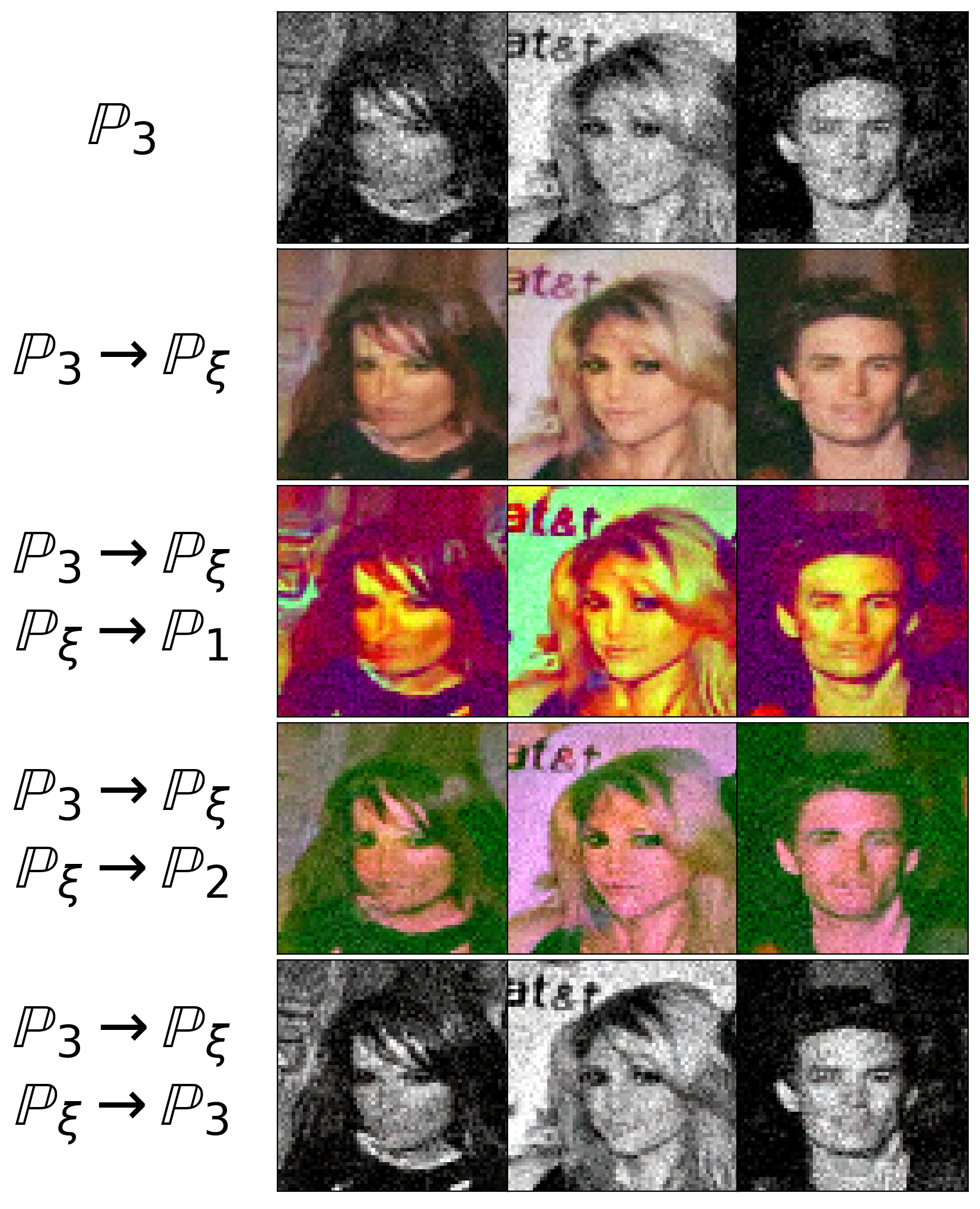}
\caption{\centering Samples $y\sim\mathbb{P}_{3}$ mapped through $\mathbb{P}_{\xi}$ to each $\mathbb{P}_{n}$.}
\end{subfigure}
\caption{Maps between subsets of Ave, celeba! dataset through the barycenter learned by our algorithm \ref{algorithm-win}.}
\label{fig:ave-celeba-through-ext}
\end{figure*}

In Figure \ref{fig:ave-celeba-through-ext}, we visualize maps between Ave, Celeba! subsets through the learned barycenter. In Figure \ref{fig:handbag-shoe-fruit-ext}, we provide additional qualitative results for computing barycenters of Handbags, Shoes, Fruit360 datasets.

\end{document}